\documentclass[a4paper,11pt]{article}
\usepackage{geometry}
\usepackage{graphicx}

\usepackage{mathptmx}      

\usepackage[utf8]{inputenc} 
\usepackage[T1]{fontenc}    
\usepackage{url}            
\usepackage{booktabs}       
\usepackage{nicefrac}       
\usepackage{microtype}      

\usepackage{tikz}
\usepackage{tikz-qtree}
\usetikzlibrary{arrows}

\usepackage{times}
\usepackage{graphicx} 
\usepackage{subcaption} 
\usepackage{amsfonts,amsmath,amssymb}
\usepackage{color}
\usepackage{amsthm}
\usepackage{comment}
\usepackage{paralist}
\usepackage{enumitem}
\setitemize{noitemsep,topsep=0pt,parsep=0pt,partopsep=0pt}

\usepackage{mathtools}

\usepackage{algorithm}
\usepackage{algorithmic}

\usepackage[normalem]{ulem}

\usepackage{caption}

\usepackage[sort,nocompress]{cite}

\usepackage{empheq}

\usepackage{url}
\usepackage[colorlinks = true, pdfstartview = FitV, linkcolor = blue, citecolor = blue, urlcolor = blue]{hyperref}

\newcommand {\uu}  { {\bf u} }

\newcommand {\bss}  { {\bf s} }

\newcommand {\xx}  { {\bf x} }
\renewcommand {\aa}  { {\bf a} }

\newcommand {\bll}  { {\bf l} }

\newcommand {\yy}  { {\bf y} }

\newcommand {\rr}  { {\bf r} }

\newcommand {\qq}  { {\bf q} }

\newcommand {\bb}  { {\bf b} }

\newcommand {\ww}  { {\bf w} }

\newcommand {\cc}  { {\bf c} }

\DeclareMathOperator*{\argmin}{arg\,min}
\DeclareMathOperator*{\argmax}{arg\,max}

\newfloat{subroutine}{htbp}{loa}
\floatname{subroutine}{Procedure}
\makeatletter\newcommand\l@subroutine{\@dottedtocline{1}{1.5em}{2.3em}}\makeatother

\usepackage{arydshln}

\begin{document}

\title{Parallel and Communication Avoiding Least Angle Regression}

\author{
	Swapnil~Das\thanks{S.~Das is with the Computer Science Division and Department of Mathematics, University of California Berkeley, 389 Soda Hall, Berkeley, CA 94720-1776, USA. e-mail: tracer@berkeley.edu.} \and
        James~Demmel\thanks{J.~Demmel is with the Computer Science Division and Department of Mathematics, University of California Berkeley, 389 Soda Hall, Berkeley, CA 94720-1776, USA. e-mail: demmel@berkeley.edu.} \and
        Kimon~Fountoulakis\thanks{K.~Fountoulakis is with the School of Computer Science, University of Waterloo, 200 University Avenue West, Waterloo, ON N2L3G1, Canada. e-mail: kfountou@uwaterloo.ca.} \and
        Laura~Grigori\thanks{L.~Grigori is with the INRIA Paris, Alpines group, France, Paris. e-mail: laura.grigori@inria.fr.} \and
        Michael.~W.~Mahoney\thanks{M.~Mahoney is with the International Computer Science Institute, Department of Statistics, University of California Berkeley, Evans Hall, 2594 Hearst Ave., Berkeley, CA 94720, USA. e-mail: mmahoney@stat.berkeley.edu.} \and
        Shenghao~Yang\thanks{S.~Yang is with the School of Computer Science, University of Waterloo, 200 University Avenue West, Waterloo, ON N2L3G1, Canada. e-mail: shenghao.yang@uwaterloo.ca.} 
}

\maketitle

\begin{abstract}

We are interested in parallelizing the Least Angle Regression (LARS) algorithm for fitting linear regression models to high-dimensional data.\
We consider two parallel and communication avoiding versions of the basic LARS algorithm.\
The two algorithms have different asymptotic costs and practical performance. One offers more speedup and the other produces more accurate output. \
The first is bLARS, a block version of LARS algorithm, where we update $b$ columns at each iteration.\
Assuming that the data are row-partitioned, bLARS reduces the number of arithmetic operations, latency, and bandwidth by a factor of $b$.\
The second is Tournament-bLARS (T-bLARS), a tournament version of LARS where processors compete by running several LARS computations in parallel to choose $b$ new columns to be added in the solution.\
Assuming that the data are column-partitioned, T-bLARS reduces latency by a factor of $b$.\
Similarly to LARS, our proposed methods generate a sequence of linear models.\
We present extensive numerical experiments that illustrate speedups up to $4$x compared to LARS without any compromise in solution quality.\



\end{abstract}

\section{Motivation and outline}
Recently there has been large growth in data for many applications in statistics, machine learning and signal processing and this poses the need for powerful computer hardware as well as new algorithms that utilize the new hardware efficiently.\ 
Commercial hardware companies started to construct multicore designs because the performance of single central processing units (CPUs) is stagnating due to heat issues, i.e., ``the Power Wall" problem \cite{PH13}.\
In terms of software and algorithm implementations for processing large-scale data, the increased number of cores might require synchronization among them and this results in data transfer between levels of a memory hierarchy or between CPUs over a network.\ For this reason the total running time of a parallel algorithm depends on the number of arithmetic operations (computational costs) and the cost of data movement (communication costs).\ The communication cost includes the ``bandwidth cost", i.e. the number of bytes, or more abstractly, number of words, sent among cores for synchronization purposes, and the ``latency cost", i.e. the number of messages sent.\ On modern computer architectures, communicating data often takes much longer than performing a floating-point operation and this gap is continuing to increase \cite{SDM10}.\ Therefore, it is especially important to design algorithms that minimize communication in order to attain high performance on modern computer architectures.\

In this paper we will propose two novel parallel and communication avoiding versions of the least angle regression algorithm which is a very popular method for sparse linear regression \cite{lars}.\
A plethora of applications in statistics \cite{lars}, machine learning \cite{NG04} and signal processing/compressed sensing \cite{CRT06} utilize sparse linear models.\ 
To the best of our knowledge there is no study on parallelizing LARS.\ 

\section{Introduction to the problem, existing models and LARS}\label{sec:intro}
Let $A\in\mathbb{R}^{m\times n}$ be a data matrix with $m$ samples and $n$ features.\ We are concerned with the problem of finding a vector $\yy:=A \xx$ that approximates a given vector $\bb\in\mathbb{R}^m$, where vector $\yy$ is a linear combination of a few columns/features of the given data matrix $A$.\ 
This means that we are looking for a coefficients vector $\xx$ that is sparse, i.e., it has few number of non-zeros.\

Over the years, many algorithms/models to solve this problem have been proposed.\ In what follows, we review the ones that to the best of our knowledge are the most important.\ 
There are two main categories of algorithms/models to solve this problem.\ The first category consists of algorithms that progressively select a subset of columns/features based on their absolute correlation 
with the residual vector  $\yy - \bb$.\ In particular, the classic Forward Selection algorithm in Section $8.5$ in \cite{Weis80} selects the first column/feature with the largest absolute correlation 
with the response $\bb$.\ Let us denote the index of the selected column with $i$, the corresponding column with $A_i$ and the corresponding coefficient with $x_i$.\ The next step of the algorithm is to solve a simple linear regression problem 
$$\min \ \frac{1}{2}\|A_ix_i-\bb\|_2^2.$$
By solving this simple regression problem we obtain the value of the optimal coefficient $x_i$.\
The residual $\rr:=A_ix_i-\bb$, which is orthogonal to $A_i$, is now considered the new response vector for the next iteration.\
Finally, we project orthogonally the remaining columns in $A$ to $A_i$.\ Then we have to repeat this process and find a new column/feature.\
After $k$ iterations we will have selected $k$ columns, and we use the $k$ columns to solve smaller ordinary regression problem using the response vector $\bb$.\ According to \cite{lars}, 
in practice the Forward Selection algorithm might be aggressive in terms of selecting features since other columns might be correlated with the selected column $A_i$
that we ignored.\ Another algorithm in this category is the Forward Stagewise algorithm \cite{HTF01,HTTW07}, which in comparison to Forward Selection is much more cautious since it requires much more steps to converge 
to a $k$-sparse model, i.e., $k$ selected columns.\ More precisely, at each iteration of the Forward Stagewise we select the column that is most correlated with the current residual and we increment the corresponding 
coefficient in the vector $\xx$ by a small amount $\pm \epsilon$, where the sign is determined based on the sign of the correlation.\ The small increment of elements in $\xx$ at each iteration is what distinguishes Forward Stagewise 
and Forward Selection.\

The second category of models is optimization based, meaning that we solve a predefined optimization problem to obtain a sparse linear model.\ There are two subclasses of optimization problems in this category,
the first is known as $\ell_1$-regularized linear regression or least absolute shrinkage and selection operator (LASSO) \cite{Tib96}, the second is $\ell_0$-regularized variants.\
Let us first define the $\ell_1$ and $\ell_0$ norms and then we will continue with presenting the optimization problems.\ The $\ell_1$ norm of a vector $\xx$ is defined as $\|\xx\|_1 := \sum_{i=1}^n |x_i|$, while the 
$\ell_0$ norm is defined as $\|\xx\|_0:=\{\mbox{number of non-zero elements in } \xx\}$.\ Equipped with these definitions we define LASSO
\begin{eqnarray}\label{eq:lasso}
\mbox{minimize} & \frac{1}{2}\|A\xx-\bb\|_2^2 \\\nonumber
\mbox{subject to} & \|\xx\|_1\le \lambda,
\end{eqnarray}
where $\lambda$ is a model parameter.\ LASSO is a convex optimization problem and can be solved in polynomial time, we discuss several serial and parallel algorithms later in this paper.\
The LASSO optimization problem is likely to have a set of sparse optimal solutions due to the sparsity inducing $\ell_1$-ball constraint.\ For details we refer the reader to \cite{Tib96}.\ 
A non-convex alternative of LASSO, but with a direct constraint on the sparsity of $\xx$ is the $\ell_0$-regularized linear regression problem 
\begin{eqnarray}\label{eq:l0lasso}
\mbox{minimize} & \frac{1}{2}\|A\xx-\bb\|_2^2 \\\nonumber
\mbox{subject to} & \|\xx\|_0 \le \tau,
\end{eqnarray}
where $\tau$ is a model parameter that bounds the number of non-zeros in $\xx$.\ This is an NP-hard problem, however, we can find local solutions by variants of gradient descent, which we discuss later in this paper.\

An important difference between the two approaches, i.e., Forward Selection or Stagewise vs LASSO, is that with the former one obtains a sequence of solutions $\xx_k$ with increasing number of non-zeros,
while the latter we obtain a solution path $\xx(\lambda)$.\ There is a question regarding how those two solution paths defer in terms of the selected features.\ 
The LARS algorithm is an algorithmic framework that unifies those two approaches.\
In particular, the LARS algorithm has been motivated by the Forward Selection and Stagewise algorithms, therefore in terms of steps it is similar to those as we will see later, 
but it is also proved in Theorem $1$ in \cite{lars} that a certain version of LARS produces a sequence of solutions $\xx_k$ that is equivalent to the solution path $\xx(\lambda)$.\
Let us now summarize the steps of the LARS algorithm.\ This algorithm is discussed in detail in Section \ref{sec:lars}.\
Similarly to Forward algorithms, at the first iteration of LARS we initialize the algorithm by selecting the column with the largest absolute correlation with vector $\bb$.\ The next step is to update vector $\yy$.\ Instead of solving a simple regression problem
like in Forward Selection (which is an aggressive strategy) or making $\epsilon$ updates to $\xx$ (which is too cautious), we define a vector $\uu$ that is equiangular with all previous chosen columns and then
we update $\yy:= \yy + \uu \gamma$.\ The step-size $\gamma\in\mathbb{R}$ is set such that the new column to be added in the next iteration has the same correlation with the new residual vector as with all other selected columns so far.\
This process might sound complicated at first but we will revisit the linear algebra behind these decisions in Section \ref{sec:lars}.\

\section{Our contributions}\label{sec:contributions}
Although there are numerous parallel optimization algorithms for $\ell_0$- and $\ell_1$-regularized regression, we are not aware of any parallel and communication avoiding versions for LARS.\ To the best of our knowledge the proposed algorithms are the first parallel versions of LARS that are also communication avoiding.\ 
Let us briefly describe the proposed algorithms and the most significant ideas that had to be developed to establish them.\

The first method is a block version of LARS which is described in Section \ref{sec:blars}.\
Instead of adding one feature at each iteration in the solution set we add $b$ features at a time.\
By blocking operations and by partitioning the data per row we are able to show that we decrease the arithmetic, latency and bandwidth costs by a factor of $b$.\ 
Extensive numerical experiments in Section \ref{sec:experiments} illustrate 
significant speedups for block LARS without compromising too much of the quality of the output compared to LARS.\ In the same section we study empirically the trade-off between the size of $b$ and the quality of the output compared to LARS.\

Careful modification of the linear algebra had to be performed in order to successfully generalize LARS to the block case and also guarantee that all steps of the algorithm are well-defined.\
More precisely, LARS has two important properties that we had to relax.\ The first is that all chosen columns at each iteration have the same absolute correlation with the residual and also they are maximally correlated.\ The second property is that the direction $\uu$ is equiangular and also has maximal correlation with the chosen columns.\ Block LARS maintains the property that the chosen columns at each iteration are maximally correlated but they are not equal, meaning that there is no column that has not been selected with larger absolute correlation with the residual than the selected ones.\ Block LARS also relaxes the second property in the sense that $\uu$ is not equiangular with all chosen columns but it is maximally correlated, i.e., there is no column that has not been selected with larger correlation than the selected ones.\ We show that block LARS at each iteration reduces the correlations for all selected columns similarly to LARS.\ Finally, if we set $b=1$ then block LARS reduces to LARS.\

The second method is a tournament block LARS method.\ In this method the data are partitioned per column and distributed to processors.\ Then each processor calls a modified version of the LARS algorithm on its local data.\
Each processor can run the modified LARS algorithm for $b$ iterations so that $b$ columns are chosen at termination of the local call to LARS.\
Using a generalized tree-reduction operation each processor/node sends its chosen columns to the parent node (starting from the bottom of the tree).\ The parent node calls again the modified LARS algorithm by utilizing 
only the columns that have been sent from the child nodes.\ This process repeats until we reach the root node where the final output is used to update the current vector $\yy$ and current set of selected columns.\
By partitioning the data per column (as opposed to per row for block LARS) and using the generalized tree-reduction we allow the nodes to work in parallel in local data and this way we reduce latency by a factor of $b$.\ 
Many of the properties of the LARS algorithm are not satisfied at a global level but some of them are maintained during the local calls to LARS.\ We discuss details in Section \ref{sec:tourlars}.\
In Section \ref{sec:experiments} we show that tournament block LARS can be faster than the original LARS without compromising the quality of the output.\ Similarly to block LARS we study the tradeoff between speed and quality of output
as we vary parameter $b$ and the number of processors.\ 


\section{Literature review for parallel models and methods}\label{sec:literature}

The dependence of the running time of parallel methods on communication requirements gave a totally new perspective on how to efficiently parallelize existing algorithms.\
Communication-avoiding algorithms became a very popular subject of study and it has been demonstrated that such algorithms exhibit large speedups on modern, distributed- and shared-memory parallel architectures through careful algorithmic modifications \cite{Ballard14}.\ Many iterative methods for linear systems and matrix decomposition algorithms have been re-organized to avoid communication and this has led to significant performance improvements over existing state-of-the-art libraries \cite{Ballard14, Ballard13, Carson15, Hoemmen10, Solomonik14, williams14}.\ 

The origins of communication-avoiding algorithms lie in the $s$-step conjugate gradients method \cite{rosendale83} by Van Rosendale's and in the work of Chronopoulos on parallel iterative methods for linear systems \cite{Chrono86}.\ 
More precisely, Chronopoulos and Gear developed $s$-step methods for symmetric linear systems \cite{chronopoulos89a,chronopoulos89b}, Chronopoulos and Swanson developed $s$-step methods for unsymmetric linear systems \cite{chronopoulos96} and Kim and Chronopoulos developed $s$-step non-symmetric Lanczos method \cite{kim92}.\ Furthermore, Demmel, Hoemmen, Mohiyuddin, and others \cite{demmel07, Hoemmen10, mohiyuddin12, mohiyuddin09} introduced the matrix powers kernel optimization which reduces the communication cost of the $s$ Krylov basis vector computations by a factor $O(s)$ for well-partitioned matrices.\ 
Finally, Carson, Demmel, Hoemmen developed communication-avoiding Krylov subspace methods \cite{Carson15, demmel07, Hoemmen10} by combining the matrix powers kernel and $s$-step methods.\

The above results are mainly focused on iterative methods for least-squares and linear systems.\ Our focus on this paper is sparse linear regression where we require the coefficients of the model to be sparse.\
As is mentioned in Section \ref{sec:intro} there are two categories of methods that can solve this problem efficiently.\ The first is LARS-type algorithms.\ 
To the best of our knowledge there are no studies on parallelizing LARS.\ However, we will see in Section \ref{sec:blars} that the computational bottleneck for LARS is computing matrix-vector products.\
Therefore, a straightforward approach for parallelizing LARS is to make use of parallel matrix-vector products.\ There are numerous works on parallelizing matrix-vector product calculations \cite{Q04}.\
In our experiments in Section \ref{sec:experiments} we do compare the two proposed methods with a LARS implementation that uses parallel matrix-vector products.\ Similarly, the proposed block LARS algorithm in Section \ref{sec:blars}
relies on matrix-matrix products which can also be efficiently parallelized \cite{Q04}.\ The proposed tournament block LARS algorithm divides the problem into smaller problems that are solved in parallel and then we aggregate the results by allowing processors to compete.\ This strategy is similar to \cite{DGHL12} for parallel QR and LU algorithms, where pivoting is performed in parallel by using a generalized tree reduction operation.\ Although we also use a generalized tree-reduction operation, at each leaf of the tree we perform a LARS operation and not a pivoting operation.\ Additionally, we modify a crucial part of the LARS algorithm, i.e., the calculation of the step-size, to guarantee that all steps are well-defined.\
Details are discussed in Section \ref{sec:tourlars}.\

Recently, there have been numerous works regarding parallel optimization algorithms.\
$\ell_1$-regularization problems often appear in statistics \cite{lars}, machine learning \cite{NG04} and signal processing/compressed sensing \cite{CRT06} 
where there is a vast amount of data available, i.e., matrix $A$ has millions if not billions of samples and features.\ 
Large scale problems are the main reason for the resurgence in methods with computationally inexpensive iterations.\ Many modern first-order methods meet the previous goal.\ For instance, for $\ell_1$-regularized least-squares problems coordinate descent methods can have up to $n$ times less computational complexity per iteration than methods which use full gradient steps while at the same time it achieves very fast progress to optimality \cite{nesterov12, richtarik14, Wright15}.\ However, it is shown in \cite{DFDM18} that the running time for such methods is often dominated by communication cost which increases with the number of processors.\
In the same work \cite{DFDM18} the authors show how to avoid communication for an $s$-step accelerated proximal block coordinate descent and demonstrate up to $5$x speedup compared to parallelized alternatives.\
Moreover, there are parallel accelerated and proximal coordinate descent methods \cite{FR15} that do not use the $s$-step technique but allow coordinate updates to happen without synchronization.\
For example, HOGWILD! \cite{recht11} is a lock-free approach to stochastic gradient descent (SGD) where each processor selects a data point, computes a gradient using its data point and updates the solution without synchronization.\ 
Finally, there are some frameworks and algorithms that attempt to reduce the communication bottleneck by reducing the number of iterations.\ For example, the CoCoA framework \cite{cocoa} reduces communication by performing coordinate descent on locally stored data points on each processor and intermittently communicating by summing or averaging the local solutions.\ 
Regarding $\ell_0$-regularization there are not many works in terms of parallel methods, a notable work is that of Needell and Woolf \cite{DW17}.\ In this paper the authors suggest an asynchronous parallel and stochastic greedy algorithms, where multiple processors asynchronously update a vector in shared memory containing information on the estimated coefficients vector $\xx$.\ Finally, one could also easily parallelize gradient-based methods for $\ell_1$ and $\ell_0$ regularization by parallelizing the computation of the gradients which relies in matrix-vector products.\

Note that parallel optimization based methods aim in solving a single instance of $\ell_1$ or $\ell_0$ regularized least-squares, ie., they produce a single sparse linear model.\ In this paper we are interested in algorithms that produce a sequence of sparse linear models.\ 

\section{Preliminaries and Assumptions}\label{sec:prem}

\subsection{Preliminaries}
Capital letters denote matrices, lower case bold letters denote vectors, lower case letters denote scalars and hallow letters denote sets.\ 
We denote with $\mathbf{0}_n$ a vector of zeros of length $n$.\ Subscript $k$ denotes the $k$th iteration of the algorithm.\ The set of positive integers is denoted by $\mathbb{Z}_+$.\
We use $[\cdot]_{set}$, to denote a function with a vector as an input that returns 
a subvector which corresponds to the indices in the subscript set.\ $A^T$ denotes the transpose of a matrix.\ 
We denote with $A_{set}$ the concatenation of columns of matrix $A$ with indices in the subscript set.\
We denote the complement of a set by using the superscript $c$.\
We use the function $sign(\cdot)$ to denote the sign function which is applied component-wise if the input is a vector.\ We use the convention that $sign(0) = 0$.\ 
We use $\|\cdot\|_\infty$ to denote the infinity norm, i.e., maximum absolute component of the input, and define $\|\cdot\|_{\infty,k}$ to be the sum of $k$ largest absolute components of the input.\
We define $abs(\cdot)$ as the absolute function which is often applied component-wise.\ We define the function $\max^b(\cdot)$ and $\argmax^b(\cdot)$ as the $b$th maximum of the input vector and the indices of the $b$ largest components of the input vector, respectively.\ If the input vector has less than $b$ components then the latter functions overwrite $b$ to be the length of the input vector.\ We define $\min^b(\cdot)$ and $\argmin^b(\cdot)$ similarly.\ The function $\mbox{min}^+ (\cdot)$ returns the minimum positive value.\ The symbol $\emptyset$ denotes the empty set.\ We denote the simple multiplication of two scalars $a$ and $b$ by $a\cdot b$.\ By $\log $ we denote the logarithm with base $2$.\

\subsection{Assumptions}
For simplicity, we assume that the columns of matrix $A$ have unit $\ell_2$ norm, and that matrix $A$ is full-rank.\ For bLARS, we also assume that every $b$ columns are linearly independent.\ 
However, minor modifications to the algorithms can be done to bypass these assumptions.\
We assume that the communication cost includes the ``bandwidth cost,'' i.e., the number of words, sent among cores for synchronization purposes, and the ``latency cost,'' i.e., the number of messages sent.\ 

\section{Least angle regression} \label{sec:lars}
In this section we review the LARS algorithm.\ 
LARS is shown in Algorithm \ref{algo:lars}.\
The termination criterion in Step $2$ of Algorithm \ref{algo:lars} is arbitrary, one can choose other criteria such as a lower bound on the maximum absolute correlation $\|\cc_k\|_\infty$, see \cite{lars}.\
Let us explain the first iteration of the algorithm.\ 
Let us assume that at the $0$th iteration we have response $\yy_0$, residual vector $\rr_0=\bb - \yy_0$, correlation vector $\cc_0 := A^T\rr_0$ and maximum absolute correlation $c_0:=\max |\cc_0|$.\ 
The algorithm starts by choosing all columns that have maximum absolute correlation
\begin{equation}\label{eq:Ak}
\mathbb{I}_0 := \{ i\in[n] \ | \ |[\cc_0]_i| = c_0 \}.
\end{equation} 
The next decision step is how to set $\mathbb{I}_1$ and $\yy_{1}$ using $\mathbb{I}_0$ and $\yy_0$.\ We will define the update as $\yy_{1}:=\yy_0 + \uu_0 \gamma_0$.\ This implies that we will have to define the vector $\uu_0$ and the step-size $\gamma_0$.\ 
Let us start with the definition of $\uu_0$.\ LARS defines $\uu_0$ as a unit-length vector that is equiangular with signed columns in matrix $A$ with index in $\mathbb{I}_0$.\ 
It is easy to see that $\uu_0:= A_{\mathbb{I}_0} (A_{\mathbb{I}_0}^TA_{\mathbb{I}_0})^{-1}sign([\cc_0]_{\mathbb{I}_0}) c_0 h_0$, where $h_0:=\|A_{\mathbb{I}_0} (A_{\mathbb{I}_0}^TA_{\mathbb{I}_0})^{-1}sign([\cc_0]_{\mathbb{I}_0})c_0\|_2^{-1}$, satisfies the requirements.\
This means that $A_{\mathbb{I}_0}^T \uu_0 = sign([\cc_0]_{\mathbb{I}_0}) c_0h_0$, which in turn implies that subject to sign changes and because the columns of $A_{\mathbb{I}_0}$ and $\uu_0$ are unit-length then $\uu_0$ is equiangular with all columns in $\mathbb{I}_0$, with cosine $\pm c_0 h_0$.\ 
To define $\gamma_0$ and to update $\mathbb{I}_1$ based on $\gamma_0$ we will need first to understand how the update rule $\yy_{1}:=\yy_0 + \uu_0 \gamma_0$ affects the correlation vector $\cc_1$ as a function of $\gamma_0$.\ For this we will make use of the auxiliary vector $\aa_0 := A^T \uu_0$ and we will use a different step-size $\gamma_j$ for each element $j$.\
In particular, we have that $[\cc_{1}]_j(\gamma_j) =  A_j^T(\bb - \yy_{0} - \uu_{0} \gamma_j) = [\cc_{0}]_j - [\aa_{0}]_j \gamma_j \ \forall j\in\mathbb{I}_0^c$
and
\begin{align}\label{eq:dec_cor}
[\cc_{1}]_j(\gamma_j)&  = sign([\cc_0]_j) (1 - \gamma_j h_{0}) c_0\ \forall j\in\mathbb{I}_0. 
\end{align}
Equation \eqref{eq:dec_cor} uses $[\aa_0]_{\mathbb{I}_0}=A_{\mathbb{I}_0}^T\uu_0= sign([\cc_0]_{\mathbb{I}_0}) c_0h_0$ and that vector $[\cc_0]_{\mathbb{I}_0}$ has components of magnitude equal to $c_0$ since it satisfies the definition in \eqref{eq:Ak}.\ Notice that if $\gamma_j = 1/h_0$ then $[\cc_{1}]_j(\gamma_j) = 0$ $\forall j\in\mathbb{I}_0$,
which means that the least-squares problem is minimized with respect to the chosen columns in $\mathbb{I}_0$. Although tempting, this is not the goal of LARS since this is an aggressive strategy similar to Forward Selection.\
As we increase $\gamma_j$ from $0$ to $1/h_0$ the absolute correlations in $\mathbb{I}_0$ are decreased \emph{identically}, see \eqref{eq:dec_cor}.\
This is because the absolute correlations for the columns in $\mathbb{I}_0$ are equal. However, the absolute correlations in $\mathbb{I}_0^c$ might increase or decrease.\
LARS' goal is to find a column in $\mathbb{I}_0^c$ whose absolute correlation becomes equal to the maximum absolute correlation as we increase $\gamma_0$.\
To find such a column we need to find $\gamma_j$ for each $j\in\mathbb{I}_0^c$ such that 
\begin{equation}\label{eq:stepsz_eq}
c_0 (1 - \gamma_j h_{0}) = |[\cc_{0}]_j - \gamma_j [\aa_{0}]_j|.
\end{equation}
Such $\gamma_j$ will guarantee that column $j \in \mathbb{I}_0^c$ has the same absolute correlation as the columns with index in $\mathbb{I}_0$.\
It remains to check if \eqref{eq:stepsz_eq} has a solution.\ It has two solutions, out of which we keep the minimum positive one
$$
\gamma_j := \mbox{min}^+ \left(\frac{c_0 - [\cc_0]_j}{c_0h_0 - [\aa_0]_j},\frac{c_0 + [\cc_0]_j}{c_0h_0 + [\aa_0]_j}\right).
$$
Out of all $\gamma_j$ where $j\in\mathbb{I}_0^c$ we choose the one with the minimum value $\gamma_0:= \min_{j\in\mathbb{I}_0^c} \gamma_j$.\
Note that the minimum step-size $\gamma_0$ corresponds to the column(s) in $\mathbb{I}_0^c$ that will be the first to have the same \emph{maximal} absolute correlation as the columns in $\mathbb{I}_0$.\
Then LARS updates the set of selected columns as $\mathbb{I}_1 := \mathbb{I}_0 \cup \{\argmin_{j\in\mathbb{I}_0^c} \gamma_j\}$.\
The chosen column is the column with the least-angle which is where LARS gets its name from.\
Finally, having the step-size $\gamma_0$ we update the response $\yy_{1} := \yy_0 + \gamma_0 \uu_0$.\


It is easy to show that our claims above hold for any iteration $k$.\ Therefore, it is easy to show that LARS guarantees that $\mathbb{A}_k \subset \mathbb{A}_{k+1}$ and $|\mathbb{A}_k| = |\mathbb{A}_{k+1}| + 1$ $\forall k$.\
Moreover, LARS decreases the maximum absolute correlation $c_k$ until it finally is equal to zero for $k=\min(m,n)$.\ Furthermore, the columns in $\mathbb{A}_k$ have maximum absolute correlations $\forall k$.\ Therefore using \eqref{eq:dec_cor} we see that LARS decreases $\|\cc_k\|_\infty$ at each iteration.
Furthermore, note that LARS also decreases $\|\cc_k\|_{\infty,k} := \mbox{ sum of $k$ largest absolute components}$; as we will see later this is a property that bLARS generalizes but for the $k\cdot b$ largest components.\


\begin{algorithm}
\caption{LARS}
\begin{algorithmic}[1]
\STATE Initialize $k:=0$, $\yy_k:=\mathbf{0}_n$, $\rr_k:=\bb$, $\cc_k := A^T\rr_k$, $i:= \argmax |\cc_k|$, $c_k:= \max |\cc_k|$, $\mathbb{I}_k := \{i\}$, $t \le \min(m,n)$ 
\WHILE {$|\mathbb{I}_k| \le t$}
\STATE $\uu_k:= A_{\mathbb{I}_k} (A_{\mathbb{I}_k}^TA_{\mathbb{I}_k})^{-1}sign([\cc_k]_{\mathbb{I}_k}) h_k c_k$, where $h_k:=\|A_{\mathbb{I}_k} (A_{\mathbb{I}_k}^TA_{\mathbb{I}_k})^{-1}sign([\cc_k]_{\mathbb{I}_k})c_k\|_2^{-1}$
\STATE $\gamma_j := \min^+ \left(\frac{c_k - [\cc_k]_j}{c_kh_k - [\aa_k]_j}, \frac{c_k + [\cc_k]_j}{c_kh_k + [\aa_k]_j}\right)$ $\forall j\in \mathbb{I}_k^c$, where $\aa_k := A^T \uu_k$
\STATE $\gamma_k:= \min_{j\in \mathbb{I}_k^c} \gamma_j$, \ $i:= \argmin_{j\in \mathbb{I}_k^c} \gamma_j$, \ $\mathbb{I}_{k+1} := \mathbb{I}_{k}\cup\{i\}$
\STATE $\yy_{k+1} := \yy_k + \uu_k \gamma_k$
\STATE $\cc_{k+1} := A^T\rr_{k+1}$, where $\rr_{k+1} := \bb - \yy_{k+1}$
\STATE $c_k:= \max |\cc_k|$
\STATE $k:= k + 1$
\ENDWHILE
\STATE Return $\mathbb{I}_k$, $\yy_k$
\end{algorithmic}
\label{algo:lars}
\end{algorithm}

\section{Parallel block Least Angle Regression}\label{sec:blars}
In this section, we describe one iteration of bLARS (without going into any details about parallelism), and then we explain how we can parallelize bLARS.\

Let us assume that at the $0$th iteration of bLARS we have response $\yy_0$, residual vector $\rr_0=\bb - \yy_0$, correlation vector $\cc_0 := A^T\rr_0$ and the $b$th maximum correlation $c_0:=\max^{b} |\cc_0|$.\ 
The algorithm chooses all columns that have larger or equal absolute correlation than the maximum $b$th absolute correlation $\mathbb{I}_0 = \{i\in[n] \ | \ |[\cc_0]_i| \ge c_0 \}$.\
Similarly to LARS, we define the update as $\yy_{1}:=\yy_0 + \uu_0 \gamma_0$, but the decision rules for selecting $\uu_0$, $\gamma_0$ and updating $\mathbb{I}_0$ and $\yy_0$ are different.\ 
bLARS defines $\uu_0$ as $\uu_0:= A_{\mathbb{I}_0} (A_{\mathbb{I}_0}^TA_{\mathbb{I}_0})^{-1}[\cc_0]_{\mathbb{I}_0} h_0$ and $h_0:=\|A_{\mathbb{I}_0} (A_{\mathbb{I}_0}^TA_{\mathbb{I}_0})^{-1}[\cc_0]_{\mathbb{I}_0}\|_2^{-1}$.\
This means that $\uu_0$ is a unit-length vector that satisfies
$A_{\mathbb{I}_0}^T\uu_0= [\cc_0]_{\mathbb{I}_0} h_0$, instead of $A_{\mathbb{I}_0}^T \uu_0 = \mbox{sign}([\cc_0]_{\mathbb{I}_0}) c_0h_0$ for LARS.\
Note that $\uu_0$ is not guaranteed to be equiangular to the chosen columns in $\mathbb{I}_0$.\ This is because $[\cc_0]_{\mathbb{I}_0}$ is not guaranteed to have components with equal value.\
On the contrary, LARS guarantees that all components of $[\cc_0]_{\mathbb{I}_0}$ are equal to the maximum absolute correlation.\
However, bLARS still guarantees that there is no column that has not been selected with absolute correlation larger than the $b$th maximum absolute correlation.\ 
Similarly to LARS, we will make use of the auxiliary vector $\aa_0 := A^T \uu_0$, but we will use different step-sizes $\gamma_j$ for each element $j$.\
In particular, we have that $[\cc_{1}]_j(\gamma_j) =  A_j^T(\bb - \yy_{0} -  \uu_{0} \gamma_j) = [\cc_{0}]_j - [\aa_{0}]_j \gamma_j\ \forall j\in\mathbb{I}_0^c$, where $\mathbb{I}_0^c$ is the complement of $\mathbb{I}_0$,
and
\begin{align}\label{eq:dec_cor_blars}
[\cc_{1}]_j(\gamma_j)& = [\cc_{0}]_j (1 - \gamma_j h_{0}) \ \forall j\in\mathbb{I}_0.
\end{align}
The last equality uses $[\aa_0]_{\mathbb{I}_0}=A_{\mathbb{I}_0}^T\uu_0= [\cc_0]_{\mathbb{I}_0} h_0$.\ 
This is different from LARS which uses $[\aa_0]_{\mathbb{I}_0}=\mbox{sign}([\cc_0]_{\mathbb{I}_0}) c_0h_0$.\ 
This means that as we increase $\gamma_j$ LARS decreases 
the absolute correlations identically, but bLARS decreases the absolute correlations with the same rate but not identically.\ However, bLARS still guarantees that if $\gamma_j = 1/h_0$ then $[\cc_{1}]_j(\gamma_j) = 0$ $\forall j\in\mathbb{I}_0$,
which means that the least-squares problem is minimized with respect to the chosen columns in $\mathbb{I}_0$.\ 
Furthermore, bLARS still guarantees that as we increase $\gamma_j$ from $0$ to $1/h_0$ the absolute correlations in $\mathbb{I}_0$ are decreased, see \eqref{eq:dec_cor_blars},
but the absolute correlations in $\mathbb{I}_0^c$ might increase or decrease.\
bLARS goal is to find $b$ columns in $\mathbb{I}_0^c$ for which their absolute correlations become larger or equal to the minimum absolute correlation of columns in $\mathbb{I}_0$ as we increase $\gamma_0$.\
To find such a column we need to find $\gamma_j$ for each $j\in\mathbb{I}_0^c$ such that 
\begin{equation}\label{eq:stepsz_eq_blars}
c_0 (1 - \gamma_j h_{0}) = |[\cc_{0}]_j - \gamma_j [\aa_{0}]_j|.
\end{equation}
Using the definition of $c_0$, such $\gamma_j$ will guarantee that column $j \in \mathbb{I}_0^c$ has the same absolute correlation as the column with index $i\in\mathbb{I}_0$ that satisfies $i=\argmax^{b} |\cc_0|$.\
Equation \eqref{eq:stepsz_eq_blars} has two solutions, we keep the minimum positive solution
$$
\gamma_j := \mbox{min}^+ \left(\frac{c_0 - [\cc_0]_j}{c_0h_0 - [\aa_0]_j},\frac{c_0 + [\cc_0]_j}{c_0h_0 + [\aa_0]_j}\right).
$$
Out of all $\gamma_j$ where $j\in\mathbb{I}_0^c$ we choose the one with the minimum $b$th value $\gamma_0:= \mbox{min}^b_{j\in\mathbb{I}^c} \gamma_j$.\
Note that the $b$th minimum step-size $\gamma_0$ corresponds to the column(s) in $\mathbb{I}_0^c$ that will be the $b$th to have the same absolute correlation with the column in $\mathbb{I}_0$ with the minimum absolute correlation.\ 
Then bLARS updates $\mathbb{I}_1 := \mathbb{I}_0 \cup \{b \mbox{ columns with } \gamma_j \ge \gamma_0\}$.\
Note that bLARS decreases $\|\cc_k\|_{\infty,k \cdot b} := \mbox{sum of $k\cdot b$ largest absolute components}$, compared to LARS which decreases $\mbox{sum of $k$ largest absolute components}$.\
It is easy to see that by setting $b=1$ then bLARS is equivalent to LARS.\

The parallel bLARS algorithm is shown in Algorithm \ref{algo:blars}.\ This algorithm is presented in great detail since this demonstrates our implementation.\   
We assume that the data matrix $A$ and any vector/set of length/cardinality $m$ are partitioned across processors, i.e., each processor holds $m/P$ components, where $P$ is the number of processors and we assume for simplicity that $m/P$ is an integer.\ More complicated two dimensional partitions could be used \cite{PHP03,C69} and may potentially improve communication cost, but we use row partition for simplicity and leave more sophisticated partitioning methods for future work.\
The main computational kernels of the algorithm are matrix-matrix and matrix-vector products, which we can parallelize efficiently 
using Message Passing Protocol (MPI) collective routines for reduction \cite{TRG2005}.\ 
We also make use of collective routines for broadcasting data \cite{TRG2005}.\
In our numerical experiments in Section \ref{sec:experiments}, we use parallel bLARS with $b=1$ as parallel LARS.\ 

\begin{algorithm}[h]
\caption{Parallel bLARS for row-partitioned data}
\begin{algorithmic}[1]
\STATE Initialize $b\in\mathbb{Z}_+$,  $t \le \min(m,n)\in\mathbb{Z}_+$, $k:=0$, $\yy_k:=\mathbf{0}_n$, $\rr_k:=\bb$ in parallel without synchronization.
\STATE Compute $\cc_k := A^T\rr_k$ in parallel using reduction.
\STATE $c_k:= \max^b |\cc_k|$, $\mathbb{I}_k :=\{i\in[n] \ | \ |[\cc_k]_i| \ge c_k \}.$
\STATE Compute $G_k := A_{\mathbb{I}_k}^TA_{\mathbb{I}_k}$ in parallel using a reduction.
\STATE Compute $L_k$, the Cholesky factor of $G_k$
\WHILE {$|\mathbb{I}_k| < t$}
\STATE $\bss_{k} := [\cc_{k}]_{\mathbb{I}_{k}}$, $\qq_{k} := (L_{k}L_{k}^T)^{-1}\bss_{k}$
\STATE $h_{k} := (\bss_{k}^T\qq_{k})^{-1/2}$, $\ww_{k} := \qq_{k} h_{k}$
\STATE The master processor broadcasts $\ww_{k}$.
\STATE Compute $\uu_{k} := A_{\mathbb{I}_{k}} \ww_{k}$ in parallel, no communication is required.
\STATE Compute $\aa_{k} := A^T \uu_{k}$ in parallel using a reduction.
\STATE $\gamma_j := \min^+ \left(\frac{c_k - [\cc_k]_j}{c_kh_k - [\aa_k]_j}, \frac{c_k + [\cc_k]_j}{c_kh_k + [\aa_k]_j}\right)$ $\forall j\in \mathbb{I}_k^c$
\STATE $\gamma_k:= \min_{j\in \mathbb{I}_k^c}^b \gamma_j$, 
\STATE $\mathbb{B}:= \argmin_{j\in \mathbb{I}_k^c}^b \gamma_j$ (note this returns $b$ indices)
\STATE $\mathbb{I}_{k+1} := \mathbb{I}_{k}\cup \mathbb{B}$
\STATE The master processor broadcasts $\gamma_k$ to all processors.
\STATE Compute $\yy_{k+1} := \yy_k + \uu_k \gamma_k$ in parallel, no communication is required.
\STATE $[\cc_{k+1}]_j := [\cc_{k}]_j(1-\gamma_k h_{k})$ $\forall j \in \mathbb{I}_{k}$, and $[\cc_{k+1}]_j = [\cc_{k}]_j - \gamma_k [\aa_{k}]_j$ $\forall j \in \mathbb{I}_{k}^c$
\STATE $c_{k+1}:= c_{k}(1-\gamma_k h_{k})$
\STATE Compute $A_{\mathbb{I}_{k}}^TA_{\mathbb{B}}$ and $A_{\mathbb{B}}^T A_{\mathbb{B}}$ in parallel using a reduction.
\STATE $H_{k+1}:= L_{k}^{-1}A_{\mathbb{I}_{k}}^TA_{\mathbb{B}}$
\STATE Solve $\Omega_{k+1}^T\Omega_{k+1} = A_{\mathbb{B}}^T A_{\mathbb{B}} - H_{k+1}^TH_{k+1}$ subject to $\Omega_{k+1}$ being a lower triangular matrix.
\STATE $L_{k+1}:= \begin{bmatrix}
	 	    L_{k} & \mathbf{0}_{k,b} \\
 		    H_{k+1} & \Omega_{k+1}
	    \end{bmatrix}$
\STATE $k:= k + 1$
\ENDWHILE
\STATE Return $\mathbb{I}_k$, $\yy_k$
\end{algorithmic}
\label{algo:blars}
\end{algorithm}

\subsection{Asymptotic costs for parallel bLARS and LARS}
 
In what follows we examine the asymptotic costs of each step of parallel bLARS in Algorithm \ref{algo:blars}.\ 
The asymptotic costs of parallel LARS are obtained by setting $b=1$.\ We also comment when a step is executed only by the master processor, by all processors independently or in parallel with synchronization.\
We model the running time of an algorithm by considering both arithmetic and communication costs.\ In particular, we model the running time of an algorithm as a sum of three terms
as
$$
\gamma F + \alpha L + \beta W,
$$
where $\gamma$, $\alpha$ and $\beta$ are hardware parameters for time per arithmetic operation, time per message sent and time per word moved, respectively.\
$F$, $L$ and $W$ are algorithm parameters for number of arithmetic operations to be executed, number of messages to be sent and number of words to be moved, respectively.\ 
We choose the $\alpha$-$\beta$ model to measure communication of algorithms for simplicity.\ More refined models exists like the LogP \cite{CKPSSSSE93} and LogGP \cite{AISS97} models.\

We assume that matrix $A$ is a dense matrix.\
Step $1$ requires $O(m/P)$ operations for initialization of $\yy_0$ and $\rr_0$ in parallel with no communication.\ 
Step $2$ requires computing $\cc_k$ which is equal to $A^T\rr_k$.\ This operation can be performed in parallel with synchronization in $O(mn/P)$ operations, $n \log P$ words
and $\log P$ messages, using a binary tree reduction algorithm in \cite{TRG2005}.\ The result of Step $2$ is reduced to the master processor.\ Step $3$ is performed by the master processor and it costs $O(n)$ operations using Introspective Selection~\cite{Musser1997}.\
Step $4$ is performed in parallel with synchronization and it requires $O(b^2 m/P)$ operations, $b^2 \log P$ words and $\log P$ messages using binary tree reduction.\ 
Step $5$ is executed by the master processor and it costs $O(b^3)$ operations.\ 
Step $7$ is executed by the master processor and it costs $O(|\mathbb{I}_k|)$ operations to compute $\bss_{k} := [\cc_{k}]_{\mathbb{I}_{k}}$.\ Since $|\mathbb{I}_k| = b(k+1)$, this requires $O(bk + b)$ operations.\
Moreover, Step $7$ requires an additional $O(b^2(k+1)^2)$ operations to compute $\qq_{k} := (L_{k}L_{k}^T)^{-1}\bss_{k}$, which is also executed by the master processor.\ 
Steps $8$ costs $O(bk + b)$ operations and it is executed by the master processor.\ 
In Step $9$, $\ww_k$ has to be broadcasted to each processor from the master processor and this costs $b(k+1)\log P$ words and $\log P$ messages using a broadcast algorithm from \cite{TRG2005}.\
Step $10$ is computed in parallel without synchronization in $O(b(k+1)m/P)$ operations, i.e., each processor multiplies its own part of the vector $A_{\mathbb{I}_k}$ with $\ww_k$.\ 
Step $11$ is executed in parallel with synchronization and it requires $O(mn/P)$ operations, $n \log P$ words and $\log P$ messages using a reduction.\ The result of Step $11$ is reduced to the master processor.\ 
Step $12$ is executed by the master processor and it requires $O(|\mathbb{I}_k^c|)$ operations, which is upper bounded by $O(n)$ operations 
in worst-case since $|\mathbb{I}_k^c| \le n$.\ 
Steps $13$ and $14$ are executed by the master processor and they require in worst-case $O(n)$ operations using Introspective Selection.\ 
Step $15$ is executed by the master processor and it costs $O(b)$ operations.\
In Step $16$ the step-size $\gamma$ is broadcasted to all processors from the master processor in $\log P$ words and $\log P$ messages.\ 
Step $17$ is executed in parallel without synchronization and it requires $O(m/P)$ operations.\ Steps $18$ and $19$ are executed by the master processor and they require $O(n)$ operations.\
Step $20$ is executed in parallel with synchronization and it requires 
$O(b^2km/P + b^2m/P)$ operations, $O(b^2k\log P + b^2\log P)$ words and $2\log P$ messages.\ The result of Step $20$ is reduced to the master processor.\
Step $21$ is executed by the master processor and it requires $O(b^3k^2)$ operations since $L_{k}$ is a lower triangular matrix.\ 
Step $22$ is executed by the master processor and it requires $O(b^3k + b^2)$ operations.\
Step $23$ is executed by the master processor and it requires $O(b^2k + b^2)$ operations.\
Notice that if we want to obtain $t$ columns using LARS then we need to run the algorithm for $t-1$ iterations.\ 
Therefore, if we want to obtain $t$ columns using bLARS then we need to run the algorithm for $(t-1)/b$ iterations.\
By using this and the above costs for each step we summarize in Table \ref{tab:costs_blars} the asymptotic costs of bLARS and LARS for obtaining a solution with $t$ columns.\
Assuming that $t \gg b$, which means that we want to output many more columns than $b$, then we observe in Table \ref{tab:costs_blars} that by using bLARS we reduce by a factor of $b$ all major 
computational and communication costs compared to LARS.\

\begin{table}
\centering
 \begin{tabular}{||c c c c||} 
 \hline
 Step(s) & Arithmetic operations (F) & Words (W) & Messages (L) \\ [0.5ex] 
 \hline\hline
 $1$ & $\frac{m}{P}$ & - & - \\ 
 $2$ & $\frac{mn}{P}$ & $n\log P$ & $\log P$  \\ 
 $3$ & $n$ & - & - \\ 
 $4$ & $\frac{b^2m}{P}$ & $b^2 \log P$ & $\log P$  \\ 
 $5$-$8$ & $\frac{t^3}{b} + \frac{t^2}{b}$ & - & - \\ 
 $9$ & - & $\frac{t^2}{b}\log P + t \log P$ & $\frac{t}{b}\log P$ \\ 
 $10$ & $\frac{t^2 m}{bP} + \frac{t m}{P}$ & - & - \\ 
 $11$ & $\frac{tmn}{bP}$ & $\frac{tn}{b}\log P$ & $\frac{t}{b}\log P$ \\ 
 $12$-$14$ & $\frac{tn}{b}$ & - & - \\ 
 $15$ & $t$ & - & - \\ 
 $16$ & - & $\frac{t}{b}\log P$ & $\frac{t}{b}\log P$ \\
 $17$-$19$ & $\frac{tm}{bP} + \frac{tn}{b}$ & - & - \\ [0.7ex] 
 $20$ & $\frac{t^2m}{P} + \frac{tbm}{P}$ & $t^2\log P + tb\log P$ & $\frac{t}{b}\log P$ \\ [0.7ex] 
 $21$-$23$ & $ t^3 + t^2 + tb$ & - & - \\ 
 Total (assuming $t\gg b$)& $\frac{tmn}{bP} + \frac{tn}{b} + \frac{t^2m}{P} + t^3$ & $\frac{tn}{b}\log P + t^2 \log P$& $\frac{t}{b}\log P$ \\ [1ex] 
 \hline
\end{tabular}
\caption{Running time costs for parallel bLARS in Big O notation. The running time costs of LARS can be obtained by setting $b=1$. The first column shows the number of step(s) of the algorithm. The second, third and forth columns show the number of operations, the number of words and the number of messages, respectively, that are required by bLARS to output a solution with $t$ columns/features}
\label{tab:costs_blars}
\end{table}

\section{Tournament block Least Angle Regression}\label{sec:tourlars}
In this section we will present tournament block LARS (Tournament-bLARS), a variation of LARS where $b$ columns are selected at each iteration using a generalized reduction on a binary tree. 
Like bLARS, Tournament-bLARS requires a lot of non-trivial modifications in order to maintain some properties of the original algorithm which we discuss in detail below. 
In comparison to parallel LARS and bLARS, for Tournament-bLARS we assume that the data matrix $A$ \emph{column-partitioned}, i.e., each processor holds $n/P$ columns, where $P$ is the number of processors and we assume that $n/P$ is an integer. 
Furthermore, we assume that vectors of length $m$ or $n$ or sets with cardinality at most $m$ or $n$ can be stored locally.

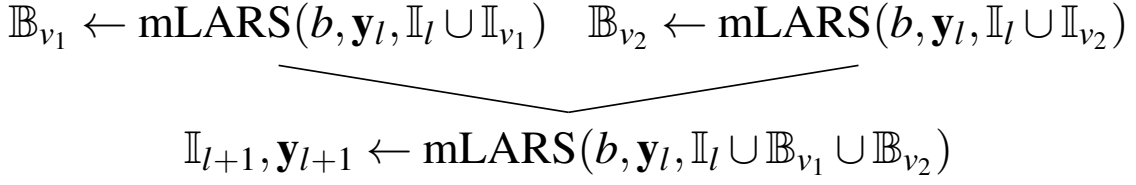
\begin{figure}[!h]
\centering
\resizebox{0.9\columnwidth}{!}{%
\begin{tikzpicture}[grow'=up]
            \Tree [. $\mathbb{I}_{l+1},\mathbf{y}_{l+1}\leftarrow\mbox{mLARS}(b,\mathbf{y}_l,\mathbb{I}_l\cup \mathbb{B}_{v_1}\cup \mathbb{B}_{v_2})$ [.$\mathbb{B}_{v_1}\leftarrow\mbox{mLARS}(b,\mathbf{y}_l,\mathbb{I}_l\cup\mathbb{I}_{v_1})$  ] [.$\mathbb{B}_{v_2}\leftarrow\mbox{mLARS}(b,\mathbf{y}_l,\mathbb{I}_l\cup\mathbb{I}_{v_2})$  ] ]  
\end{tikzpicture}
}
\vspace{-0.4cm}
\caption{Binary tree for one iteration of T-bLARS.\ The nodes at the bottom of the tree communicate columns in $\mathbb{B}_{v_1}$ and $\mathbb{B}_{v_2}$.}
\label{fig:bin_tree}
\end{figure}

Let us now describe one iteration of T-bLARS.\ Let us assume that at the $l$th iteration we have response $\yy_l$ and we have selected columns $\mathbb{I}_l$.\
Furthermore, let us assume that $P=2$, i.e., $2$ processors.\ Each processor gets $n/P$ columns, which we denote with index sets $\mathbb{I}_{v_1}$ and $\mathbb{I}_{v_2}$.\ 
T-bLARS requires running a modified version of LARS (mLARS), which we discuss later, as a reduction on a binary tree.\ For a visual
explanation see Figure \ref{fig:bin_tree}.\ The algorithm starts at the bottom of the tree by calling mLARS for each node in parallel.\ Nodes $v_1$ and $v_2$ return candidate columns with indices in the sets $\mathbb{B}_{v_1}$ and $\mathbb{B}_{v_2}$, respectively.\ 
Columns $\mathbb{B}_{v_1}\cup\mathbb{B}_{v_2}$ are sent to node $v_3$, which is the parent of $v_1$ and $v_2$.\ 
Finally, the node $v_3$ calls mLARS using columns in $\mathbb{I}_l\cup \mathbb{B}_{v_1}\cup \mathbb{B}_{v_2}$ which returns the new response $\yy_{l+1}$ and index set $\mathbb{I}_{l+1}$. Then this process is repeated.
Details are provided in Algorithm \ref{algo:tournamentblars}.\

\paragraph{Modified LARS.}
We mentioned that each node calls a modified version of LARS Algorithm \ref{algo:modlars}. Let us now comment on this algorithm and why LARS needs to be modified in order for Tournament-bLARS to be a well-defined algorithm.
The problem is caused due to the fact that each processor on any level of the binary tree runs mLARS independently of other processors and on data that might not overlap. This may result in violation of a basic rule of LARS, which is that \emph{there is no column that has not been selected with larger absolute correlation than the current known maximum absolute correlation $c_k$}.

Similarly to LARS, mLARS chooses one column at each iteration. 
Each call to mLARS operates on the columns with indices in $\mathbb{I}_\nu\cup \mathbb{I}_l$, where $\nu$ is the index of the node in the binary tree and $\mathbb{I}_l$ is the set of indices of columns that have been selected at the $l$th iteration of Tournament-bLARS.
If $\mathbb{I}_l$ does not include any index with maximum absolute correlation among the indices in $\mathbb{I}_\nu\cup \mathbb{I}_l$, then equation \eqref{eq:stepsz_eq} might not have a non-negative solution. 
This affects the step-size calculation, which for LARS is computed by solving equation \eqref{eq:stepsz_eq} with the constraint that $\gamma\ge 0$. 
To guarantee that a meaningful step-size is calculated at each iteration of mLARS we propose using stepLARS in Procedure~\ref{proc:step_size}. 
Briefly, stepLARS detects violations to the above basic rule of LARS.
If it detects a violation it checks if \eqref{eq:stepsz_eq} still has a non-negative solution and sets $\gamma_k$ appropriately. If it cannot resolve it (equation \ref{eq:stepsz_eq} does not have a non-negative solution) then it sets $\gamma_k=0$. By setting $\gamma_k=0$ we guarantee that the response $\yy_k$ is not updated in current iteration. 
Setting $\gamma_k$ to a positive value would be a ``mistake" since as we show in Step $14$ of stepLARS Procedure~\ref{proc:step_size} this would result in decreasing the current known maximum correlation $c_k$ of mLARS but at the same time it \emph{increases} the absolute correlation of columns that violate the LARS property. This makes violation of the LARS property even larger.

If $\gamma_k=0$ then mLARS at Step $18$ adds the column with the largest absolute correlation that also violates the LARS property in the set of selected columns.
This decision guarantees that a violation will not happen again during the execution of mLARS.
This is because similarly to LARS, mLARS guarantees that once $c_k$ is maximal then it will remain like this for all iterations and this ensures that \eqref{eq:stepsz_eq} always has at least one non-negative solution.
More details are described in mLARS Algorithm \ref{algo:modlars} and Procedure~\ref{proc:step_size}.\

\begin{algorithm}[h]
\caption{T-bLARS}
\begin{algorithmic}[1]
\STATE Initialize $l:=0$, $\yy_l:=\mathbf{0}_n$, $t\in\mathbb{Z}_+$, $b\in\mathbb{Z}_+$, $L_l=0$, where $L_l$ is the Cholesky factor.
\STATE Initialize $\mathbb{I}_l = \emptyset$
\WHILE {$|\mathbb{I}_l| < t$}
\FOR {all levels of the tree from bottom to the root}
\IF {at the bottom of the tree}
\STATE Let $\mathbb{I}_v$ be the columns of node $v$ in the tree.\ For all nodes $v$ in the current level of the binary tree call $\mathbb{B}_{v} \leftarrow \mbox{mLARS}(b,\yy_l,\mathbb{I}_l\cup\mathbb{I}_v,L_l)$.\
\ELSIF {not at root of the binary tree}
\STATE Let $\mathbb{B}_v$ be the columns selected by child nodes of $v$.\
For all nodes $v$ in the current level of the binary tree call 
$\tilde{\mathbb{B}}_{v}  \leftarrow \mbox{mLARS}(b,\yy_l,\mathbb{I}_l\cup\mathbb{B}_v,L_l)$,
where $\tilde{\mathbb{B}}_{v}$ are the selected $b$ columns out of $\mathbb{B}_v$.\ 
\STATE Send columns $\tilde{\mathbb{B}}_{v}$ for each node $v$ to the processor of the parent node of $v$.
\ELSE
\STATE $\yy_{l+1},\mathbb{I}_{l+1},\mathbb{B}_{l+1},L_{l+1}  \leftarrow \mbox{mLARS}(b,\yy_l, \mathbb{I}_l\cup\mathbb{B}_v,L_l)$
\STATE Broadcast selected columns with index in $\mathbb{B}_{l+1}$, $\yy_{l+1}$, and $L_{l+1}$ to all processors.\ 
Note that we only communicate the part of $L_{l+1}$ that gets updated by the root node.\ 
\ENDIF
\ENDFOR
\STATE $l := l +1$
\ENDWHILE
\STATE Return $\mathbb{I}_l$, $\yy_l$
\end{algorithmic}
\label{algo:tournamentblars}
\end{algorithm}

\begin{subroutine}[h]
\caption{Step-size for modified LARS (stepLARS)}
\label{proc:step_size}
\begin{algorithmic}[1]
\STATE Input: $c_k$, $h_k$, $\cc_k$, $\aa_k$ and an index $j$
\IF {$c_k \ge |[\cc_k]_j|$}
\IF {$[\cc_k]_j$ and $[\aa_k]_j$ have the same sign}
\STATE Equation $c_k (1 - \gamma h_k) = |[\cc_k]_j - \gamma [\aa_k]_j|$ has at least one positive solution, we select the minimum positive one $\gamma:= \min^+ \left(\frac{c_k - [\cc_k]_j}{c_kh_k - [\aa_k]_j}, \frac{c_k + [\cc_k]_j}{c_kh_k + [\aa_k]_j}\right)$.
\ELSE
\STATE Equation $c_k (1 - \gamma h_k) = |[\cc_k]_j - \gamma [\aa_k]_j|$ has one positive solution that is $\gamma:= \frac{c_k - |[\cc_k]_j|}{c_kh_k + |[\aa_k]_j|}$.
\ENDIF
\STATE
\IF {$[\cc_k]_j$ and $[\aa_k]_j$ have the same sign and $[\cc_k]_j h_k \le [\aa_k]_j$}
\STATE Equation $c_k (1 - \gamma h_k) = |[\cc_k]_j - \gamma [\aa_k]_j|$ has one positive solution that is $\gamma:= \frac{c_k - |[\cc_k]_j|}{c_kh_k - |[\aa_k]_j|}$.
\ELSIF {$[\cc_k]_j$ and $[\aa_k]_j$ have the same sign and $[\cc_k]_j h_k > [\aa_k]_j$}
\STATE Equation $c_k (1 - \gamma h_k) = |[\cc_k]_j - \gamma [\aa_k]_j|$ does not have a positive solution. But as $\gamma$ increases $c_k (1 - \gamma h_k) $ and $|[\cc_k]_j - \gamma [\aa_k]_j|$ decrease, therefore, we set $\gamma$ to its maximum value
$\gamma := 1/h_k$.
\ELSE
\STATE Equation $c_k (1 - \gamma h_k) = |[\cc_k]_j - \gamma [\aa_k]_j|$ does not have a positive solution.  
In this case, as $\gamma$ increases $|[\cc_k]_j - \gamma [\aa_k]_j|$ increases and $c_k (1 - \gamma h_k)$ decreases.
Therefore, we set $\gamma:=0$, which subject to $\gamma \ge 0$ minimizes the error $|[\cc_k]_j - \gamma [\aa_k]_j| - c_k (1 - \gamma h_k)$.
\ENDIF
\ENDIF
\STATE Return $\gamma$
\end{algorithmic}
\end{subroutine}

\begin{algorithm}[h]
\caption{Modified Least Angle Regression (mLARS)}
\begin{algorithmic}[1]
\STATE Input: number of columns $b\in\mathbb{Z}_+$, response $\tilde{\yy}$, column index sets $\tilde{\mathbb{I}}_0 \cup \tilde{\mathbb{I}}_v$ (third input) and Cholesky factor $\tilde{L}$ (forth input)
\STATE Initialize: $k:=0$, $\mathbb{B}:=\emptyset$, $L_k:= \tilde{L}$, $\mathbb{I}_k:=\tilde{\mathbb{I}}_0$
\STATE $\rr_k:= \bb - \tilde{\yy}$
\STATE $\cc_k := A^T_{\mathbb{I}_k\cup \tilde{\mathbb{I}}_v} \rr_k$
\STATE $c_k:= \max |[\cc_k]_{\mathbb{I}_k}|$. Note that we abuse notation here for $[\cc_k]_{\mathbb{I}_k}$. Since $\cc_k\in\mathbb{R}^{|\mathbb{I}_k\cup \tilde{\mathbb{I}}_v|}$ and by usual convention its components are indexed from $1$ to $|\mathbb{I}_k\cup \tilde{\mathbb{I}}_v|$ which might not overlap with the indices in $\mathbb{I}_k$. We assume that the components of $\cc_k$ are indexed using the indices in $\mathbb{I}_k\cup \tilde{\mathbb{I}}_v$. We use this abuse of notation at other steps of this algorithm because it simplifies notation.
\IF {$\mathbb{I}_k = \emptyset$}
\STATE $c_k:= \max |[\cc_k]|$, $\mathbb{I}_k := \{\argmax |\cc_k|\}$, $L_k = (A_{\mathbb{I}_k}^TA_{\mathbb{I}_k})^{1/2}$. 
\ENDIF
\WHILE {$|\mathbb{I}_k| < |\tilde{\mathbb{I}}_0| + b$}
\STATE $\bss_k := [\cc_k]_{\mathbb{I}_k}$
\STATE $\qq_k := (L_kL_k^T)^{-1}\bss_k$
\STATE $h_k := (\bss_k^T\qq_k)^{-1/2}$
\STATE $\ww_k := \qq_k h_k$
\STATE $\uu_k := A_{\mathbb{I}_k} \ww_k$
\STATE $\aa_k := A^T_{\mathbb{I}_k\cup\tilde{\mathbb{I}}_v} \uu_k$
\STATE $\gamma_j \leftarrow \mbox{stepLARS}(c_k, h_k, \cc_k, \aa_k, j)$ $\forall j\in \tilde{\mathbb{I}}_v\backslash\mathbb{I}_{k}$
\STATE If there are $\gamma_j$ that are equal to zero, set $\gamma_k$ to zero. Otherwise, set $\gamma_k$ to the minimum nonzero $\gamma_j$.
\STATE If there are $\gamma_j$ that are equal to zero, set $i$ to the $j$th column with the largest $|[\cc_k]_j|$. 
Otherwise, set $i$ to the $j$th column with the minimum nonzero $\gamma_j$. 
\STATE $\yy_{k+1} := \yy_k + \uu_k \gamma_k$
\STATE $[\cc_{k+1}]_j := [\cc_{k}]_j(1-\gamma h_{k})$ $\forall j \in \mathbb{I}_{k}$, and $[\cc_{k+1}]_j = [\cc_{k}]_j - \gamma [\aa_{k}]_j$ $\forall j \in \tilde{\mathbb{I}}_v\backslash\mathbb{I}_{k}$
\STATE $\mathbb{I}_{k+1} := \mathbb{I}_{k}\cup\{i\}$, $\mathbb{B} := \mathbb{B}\cup\{i\}$
\STATE $c_{k+1}:= \max |[\cc_{k+1}]_{\mathbb{I}_{k+1}}|$
\STATE Compute $A_{\mathbb{I}_{k}}^TA_{i}$ and $A_{i}^T A_{i}$.
\STATE $\bll_{k+1}:= L_{k}^{-1}A_{\mathbb{I}_{k}}^TA_{i}$
\STATE $\omega_{k+1} := (A_{i}^T A_{i} - \bll_{k+1}^T\bll_{k+1})^{1/2}$
\STATE $L_{k+1}:= \begin{bmatrix}
	 	    L_{k} & \mathbf{0}_{k} \\
 		    \bll_{k+1} & \omega_{k+1}
	    \end{bmatrix}$
\STATE $k:= k + 1$	   
\ENDWHILE
\STATE Return $\yy_k$, $\mathbb{I}_k$, $\mathbb{B}, L_k$
\end{algorithmic}
\label{algo:modlars}
\end{algorithm}

\subsection{Asymptotic costs for parallel implementation of Tournament-bLARS}

In this subsection we examine the asymptotic costs for Tournament-bLARS Algorithm \ref{algo:tournamentblars}.\
We start first by the asymptotic costs of mLARS Algorithm \ref{algo:modlars}, which is used by Tournament-bLARS at every iteration.\


Before we compute the asymptotic costs for mLARS we have to bound the cardinality of some sets.\ The cardinality $ \mathbb{I}_v$ is bounded by $|\mathbb{I}_v|\le n/P$.\ 
Let $l$ be the $l$th iteration of Tournament-bLARS, and $\mathbb{I}_l$ be the current selected columns of Tournament-bLARS.\
Then $|\mathbb{I}_l| \le l b$.\
Assuming that we are on the $k$th iteration of mLARS then $|\mathbb{I}_k| \le |\mathbb{I}_l| + b \le lb+b$, and $|\mathbb{I}_k\cup \tilde{\mathbb{I}}_v| \le lb + b + n/P$ for all $k$ if node $v$ is at the bottom of the tree, i.e., $\tilde{\mathbb{I}}_v := \mathbb{I}_v$, otherwise $|\mathbb{I}_k\cup \tilde{\mathbb{I}}_v| \le lb + b + 2b$ for all $k$ because node $v$ not at the bottom of the tree, i.e., $\tilde{\mathbb{I}}_v := \mathbb{B}_v$.\
The cardinality of $\tilde{\mathbb{I}}_v \backslash \mathbb{I}_k$ is bounded by $n/P$ if $v$ is a leaf node because $|\tilde{\mathbb{I}}_v \backslash \mathbb{I}_k| \le |\tilde{\mathbb{I}}_v| = |\mathbb{I}_v| \le n/P$, or otherwise bounded by $2b$ because $|\tilde{\mathbb{I}}_v| = |\mathbb{B}_v| \le 2b$.\
Using these bounds we will compute the asymptotic costs of each step of mLARS.\ Note that there is no parallelism for each individual run of mLARS. Therefore, we only report results for arithmetic operations.\

Step $3$ costs $O(m)$ operations.\
Step $4$ costs $O(mn/P+ mlb + mb)$ at leaf node and $O(mlb+3mb)$ otherwise.\ 
Step $5$ costs $O(lb + b)$.\
Step $7$ costs $O(n/P + lb + b + m)$ at leaf node and $O(lb + 3b + m)$ otherwise.\ 
Step $10$ costs $O(lb^2 + b^2)$.\
Step $11$ costs $O(b(lb +b)^2)$.\
Step $12$ to $13$ cost $O(lb^2 + b^2)$.\
Step $14$ costs $O(mlb^2+mb^2)$.\
Step $15$ costs $O(bmn/P+ mlb^2 + mb^2)$ at leaf node and $O(mlb^2+3mb^2)$ otherwise.\
Steps $16$ to $18$ cost $O(bn/P)$ at leaf node and $O(2b^2)$ otherwise.\
Step $19$ costs $O(m)$.\
Steps $20$ to $21$ cost $O(bn/P + lb^2 + b^2)$ at leaf node and $O(lb^2 + 3b^2)$ otherwise.\
Step $22$ costs $O(lb^2 + b^2)$.\
Step $23$ costs $O(mlb^2 + mb^2)$.\
Step $24$ costs $O(b(lb + b)^2)$.\
Steps $25$ to $26$ cost $O(lb^2 + b^2)$.\
For $t$ columns we need to run Tournament-bLARS for $t/b$ iterations and each iteration makes $\log P$ parallel calls to mLARS which results in 
$$
t/b \cdot  (\mbox{arithmetic cost of mLARS at leaf node}) ~ + ~ t/b \cdot (\mbox{arithmetic cost of mLARS at non-leaf node}) \cdot \log P
$$
total operations.\ Therefore, in Big O notation Tournament-bLARS requires 
$$
F= O\left(\frac{tmn}{P} + \frac{tmn}{bP} + \left(t^2m + t^3\right) \log P \right) 
$$ 
operations.\
Communication occurs $\log P$ times because of the binary tree and another $\log P$ times to broadcast data from the root node to the rest of the nodes.\ Therefore Tournament-bLARS requires 
$$
L=2\frac{t}{b} \log P
$$ 
messages.\ Each node (except of the root) communicates $bm$ words for columns in $\mathbb{B}$.\ Therefore the execution of the binary tree requires $t m\log P$ words.\ Broadcasting data from the root node to the rest of the nodes at Step $12$ costs a total of 
$$
W= O\left(\left(tm + \frac{tm+t^2}{b} + tb\right)\log P\right)
$$ 
words.\

\section{Comparison of asymptotic costs}\label{sec:costs}

In this section, we compare the asymptotic costs of parallel LARS, bLARS and T-bLARS. The results are shown in Table~\ref{tab:asymptcosts}.\
Note that parallel bLARS becomes faster than parallel LARS for $b>1$.\ 
Parallel bLARS and T-bLARS have similar latency costs.\ However, an important difference is that the number of words for parallel bLARS depends on the number of columns $n$ while the number of words for T-bLARS depend on the number of rows $m$.\ T
his is due to the fact that for parallel bLARS we partition the data per row, while for T-bLARS we partition the data per column.\ Therefore, in the high-dimensional regression setting where $n \gg m$, T-bLARS requires communicating much fewer words than bLARS.
We compare the two methods empirically in Section \ref{sec:experiments}.\

We note that even though the results in Table~\ref{tab:asymptcosts} are obtained by assuming matrix $A$ is dense, the complexity bounds trivially extend to sparse matrices as long as we have balanced partitions, i.e., the local sparse matrices stored at different processors should have similar number of nonzero entries. In the balanced sparse case, we simply replace $mn$ with the number of nonzeros $\mbox{nnz}(A)$ and obtain the arithmetic complexity for all methods. The communication costs stay the same. In Section 10 we use balanced partition to deal with sparse matrices. 

\begin{table}
\centering
\resizebox{\columnwidth}{!}{%
 \begin{tabular}{||c c c c||} 
 \hline
 Method & Arithmetic operations & Words communicated& Messages \\ [0.5ex] 
 \hline\hline
 LARS & $\frac{tmn}{P} + \frac{t^2m}{P} + tn + t^3$ & $tn\log P + t^2 \log P$& $t\log P$ \\ [3ex] 
 bLARS & $\frac{tmn}{bP} + \frac{tn}{b} + \frac{t^2m}{P} + t^3$ & $\frac{tn}{b}\log P + t^2 \log P$& $\frac{t}{b}\log P$ \\ [3ex] 
 T-bLARS & $\frac{tmn}{P} + \frac{tmn}{bP} + \left(t^2m + t^3\right) \log P$ & $\left(tm + \frac{tm}{b} + tb\right)\log P+ \frac{t^2}{b}\log P$ & $\frac{t}{b} \log P$\\ [1ex] 
 \hline
\end{tabular}
}
\caption{Asymptotic costs for parallel LARS, bLARS, T-bLARS.\ Here, $t$ is the required number of columns to be outputted by all algorithms.\ We assume that $t\gg b$ and that matrix $A$ is dense.}
\label{tab:asymptcosts}
\end{table}

\section{Empirical performance}\label{sec:experiments}

\begin{figure}[h]
	\centering
	\begin{subfigure}{.33\textwidth}
	\captionsetup{justification=centering}
  		\centering
  		\includegraphics[width=\textwidth]{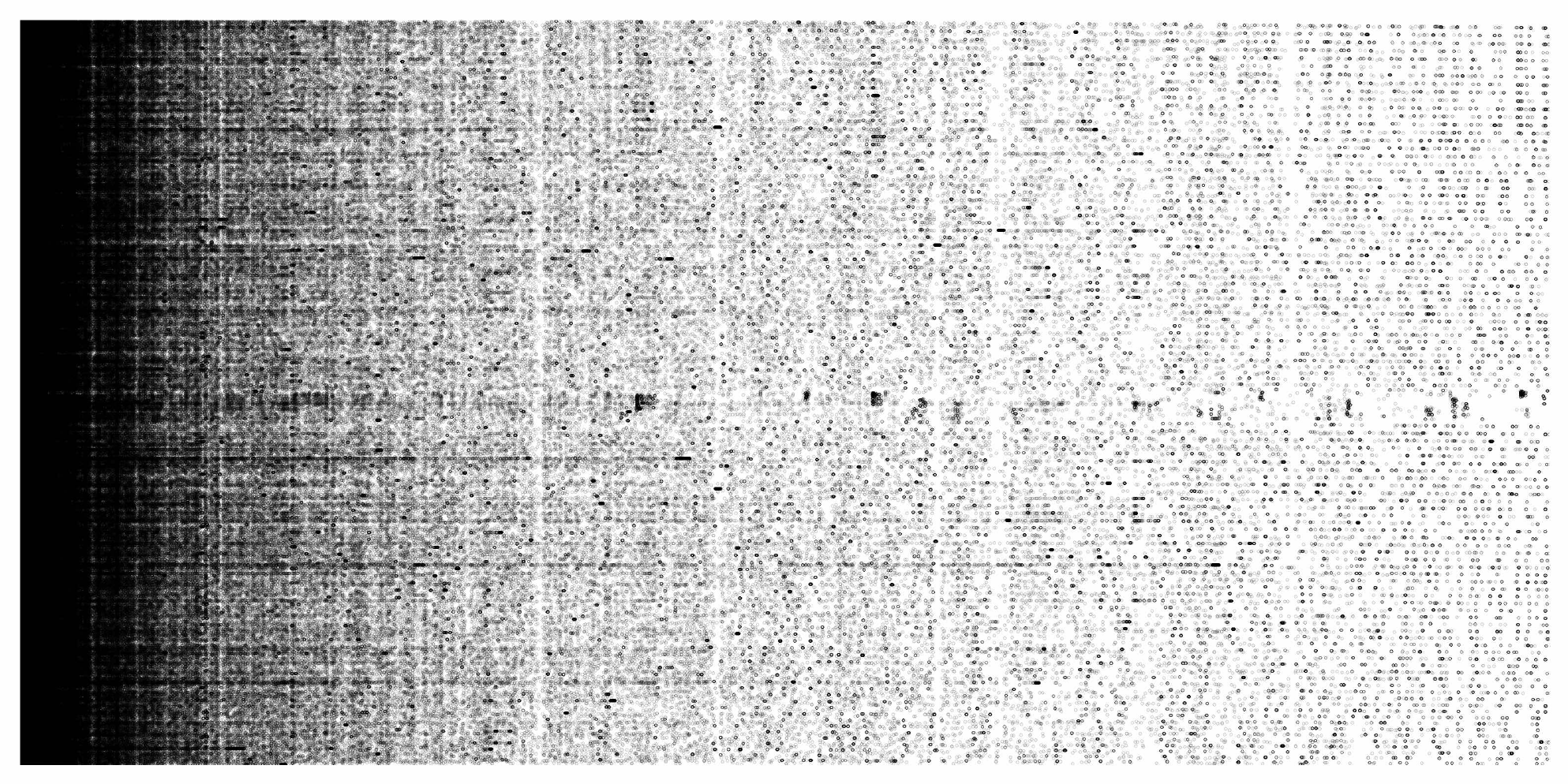}
		\caption{sector}
	\end{subfigure}%
	\begin{subfigure}{.33\textwidth}
	\captionsetup{justification=centering}
  		\centering
  		\includegraphics[width=\textwidth]{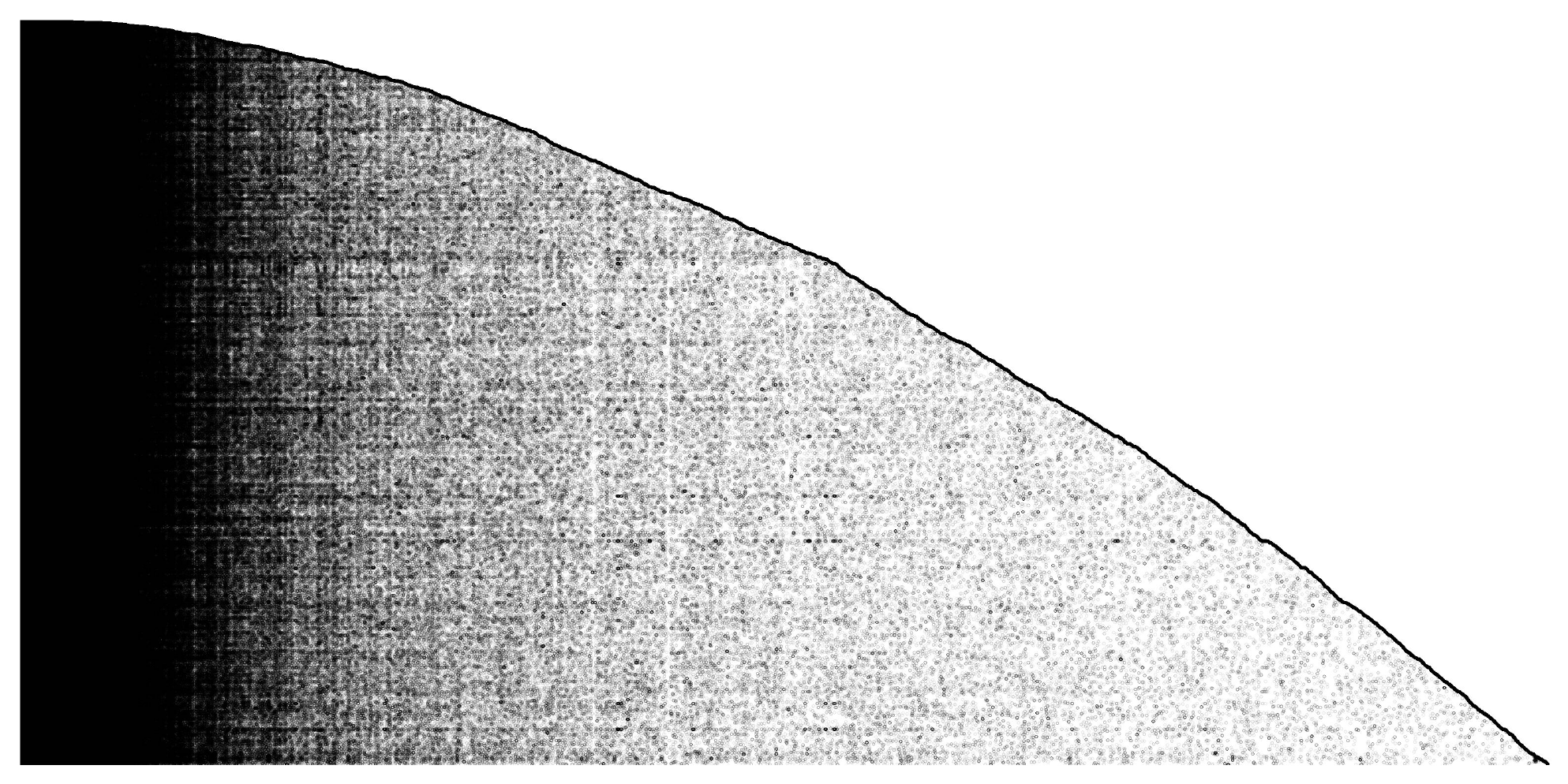}
		\caption{E2006\_tfidf}
	\end{subfigure}
	\begin{subfigure}{.33\textwidth}
	\captionsetup{justification=centering}
  		\centering
  		\includegraphics[width=\textwidth]{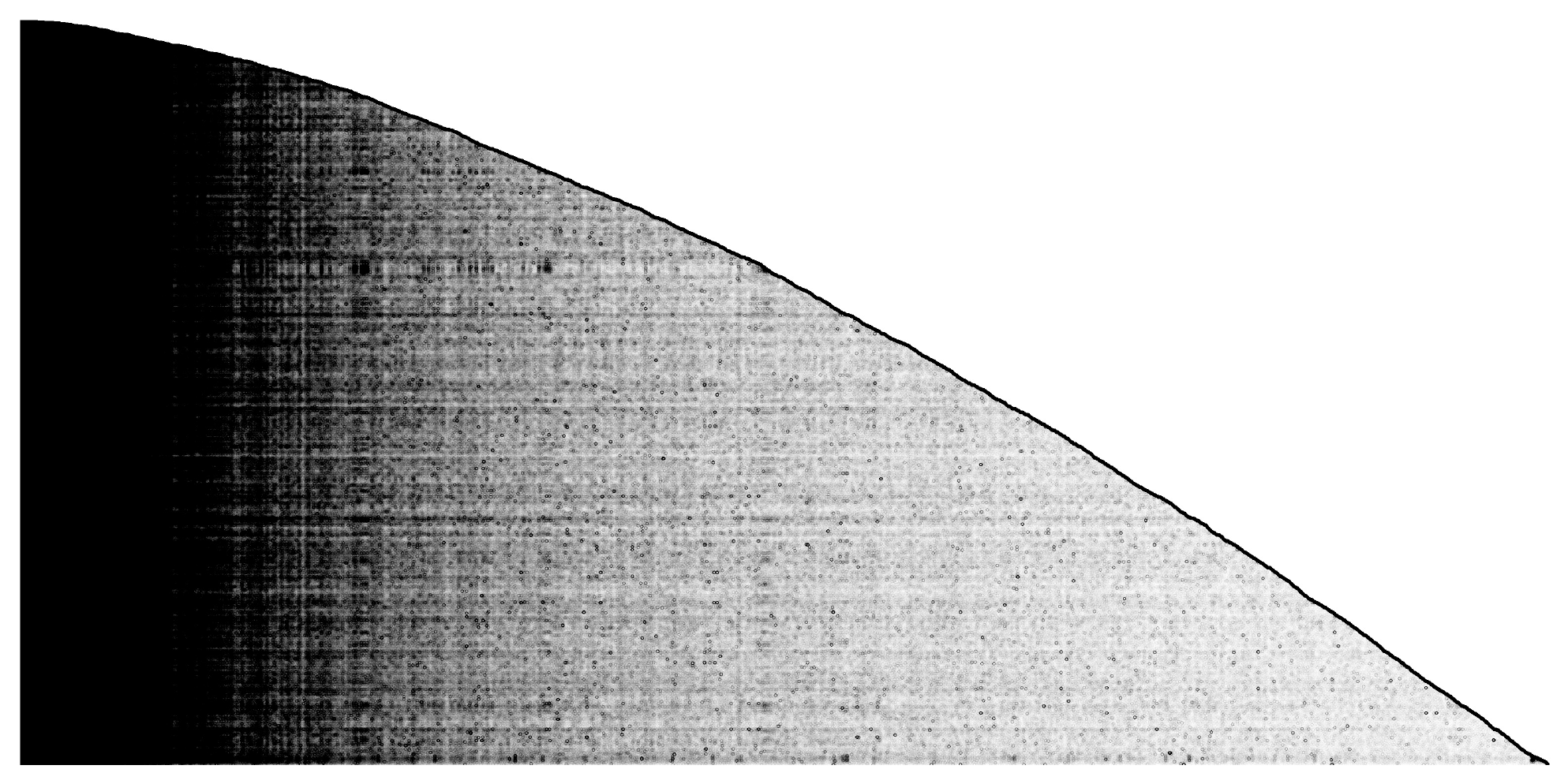}
		\caption{E2006\_log1p}
	\end{subfigure}
	\begin{subfigure}{.33\textwidth}
	\captionsetup{justification=centering}
  		\centering
  		\includegraphics[width=\textwidth]{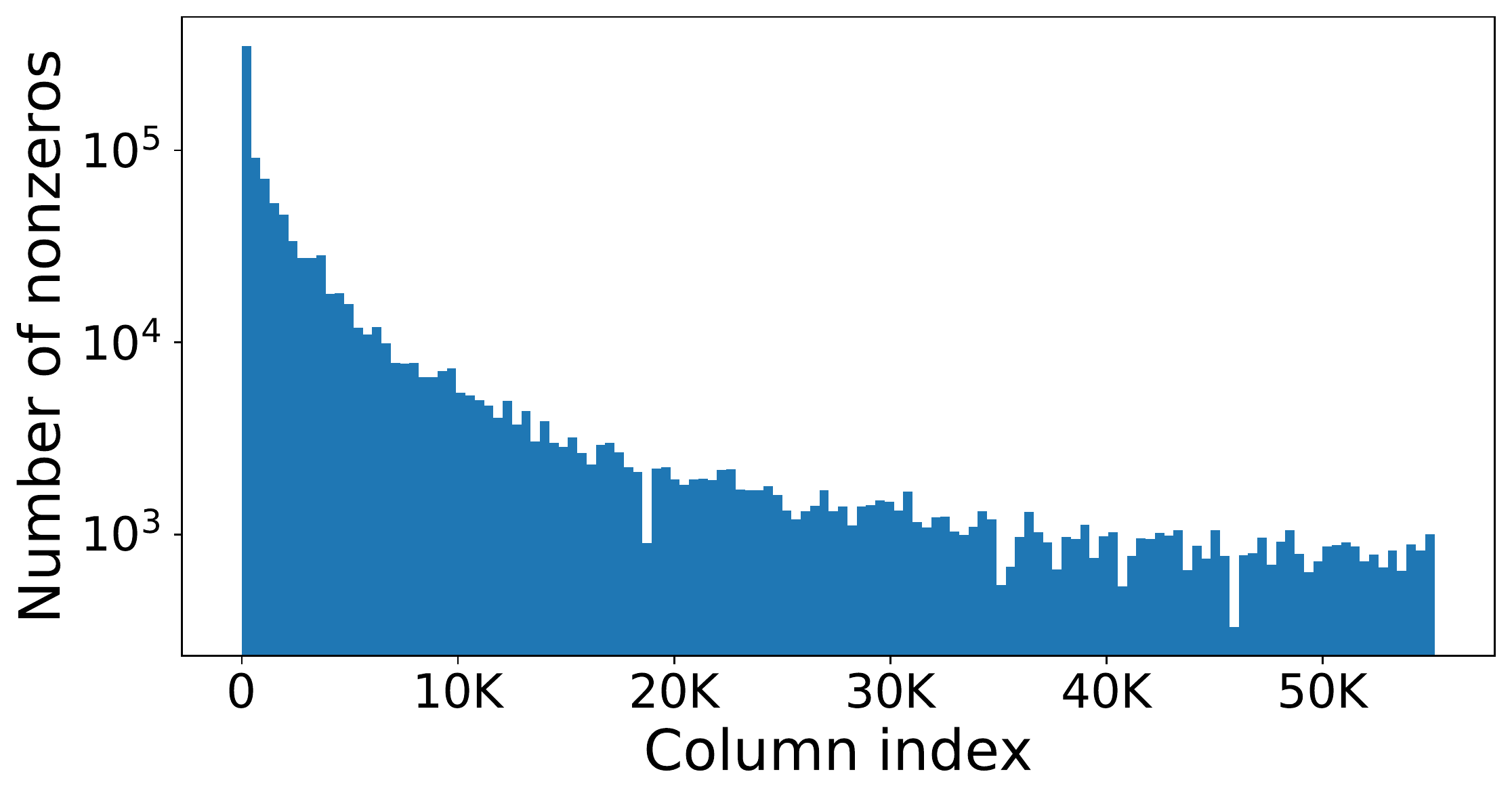}
		\caption{sector}
	\end{subfigure}%
	\begin{subfigure}{.33\textwidth}
	\captionsetup{justification=centering}
  		\centering
  		\includegraphics[width=\textwidth]{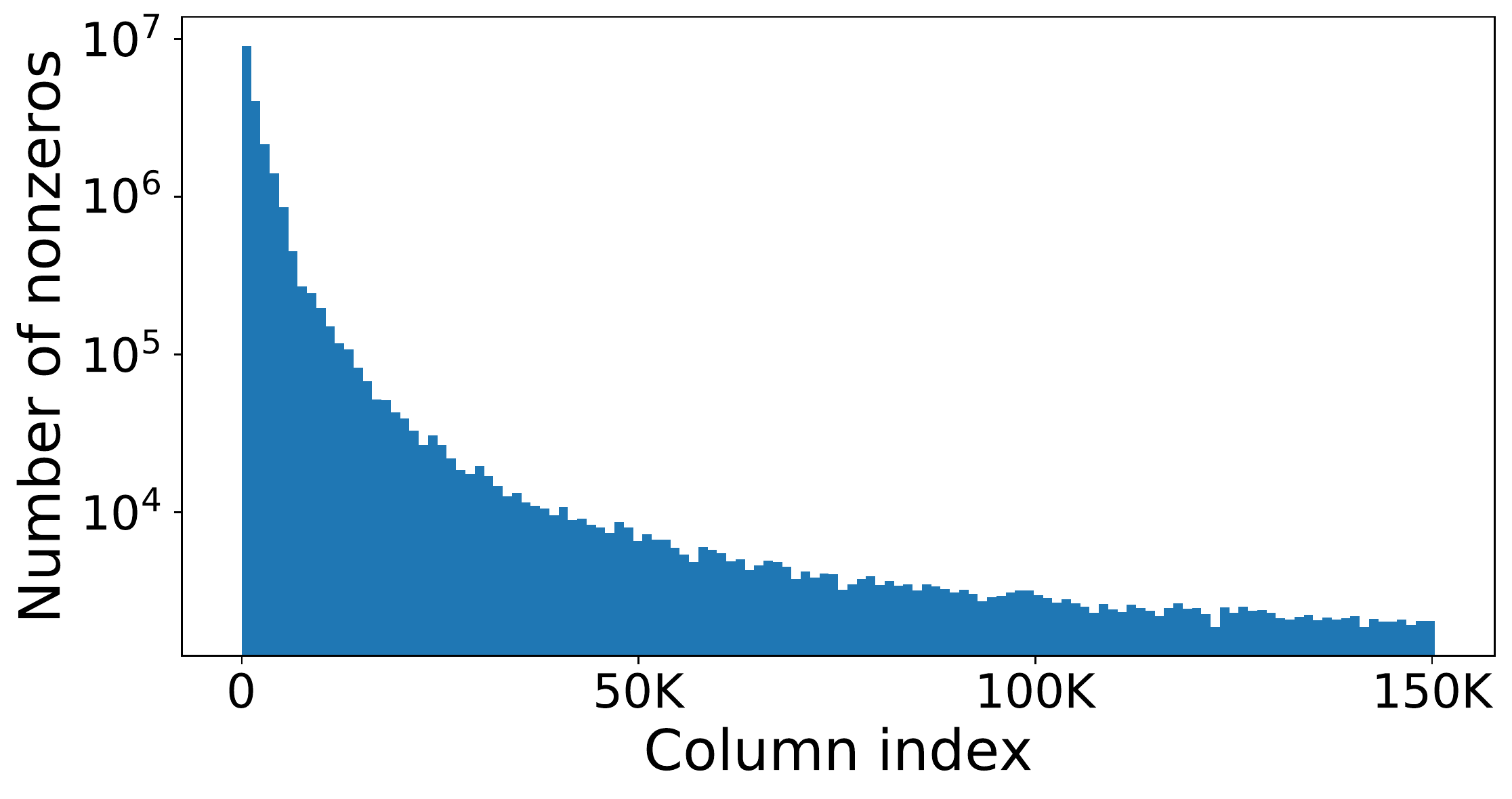}
		\caption{E2006\_tfidf}
	\end{subfigure}
	\begin{subfigure}{.33\textwidth}
	\captionsetup{justification=centering}
  		\centering
  		\includegraphics[width=\textwidth]{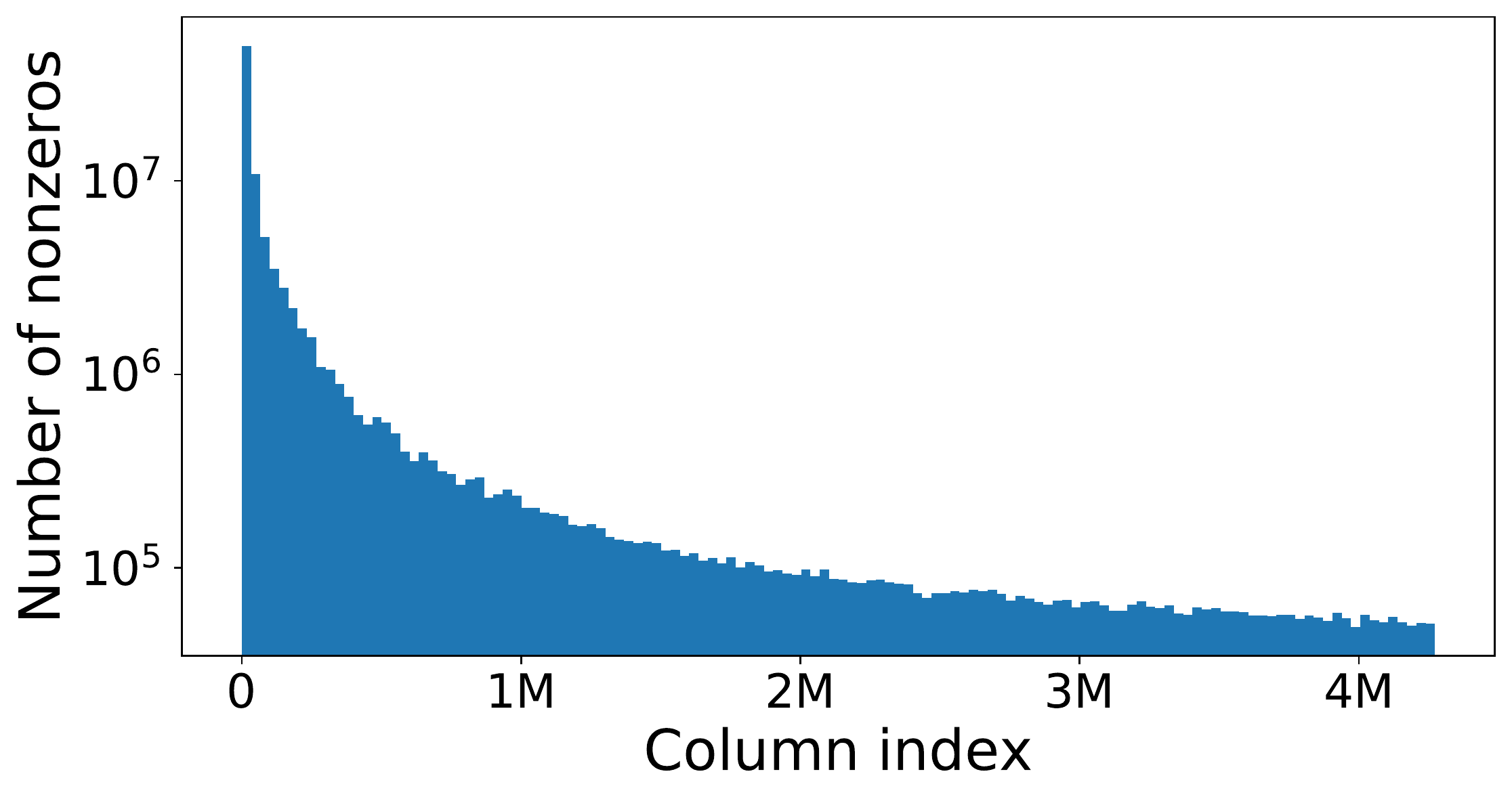}
		\caption{E2006\_log1p}
	\end{subfigure}
	\caption{Sparsity pattern and distribution of nonzeros for sparse datasets sector and E2006. The histograms (d)-(f) are drawn on 128 equally spaced bins.}
	\label{fig:sparsity}
\end{figure}

This section contains two parts. First, we evaluate and compare the solution quality of bLARS and T-bLARS for a range of block sizes $b$ and processors $P$. Second, we present a comprehensive list of plots that demonstrate both overall speedups and more detailed running time breakdowns from increasing $b$ and $P$. We carry out the experiments on four regression datasets summarized in Table~\ref{tab:LRprobs}. The data matrices for sector and E2006 are sparse and column-wise unbalanced, i.e., the distribution of nonzeros per column is skewed (Figure~\ref{fig:sparsity}). In order to balance the computation workload on all processors, for T-bLARS, we distribute the columns of these sparse matrices so that the partitioned columns at each processor have roughly the same number of nonzeros. Other column partitioning could also be used. We discuss the effect of column partition on solution quality of T-bLARS in the next subsection. For comparison purposes we limit both algorithms to collect the first 75 columns. We implemented the code in Python and used the optimized mpi4py library~\cite{mpi4py}. The code is run on a computer cluster with distributed memory. Each node in the cluster comes with 2 x Intel E5-2683 v4, 128 GB of RAM.

\begin{table}[h!]
\centering
\begin{tabular}{lccc}
\hline
Dataset& $m$	& $n$		 & \rm{nnz($A$)}/$mn$  \\\hline
sector	        & $6412$    & $55197$	& $0.003$ \\
YearPredictionMSD	        & $463715$      & $90$	& $1.00$  \\
E2006\_log1p & $16087$      & $4272227$	& $0.001$  \\ 
E2006\_tfidf	        & $16087$      & $150360$	& $0.008$\\\hline
\end{tabular}
\caption{Properties of the datasets that we consider.\ \rm{nnz($A$)} denotes the number of nonzeros in matrix $A$, consequently, the fourth column gives the (relative) sparsity of $A$.\ The first four are synthetic data. The regression datasets can be downloaded from \cite{Chang11} as part of the LIBSVM Data package.\ The E2006 and Year datasets are the three largest regression datasets in LIBSVM.}
\label{tab:LRprobs}
\end{table}

\subsection{Solution quality}

\begin{figure}[h]
	\centering
	\begin{subfigure}{.45\textwidth}
	\captionsetup{justification=centering}
  		\centering
  		\includegraphics[width=\textwidth]{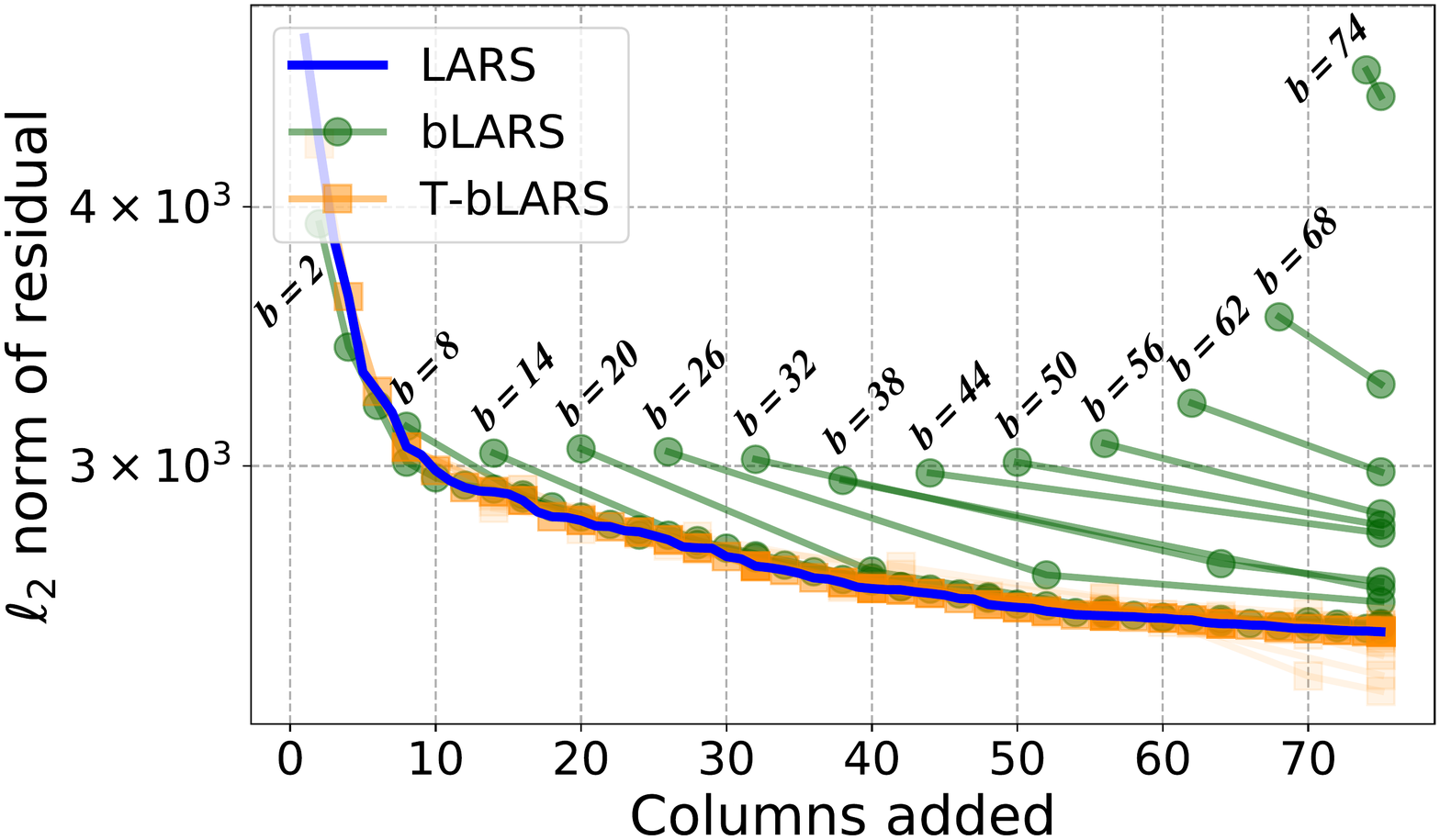}
		\caption{sector}
	\end{subfigure}%
	\begin{subfigure}{.45\textwidth}
	\captionsetup{justification=centering}
  		\centering
  		\includegraphics[width=\textwidth]{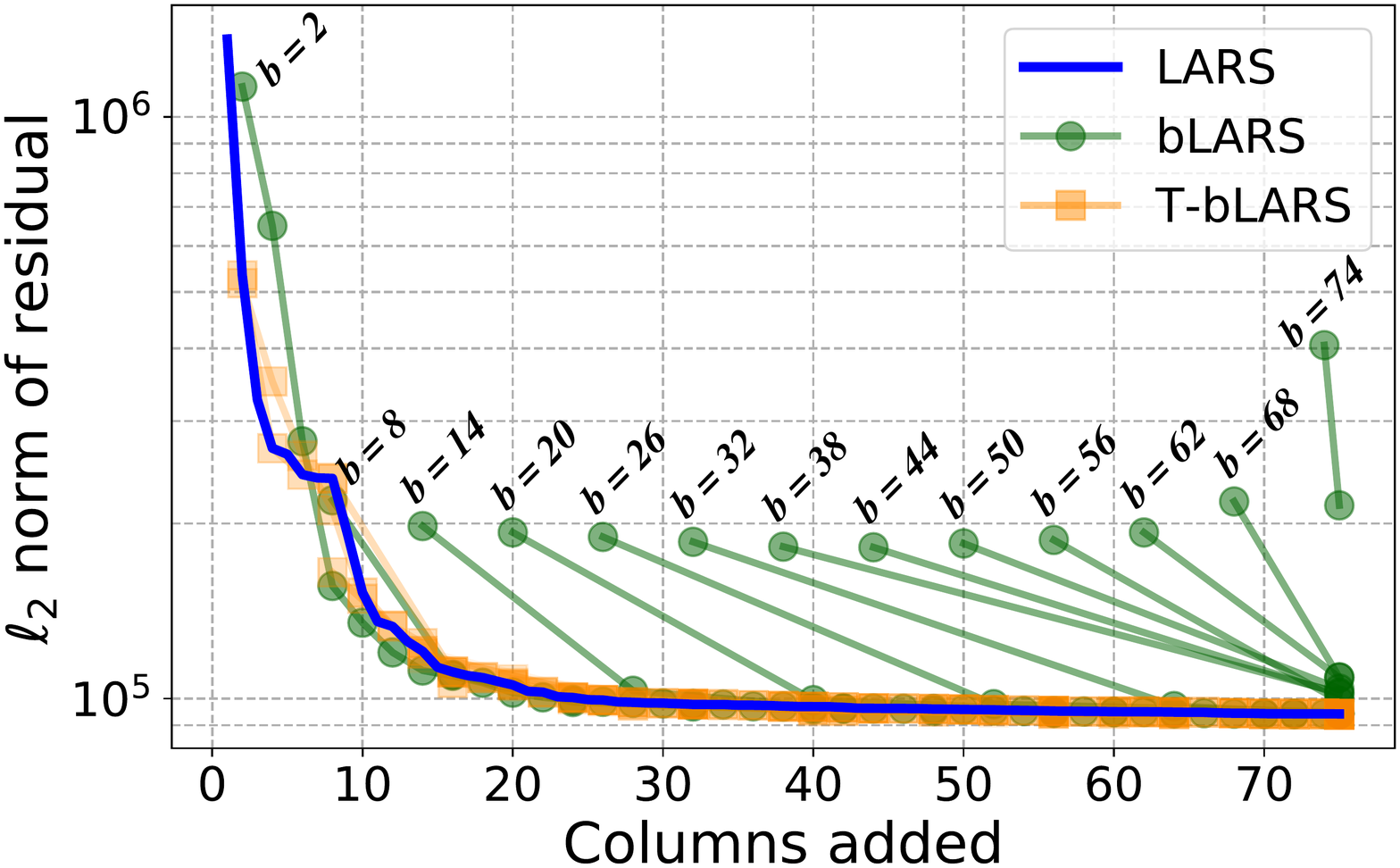}
		\caption{Year.}
	\end{subfigure}
	\begin{subfigure}{.45\textwidth}
	\captionsetup{justification=centering}
  		\centering
  		\includegraphics[width=\textwidth]{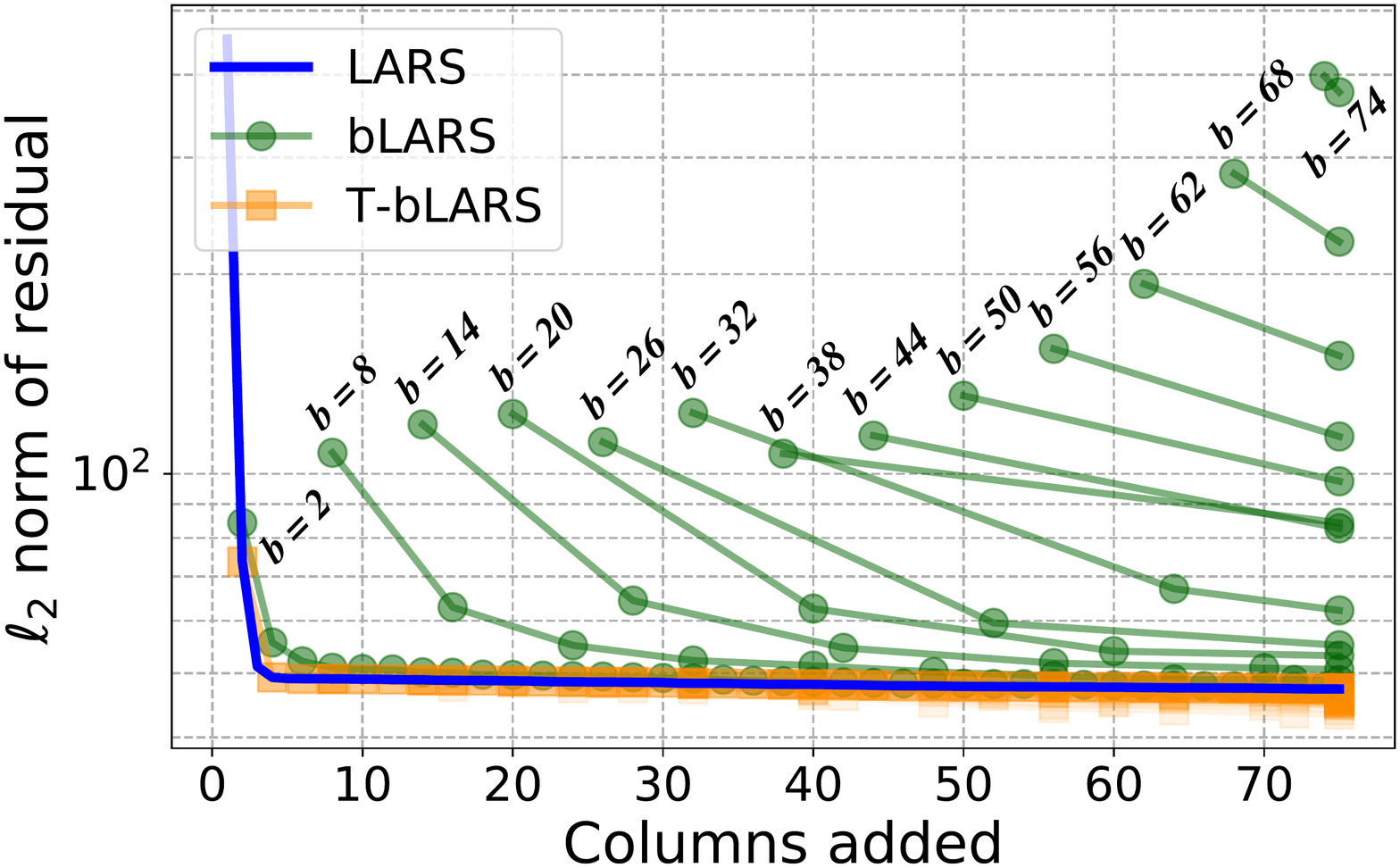}
		\caption{E2006\_tfidf}
	\end{subfigure}
	\begin{subfigure}{.45\textwidth}
	\captionsetup{justification=centering}
  		\centering
  		\includegraphics[width=\textwidth]{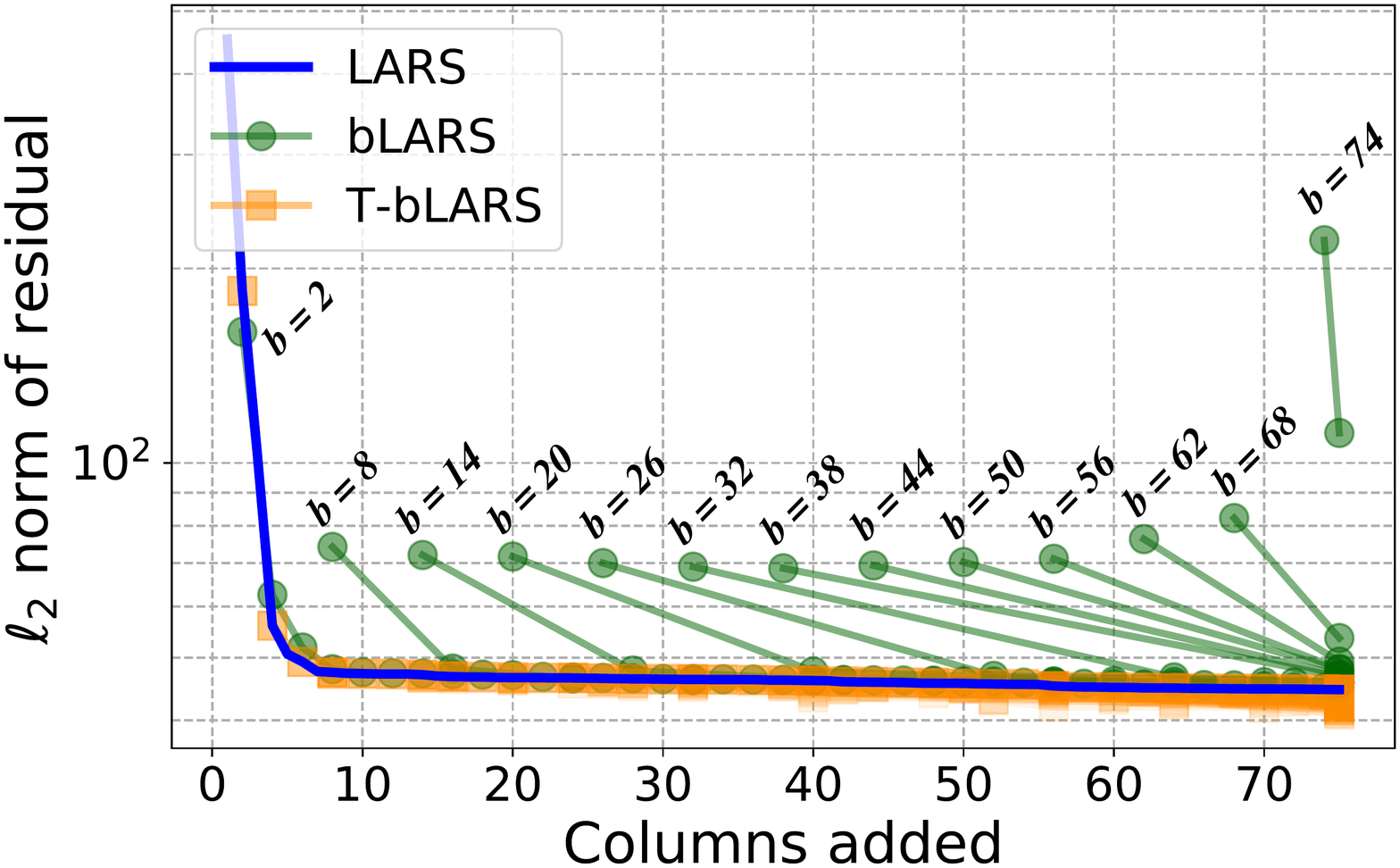}
		\caption{E2006\_log1p}
	\end{subfigure}
	\caption{$\ell_2$-norm of residuals. \ For T-bLARS each line corresponds to a setting of $P$ and $b$.\ We do not show all legends for T-bLARS to ease readability, most settings give similar quality.\ For bLARS each line corresponds to a different $b$.\ Note that $P$ does not affect the quality of bLARS.\ $75$ columns where chosen for all experiments.}
	\label{fig:residual_real}
\end{figure}

\begin{figure}[h]
	\centering
	\begin{subfigure}{.45\textwidth}
	\captionsetup{justification=centering}
  		\centering
  		\includegraphics[width=\textwidth]{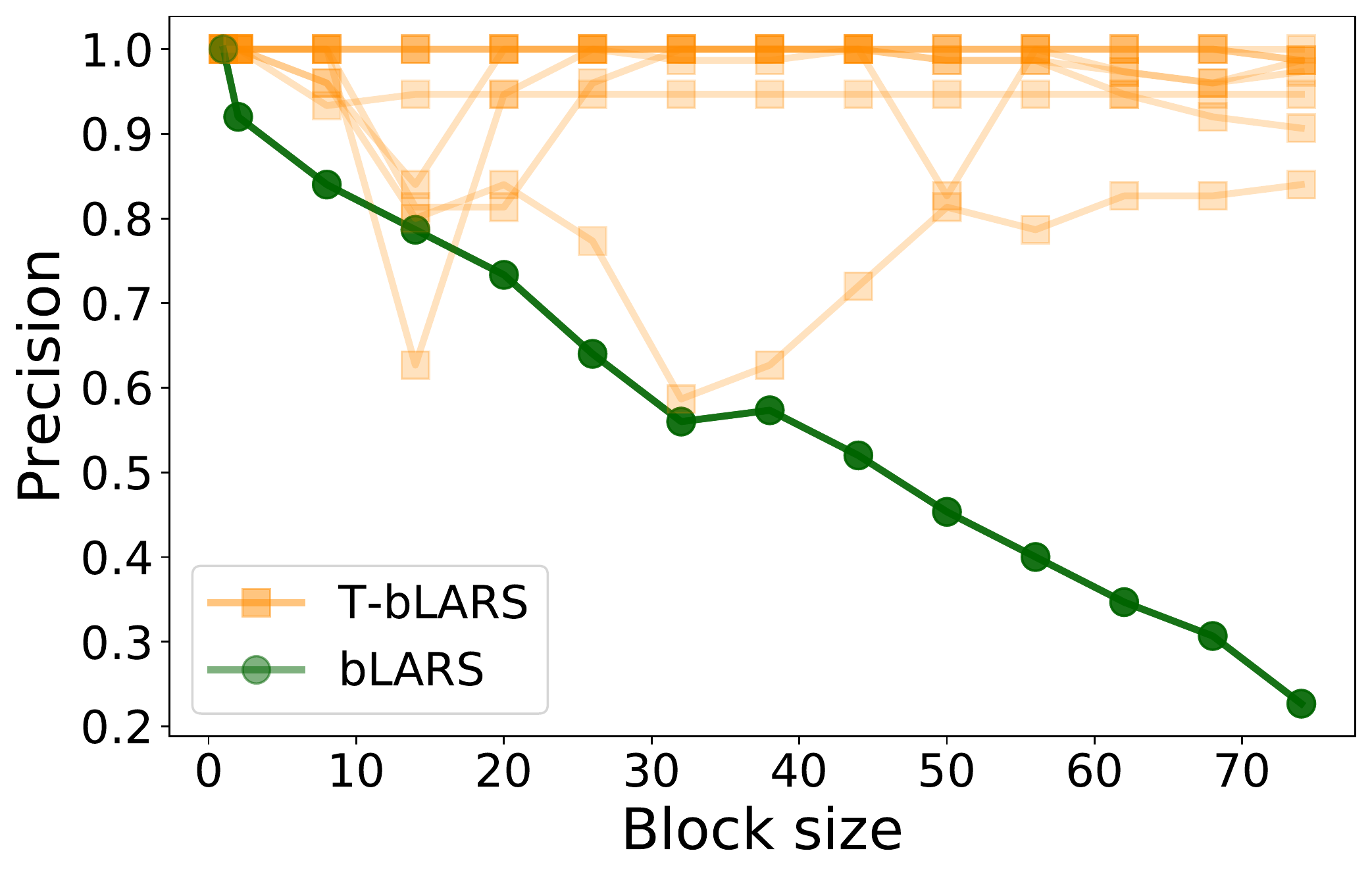}
		\caption{sector}
	\end{subfigure}%
	\begin{subfigure}{.45\textwidth}
	\captionsetup{justification=centering}
  		\centering
  		\includegraphics[width=\textwidth]{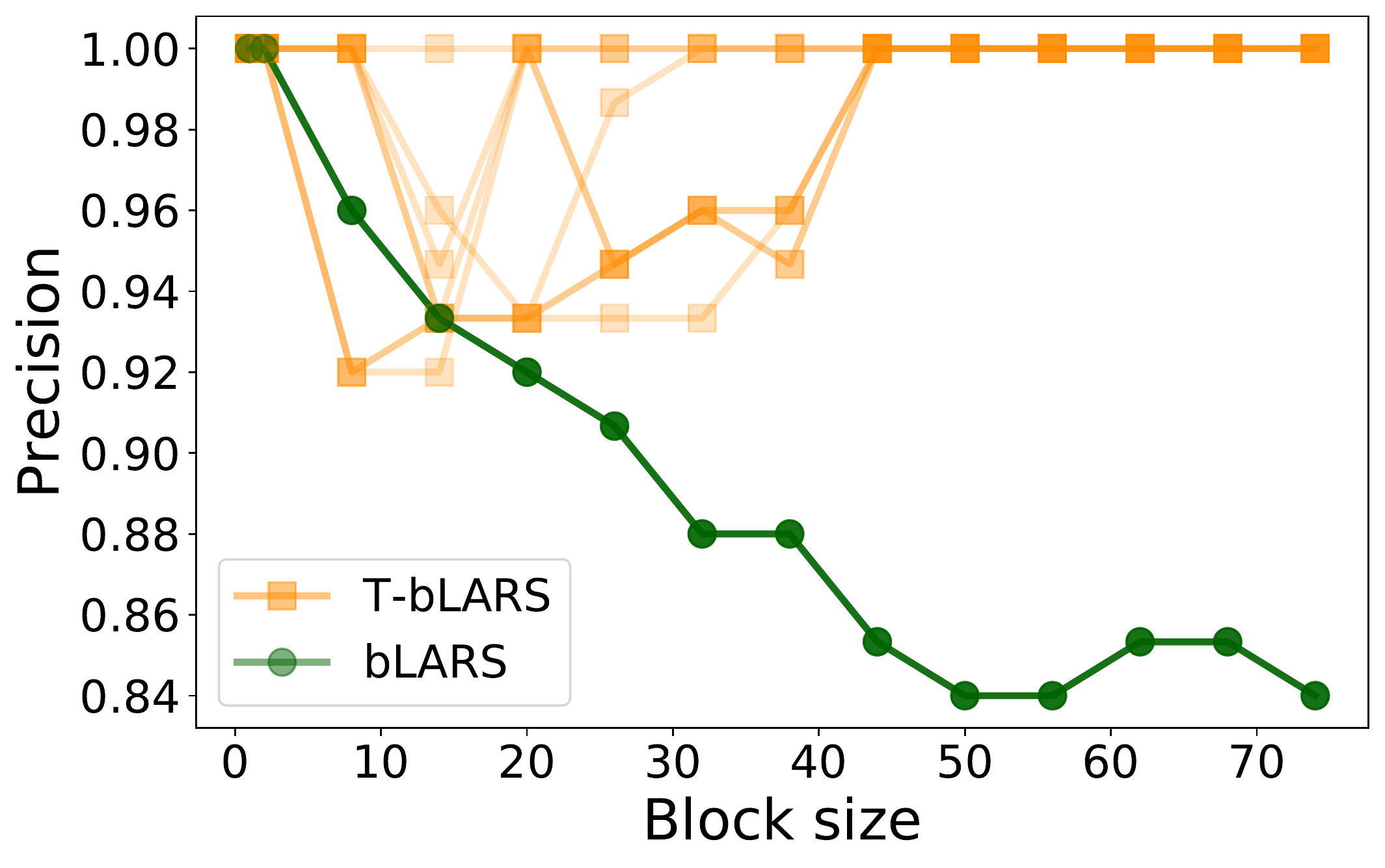}
		\caption{Year.}
	\end{subfigure}
	\begin{subfigure}{.45\textwidth}
	\captionsetup{justification=centering}
  		\centering
  		\includegraphics[width=\textwidth]{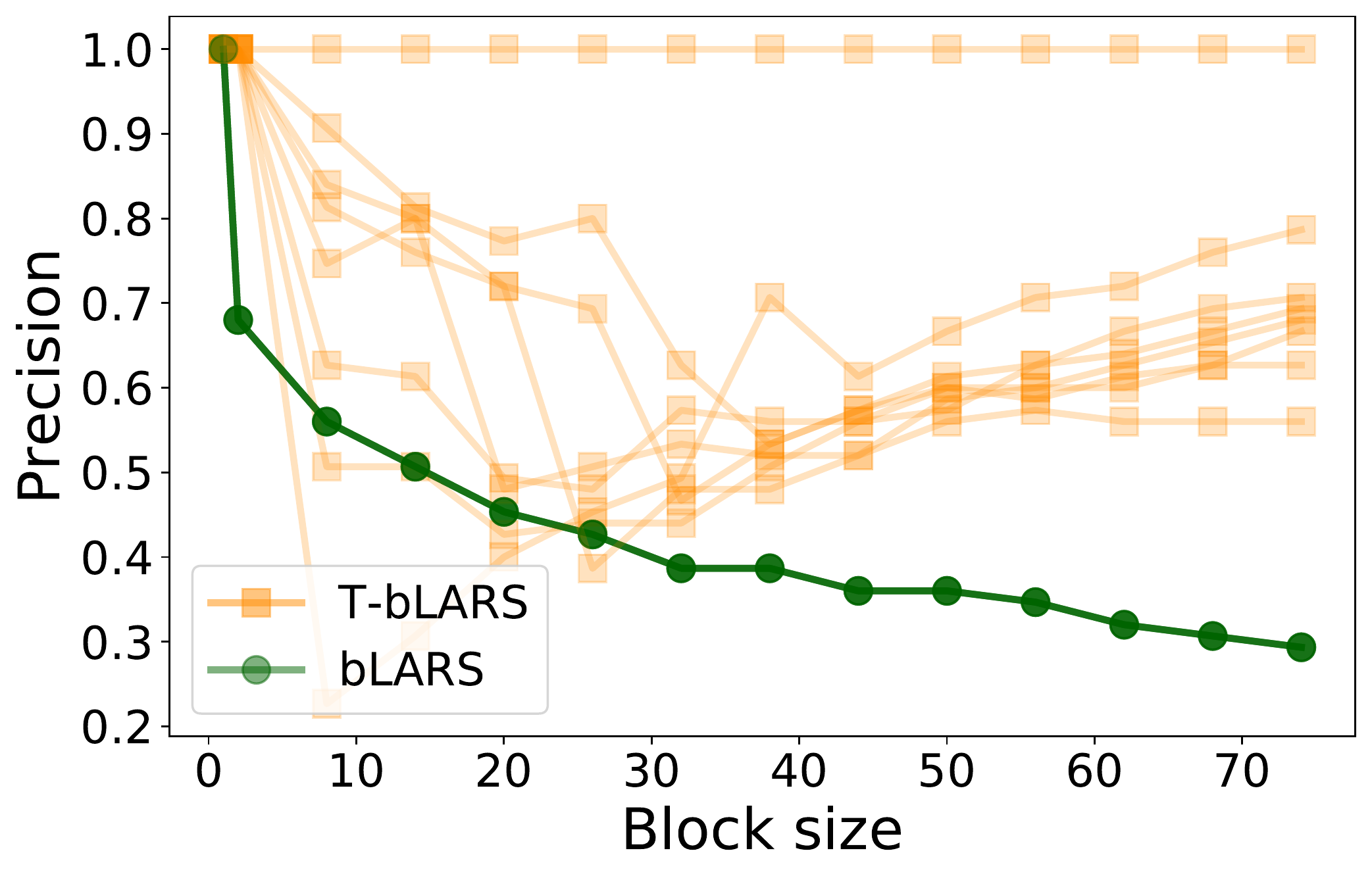}
		\caption{E2006\_tfidf}
	\end{subfigure}
	\begin{subfigure}{.45\textwidth}
	\captionsetup{justification=centering}
  		\centering
  		\includegraphics[width=\textwidth]{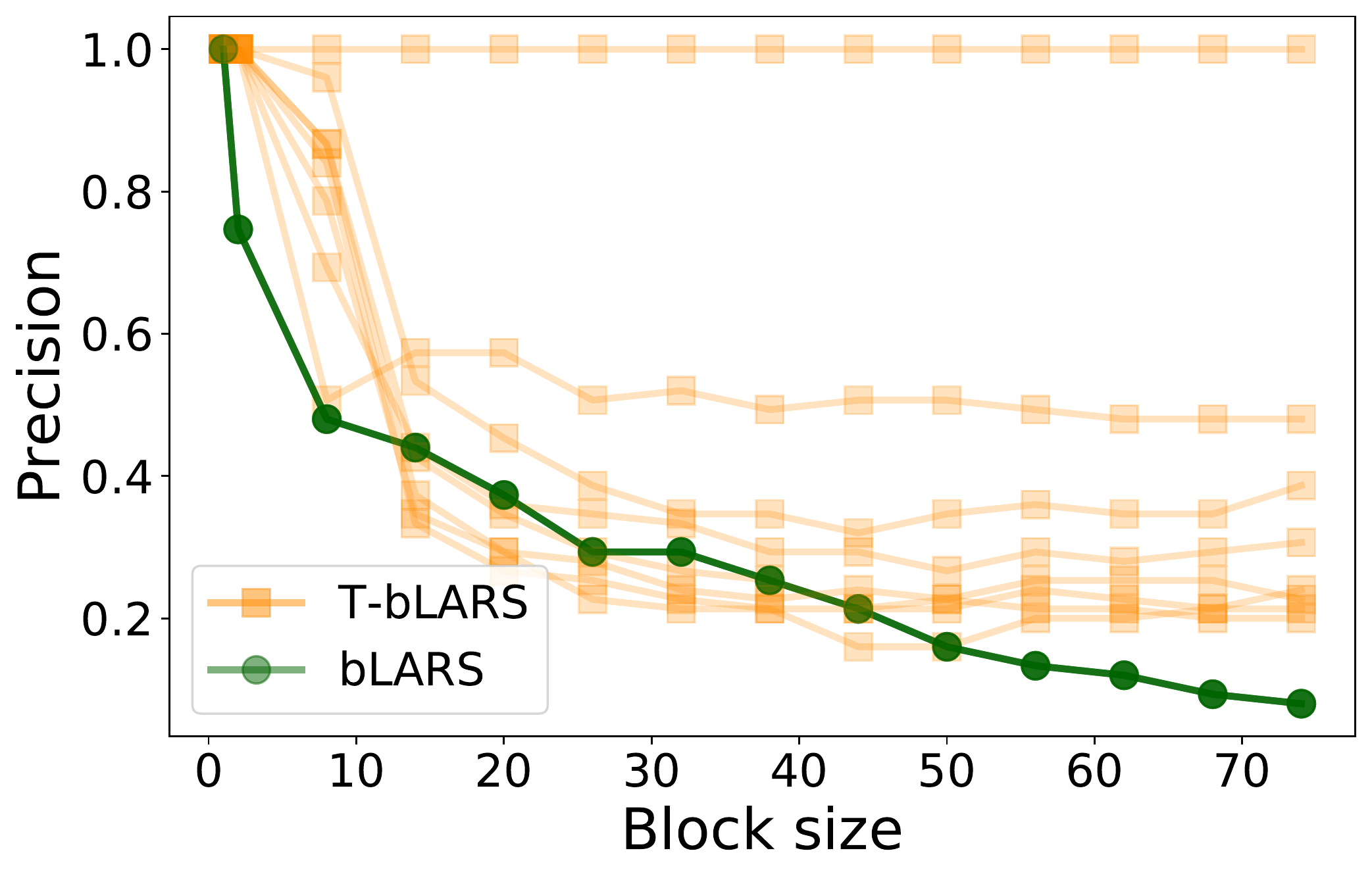}
		\caption{E2006\_log1p}
	\end{subfigure}
	\caption{Precision in column selection.\ For both bLARS and T-bLARS each line corresponds to a setting of $P$. Note that different $P$'s give rise to different row partitions for bLARS and different column partitions for T-bLARS. Row partitions do not affect the precision of bLARS.}
	\label{fig:precision_real}
\end{figure}

We use two metrics to measure solution quality. One metric is, for a given parameter $b$, the value of the $\ell_2$-norm of the residual vector versus the number of columns added at each iteration (Figure~\ref{fig:residual_real}). For the second metric, since LARS is primarily used for column selection in regression, we treat the columns selected by LARS as the ground truth, and we use precision in column selection to measure performance, i.e., we compare the percentage of columns selected by bLARS and T-bLARS that overlap with the columns selected by LARS (Figure~\ref{fig:precision_real}).

\begin{figure}[h]
	\centering
	\begin{subfigure}{.245\textwidth}
	\captionsetup{justification=centering}
  		\centering
  		\includegraphics[width=\textwidth]{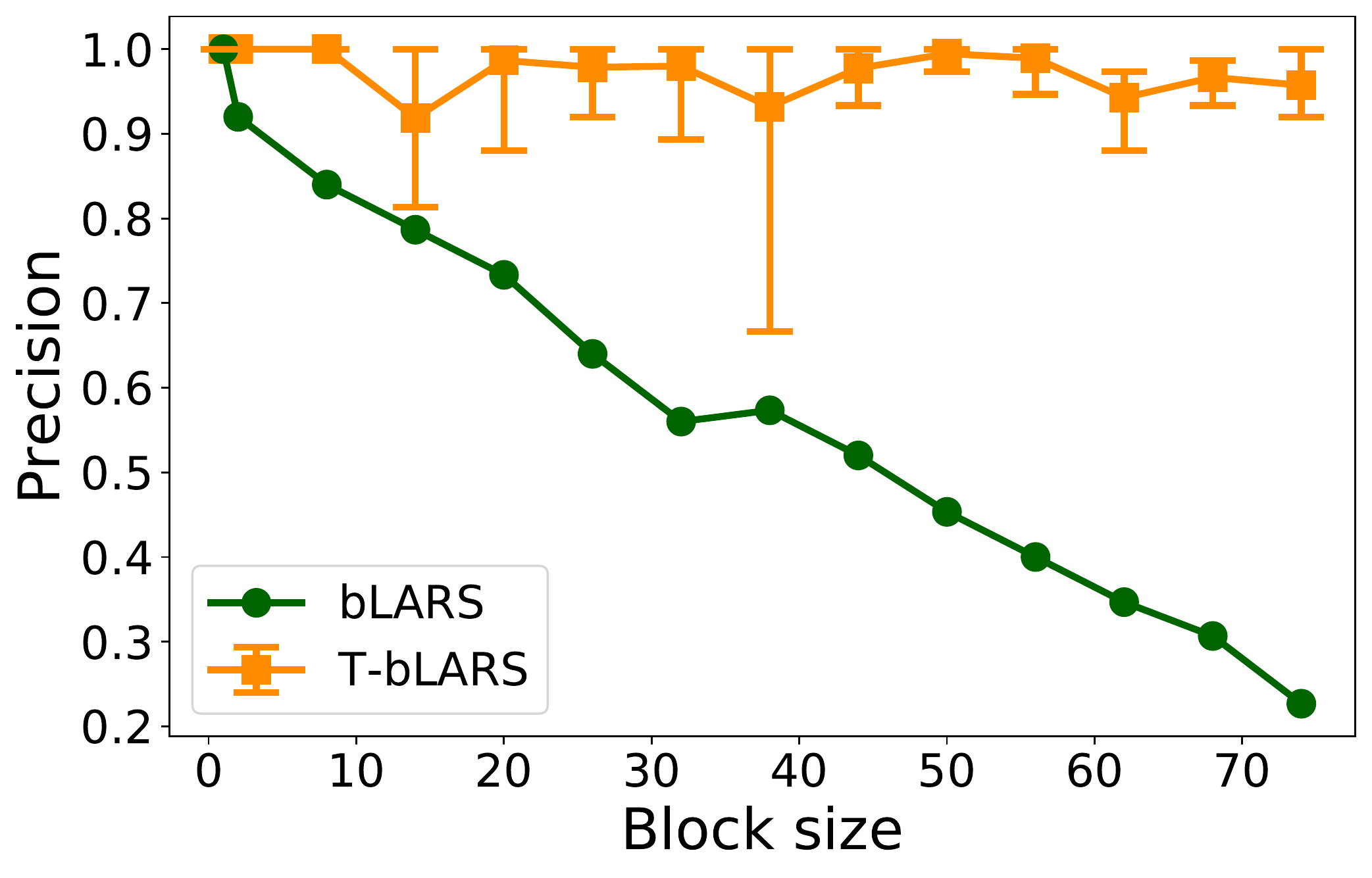}
		\caption{sector}
	\end{subfigure}%
	\begin{subfigure}{.245\textwidth}
	\captionsetup{justification=centering}
  		\centering
  		\includegraphics[width=\textwidth]{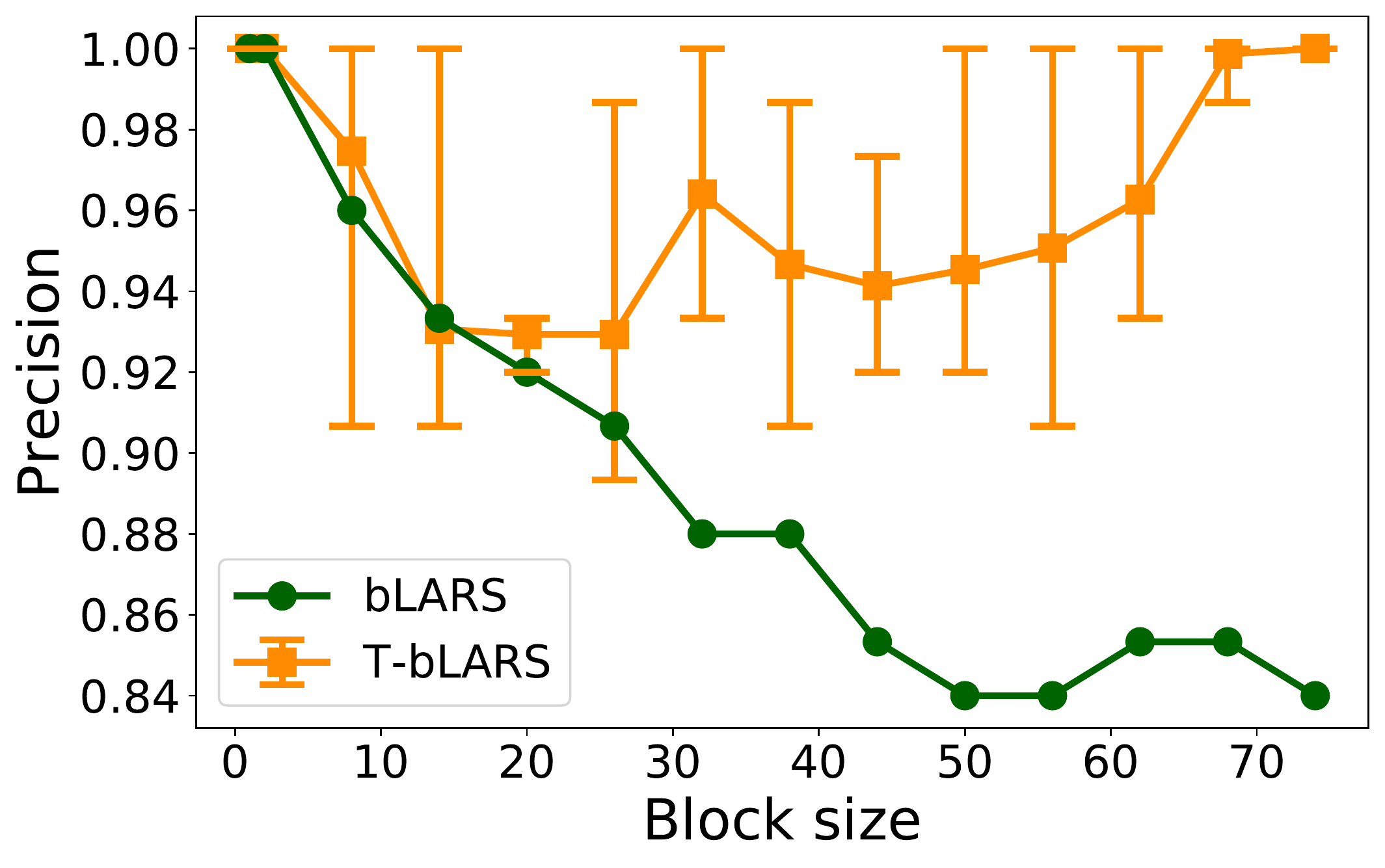}
		\caption{Year.}
	\end{subfigure}
	\begin{subfigure}{.245\textwidth}
	\captionsetup{justification=centering}
  		\centering
  		\includegraphics[width=\textwidth]{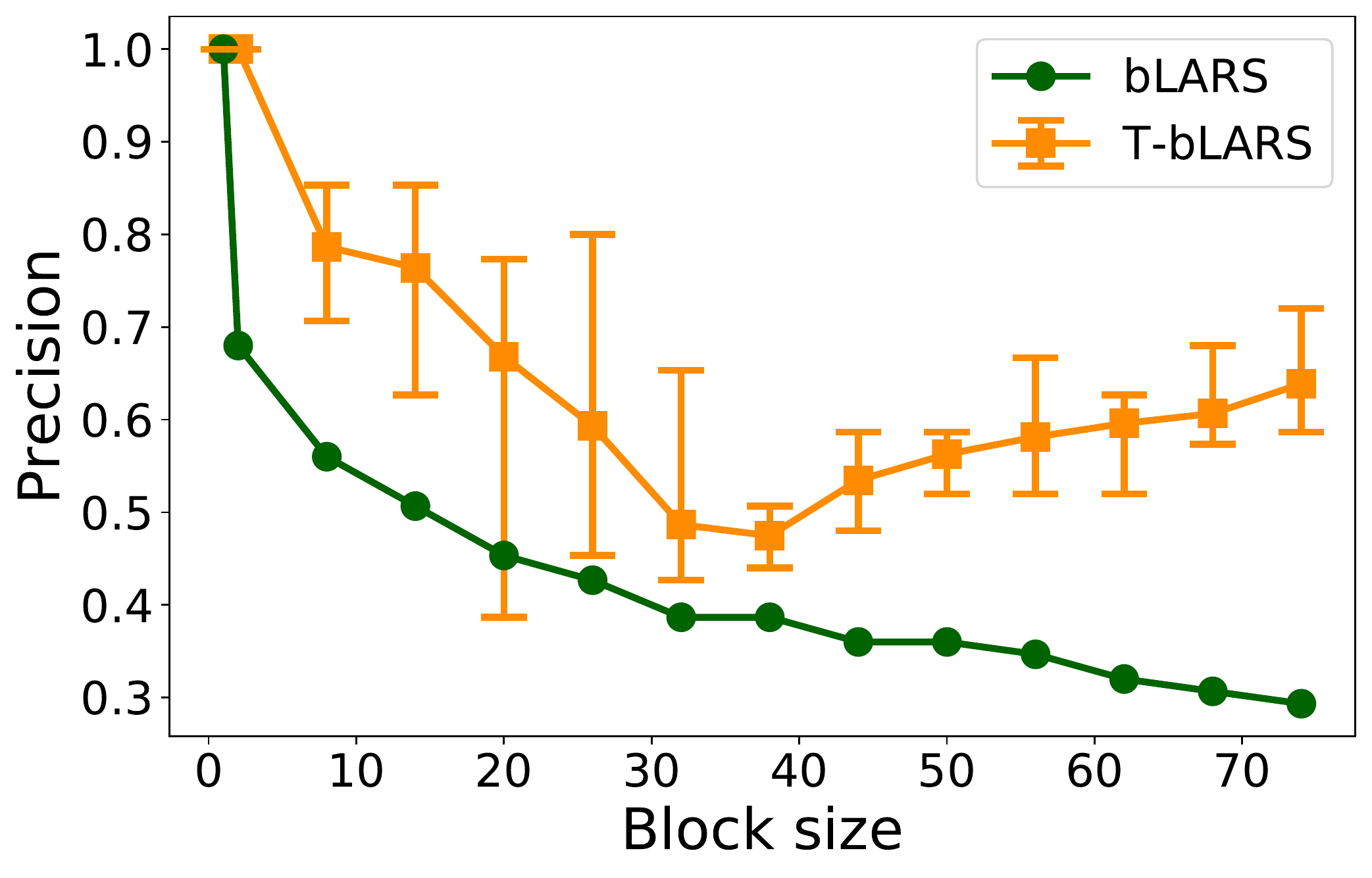}
		\caption{E2006\_tfidf}
	\end{subfigure}
	\begin{subfigure}{.245\textwidth}
	\captionsetup{justification=centering}
  		\centering
  		\includegraphics[width=\textwidth]{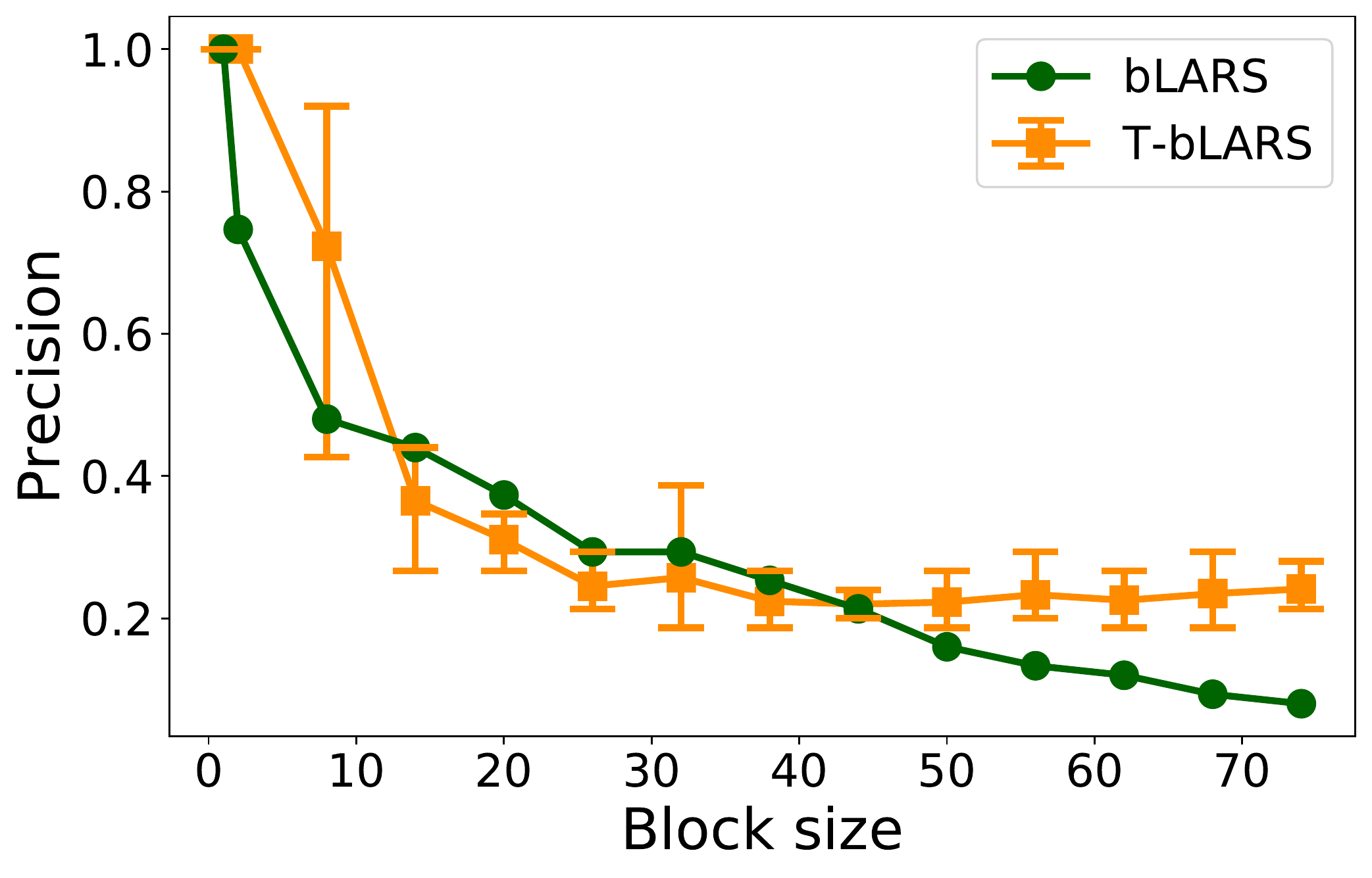}
		\caption{E2006\_log1p}
	\end{subfigure}
	\caption{Effects of column partitions on the precision of column selection for T-bLARS. We fix $P=128$ and run T-bLARS on 10 random column partitions. The bars for each $b$ show the minimum and maximum precisions over the 10 runs, and the line in the middle connects the mean.}
	\label{fig:random_partition}
\end{figure}

Observe that T-bLARS is overall more successful in terms of both data fitting and column selection. The $\ell_2$-norm of the residual produced by T-bLARS is nearly identical to that of LARS on all datasets and for all choices of $b$ and $P$. On the other hand, bLARS has higher residuals as $b$ increases. For column selection, we see a decrease in precision for both methods when $b > 1$, but in most settings T-bLARS recovers more columns than bLARS. In particular, the precision of bLARS keeps dropping quickly as $b$ increases, while on three out of four datasets the precision of T-bLARS goes up again for larger $b$. 
This makes sense because for T-bLARS, the larger the block size is, the more columns will be sent from leaf nodes to non-leaf nodes to choose from. 

For bLARS, how rows are partitioned among processors does not affect the columns selected by the algorithm. For T-bLARS, different column partitions can lead to different tournaments at non-leaf nodes and thus cause T-bLARS to select different columns at the root node. Figure~\ref{fig:random_partition} shows a range of precision results for T-bLARS over 10 random partitions of columns into $P=128$ processors. We observe that T-bLARS still has a higher precision than bLARS in most cases. Determining the best column partitions that would yield the highest precision for T-bLARS in terms of column selection is interesting both in theory and in practice, but it is beyond the scope of this work.

\subsection{Speedup}

\begin{figure}[h]
	\centering
	\begin{subfigure}{.33\textwidth}
	\captionsetup{justification=centering}
  		\centering
  		\includegraphics[width=\textwidth]{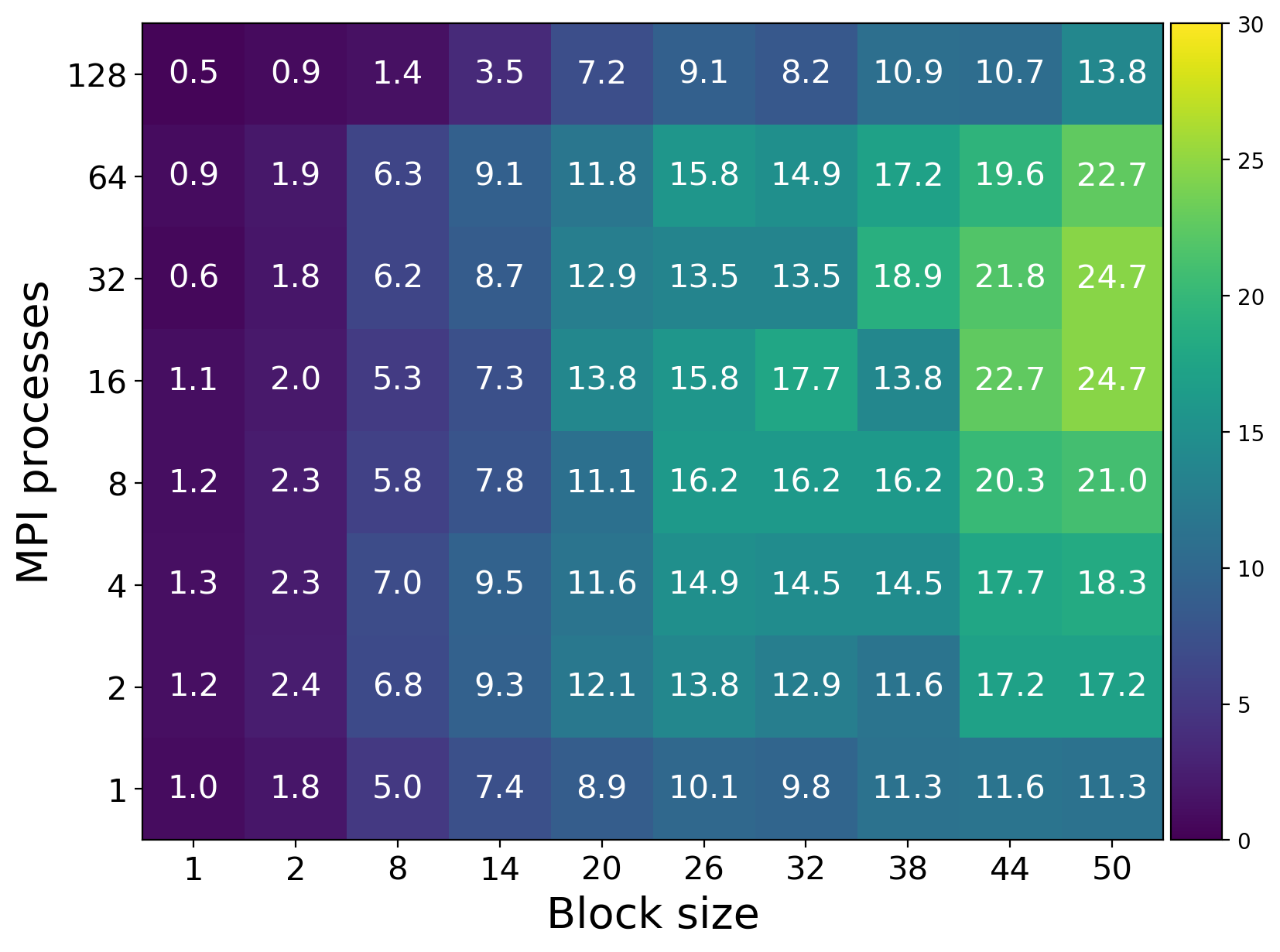}
		\caption{sector, bLARS}
	\end{subfigure}%
	\begin{subfigure}{.33\textwidth}
	\captionsetup{justification=centering}
  		\centering
  		\includegraphics[width=\textwidth]{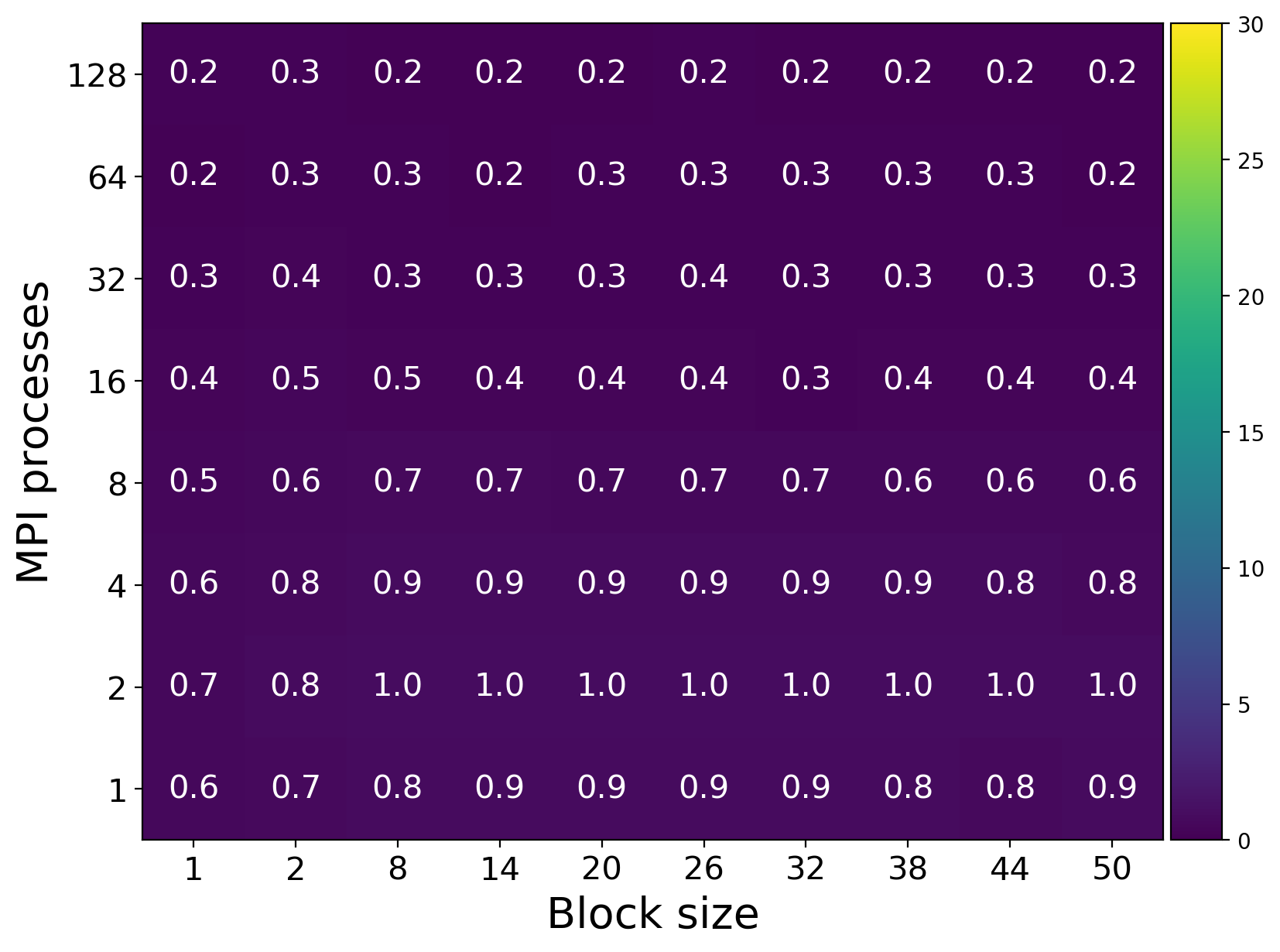}
		\caption{sector, T-bLARS}
	\end{subfigure}
	\begin{subfigure}{.33\textwidth}
	\captionsetup{justification=centering}
  		\centering
  		\includegraphics[width=\textwidth]{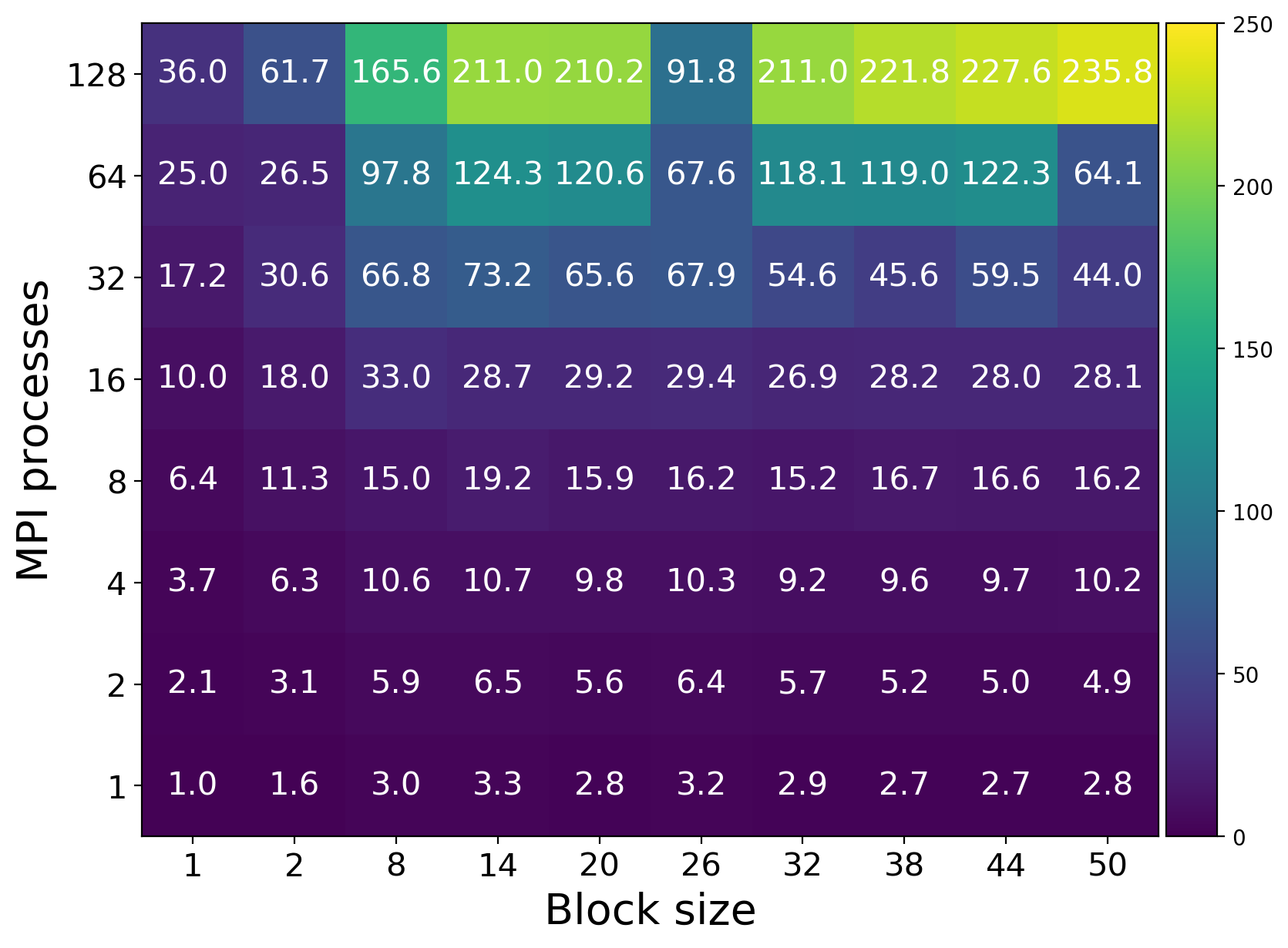}
		\caption{Year., bLARS}
	\end{subfigure}%
	\begin{subfigure}{.33\textwidth}
	\captionsetup{justification=centering}
  		\centering
  		\includegraphics[width=\textwidth]{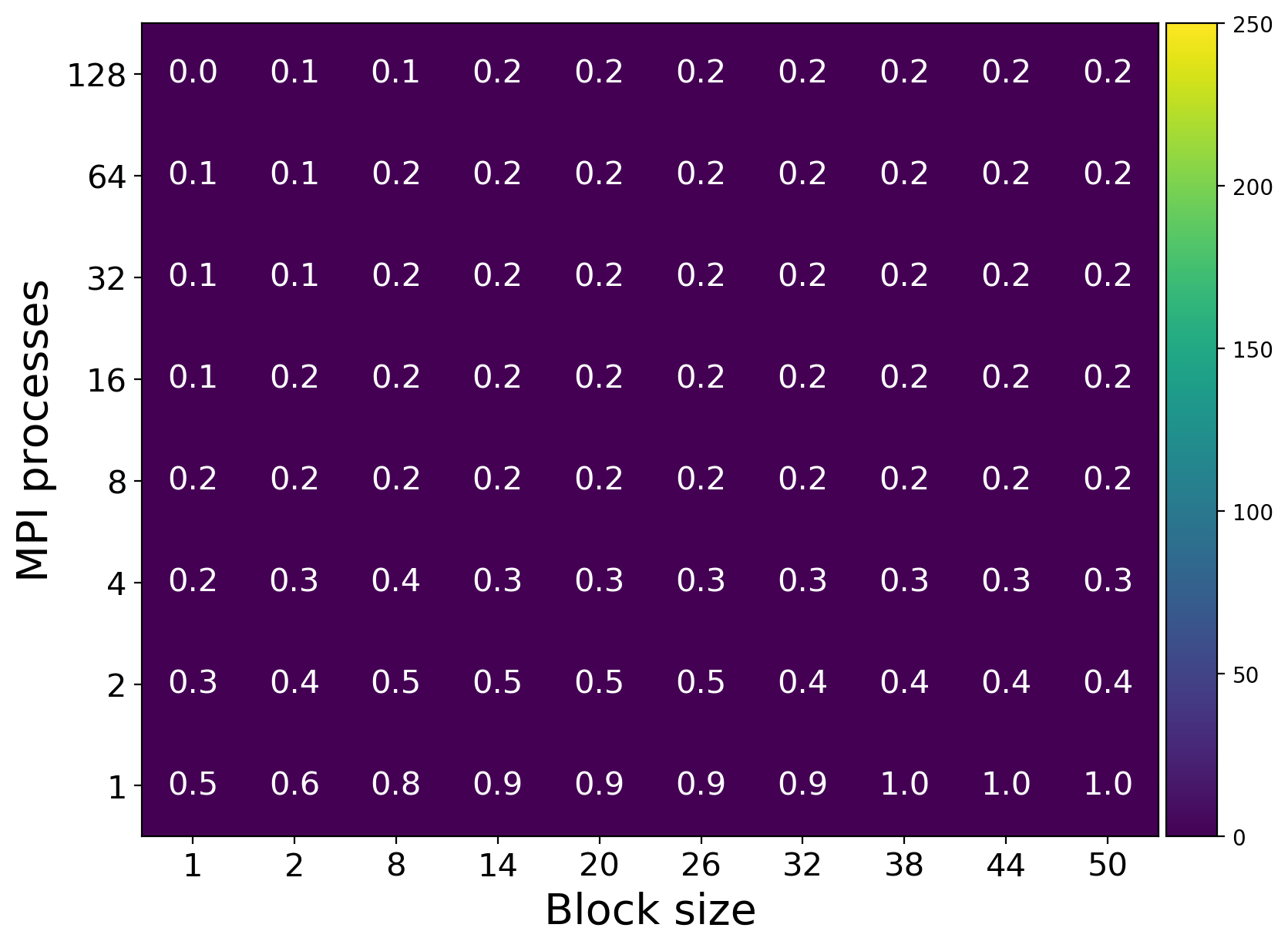}
		\caption{Year., T-bLARS}
	\end{subfigure}
	\begin{subfigure}{.33\textwidth}
	\captionsetup{justification=centering}
  		\centering
  		\includegraphics[width=\textwidth]{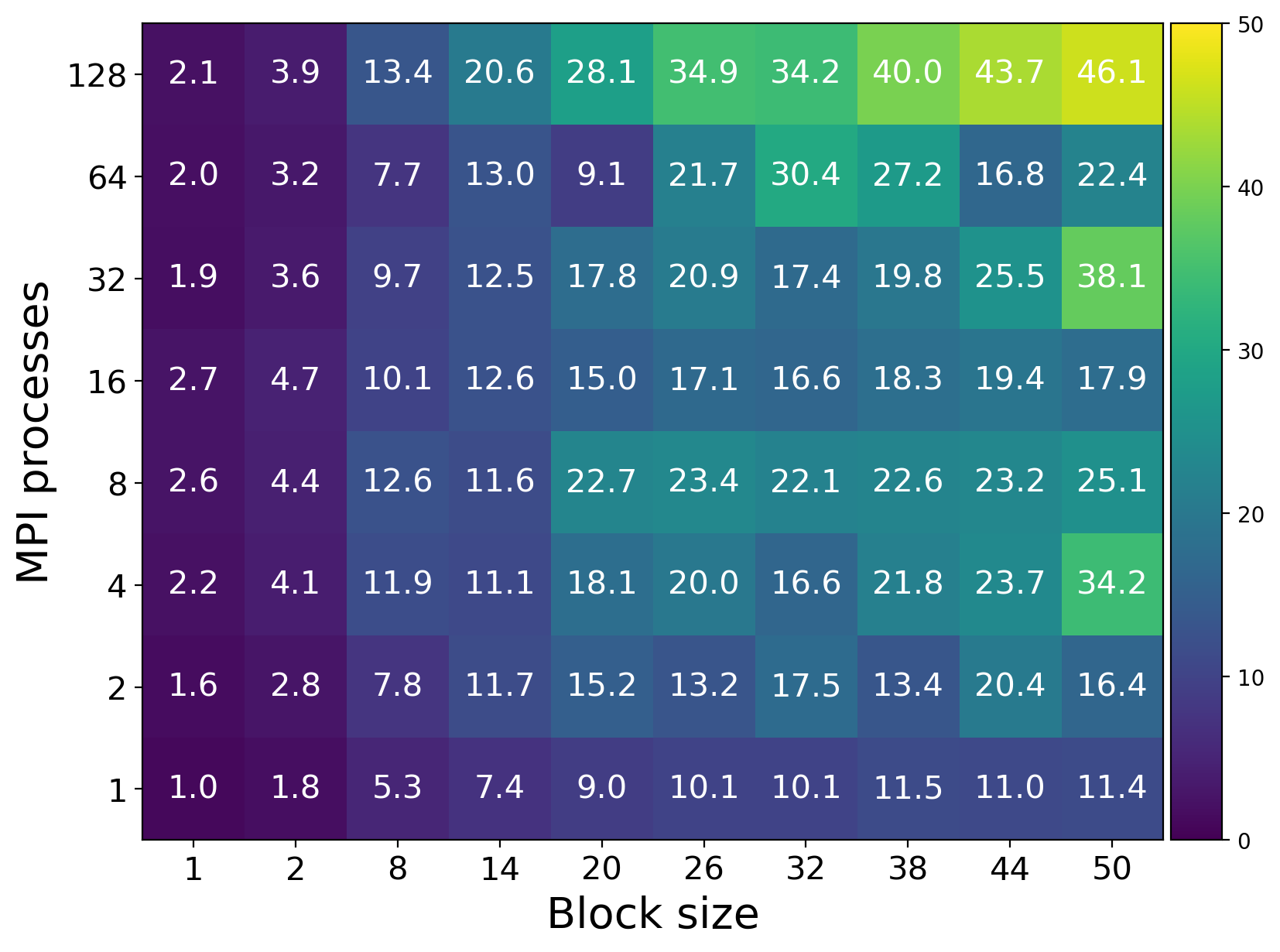}
		\caption{E2006\_tfidf, bLARS}
	\end{subfigure}%
	\begin{subfigure}{.33\textwidth}
	\captionsetup{justification=centering}
  		\centering
  		\includegraphics[width=\textwidth]{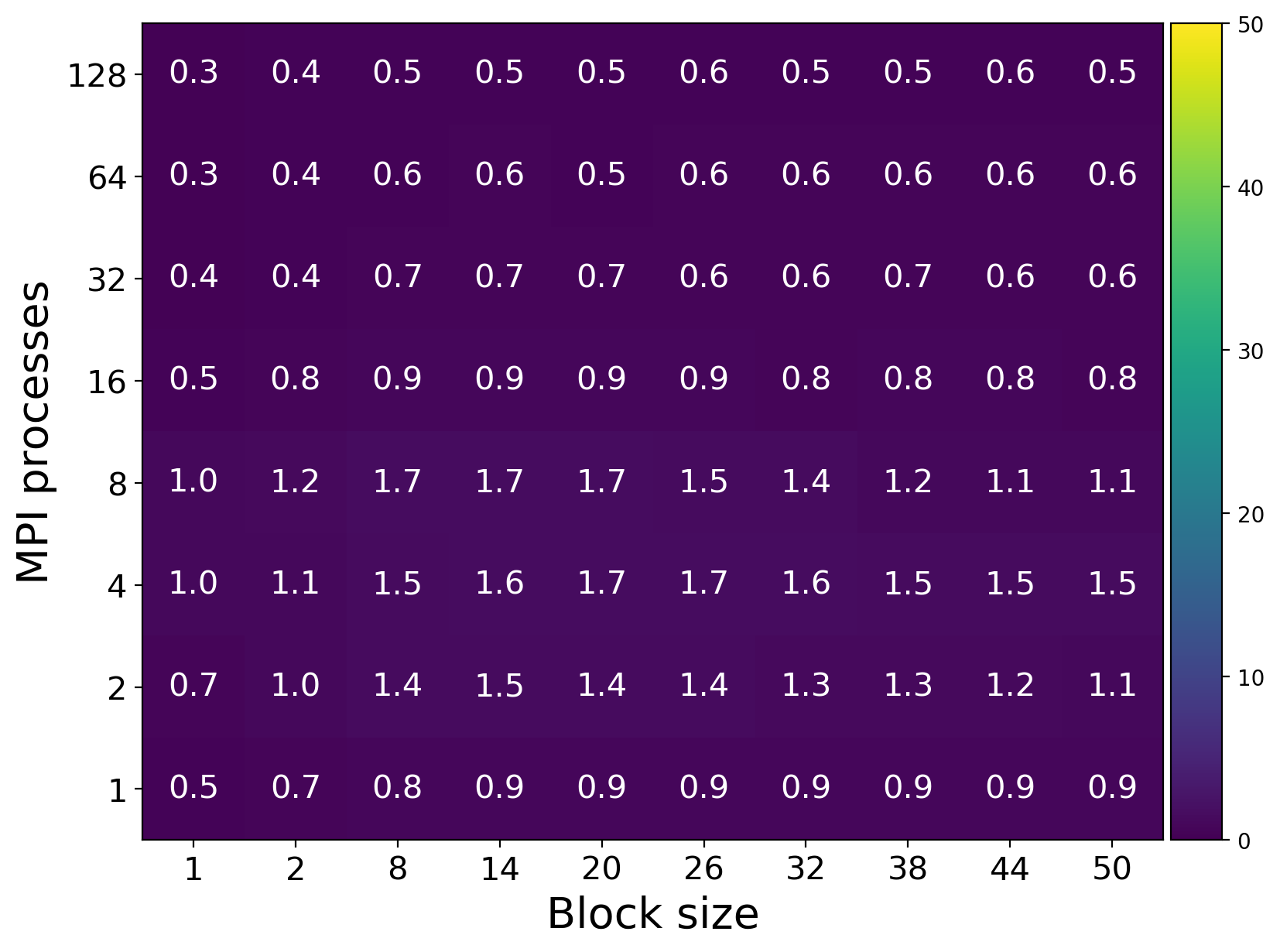}
		\caption{E2006\_tfidf, T-bLARS}
	\end{subfigure}
	\begin{subfigure}{.33\textwidth}
	\captionsetup{justification=centering}
  		\centering
  		\includegraphics[width=\textwidth]{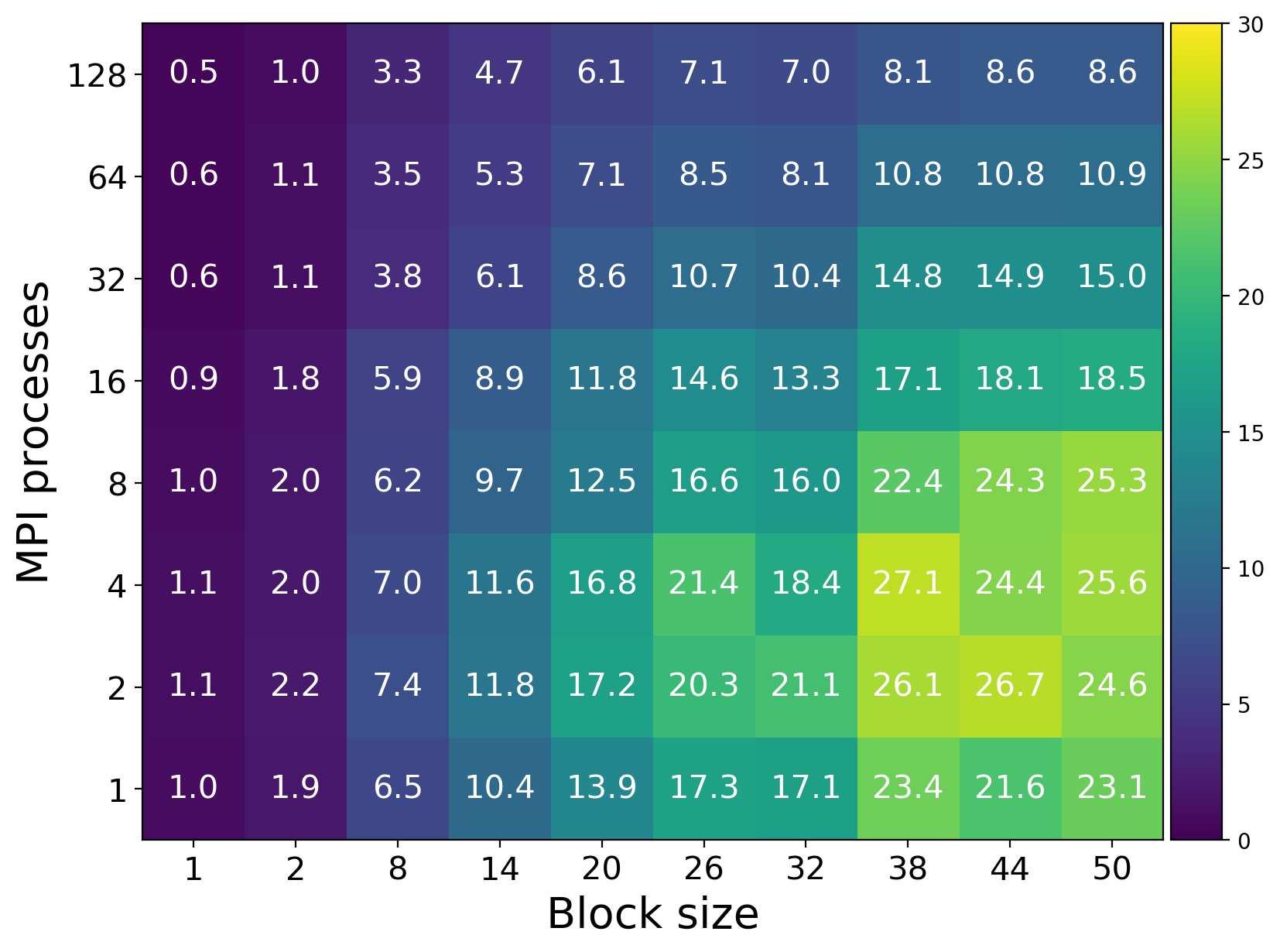}
		\caption{E2006\_log1p, bLARS}
	\end{subfigure}%
	\begin{subfigure}{.33\textwidth}
	\captionsetup{justification=centering}
  		\centering
  		\includegraphics[width=\textwidth]{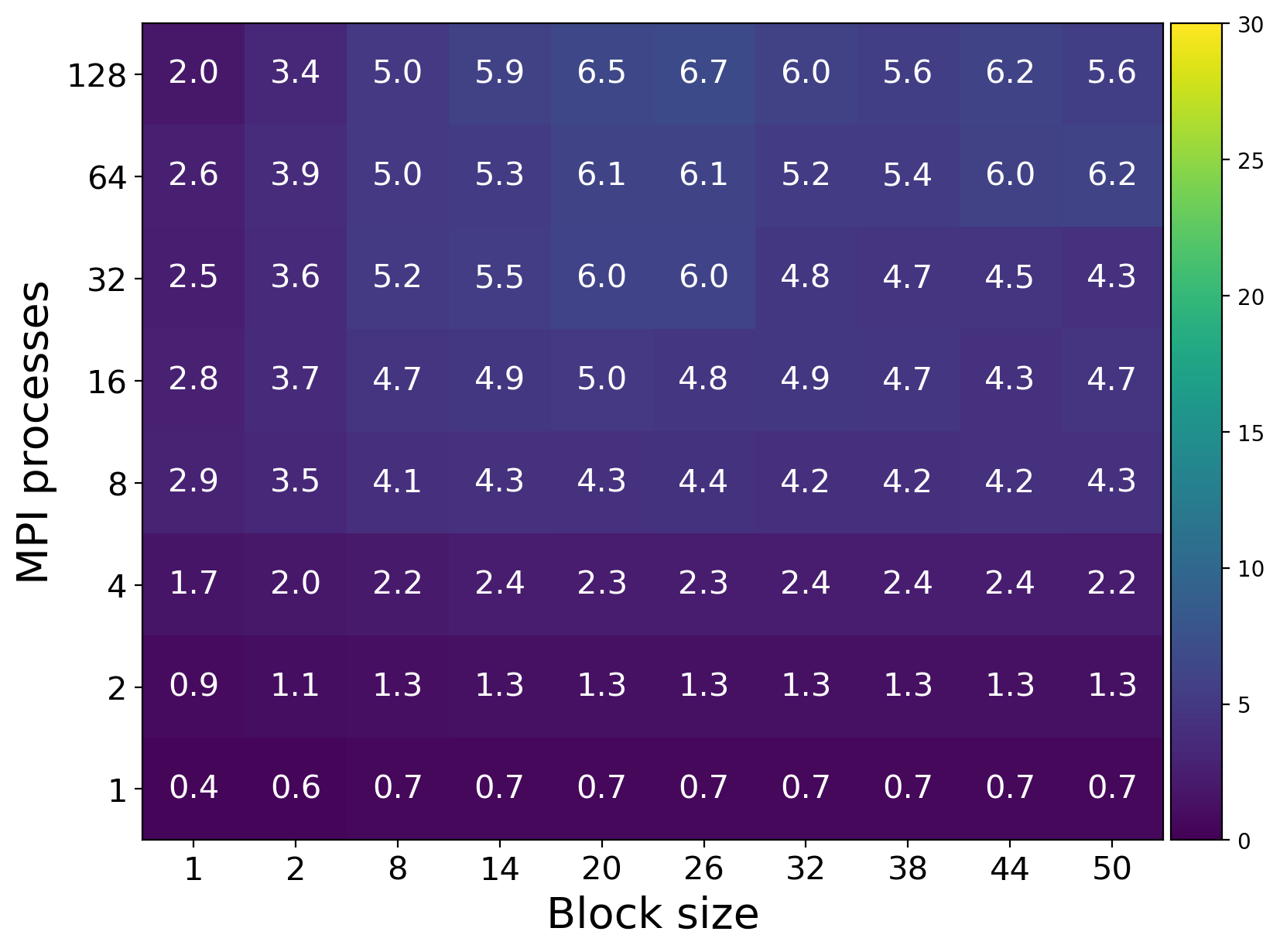}
		\caption{E2006\_log1p, T-bLARS}
	\end{subfigure}
	\caption{Total speedup.}
	\label{fig:speedup_real}
\end{figure}

\begin{figure}[h]
	\centering
	\begin{subfigure}{.245\textwidth}
	\captionsetup{justification=centering}
  		\centering
  		\includegraphics[width=\textwidth]{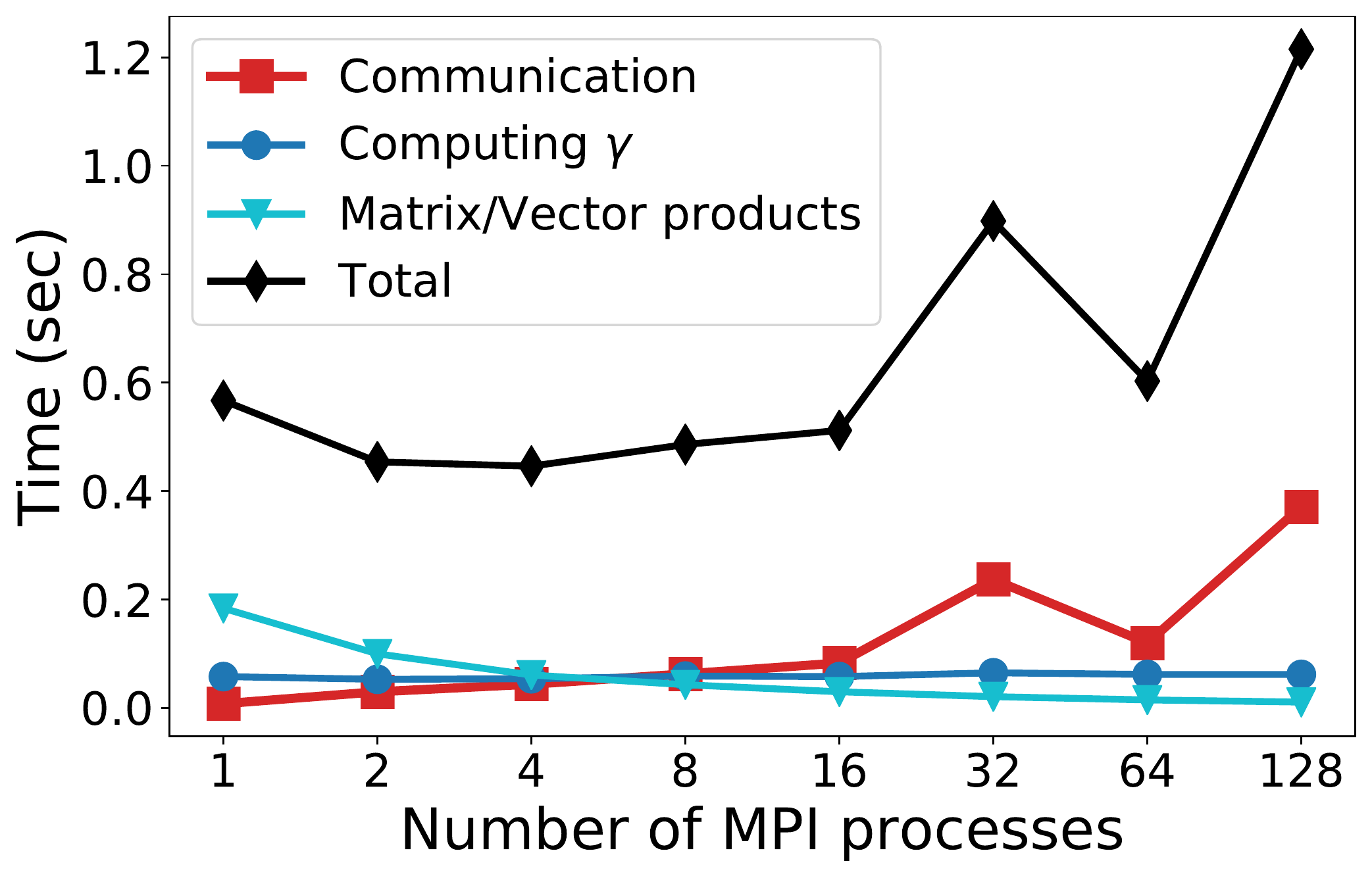}
		\caption{sector, bLARS}
	\end{subfigure}%
	\begin{subfigure}{.245\textwidth}
	\captionsetup{justification=centering}
  		\centering
  		\includegraphics[width=\textwidth]{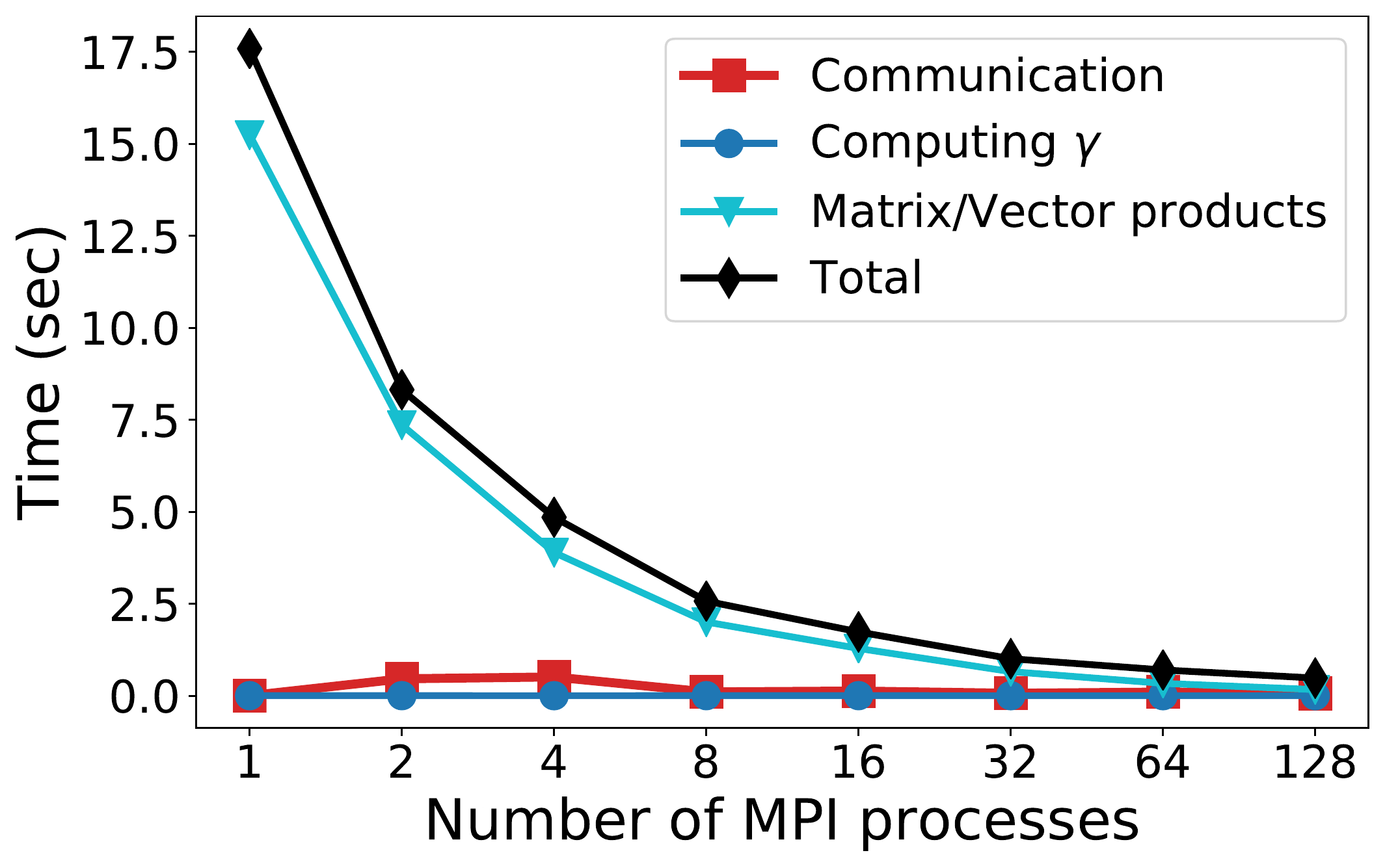}
		\caption{Year., bLARS}
	\end{subfigure}
	\begin{subfigure}{.245\textwidth}
	\captionsetup{justification=centering}
  		\centering
  		\includegraphics[width=\textwidth]{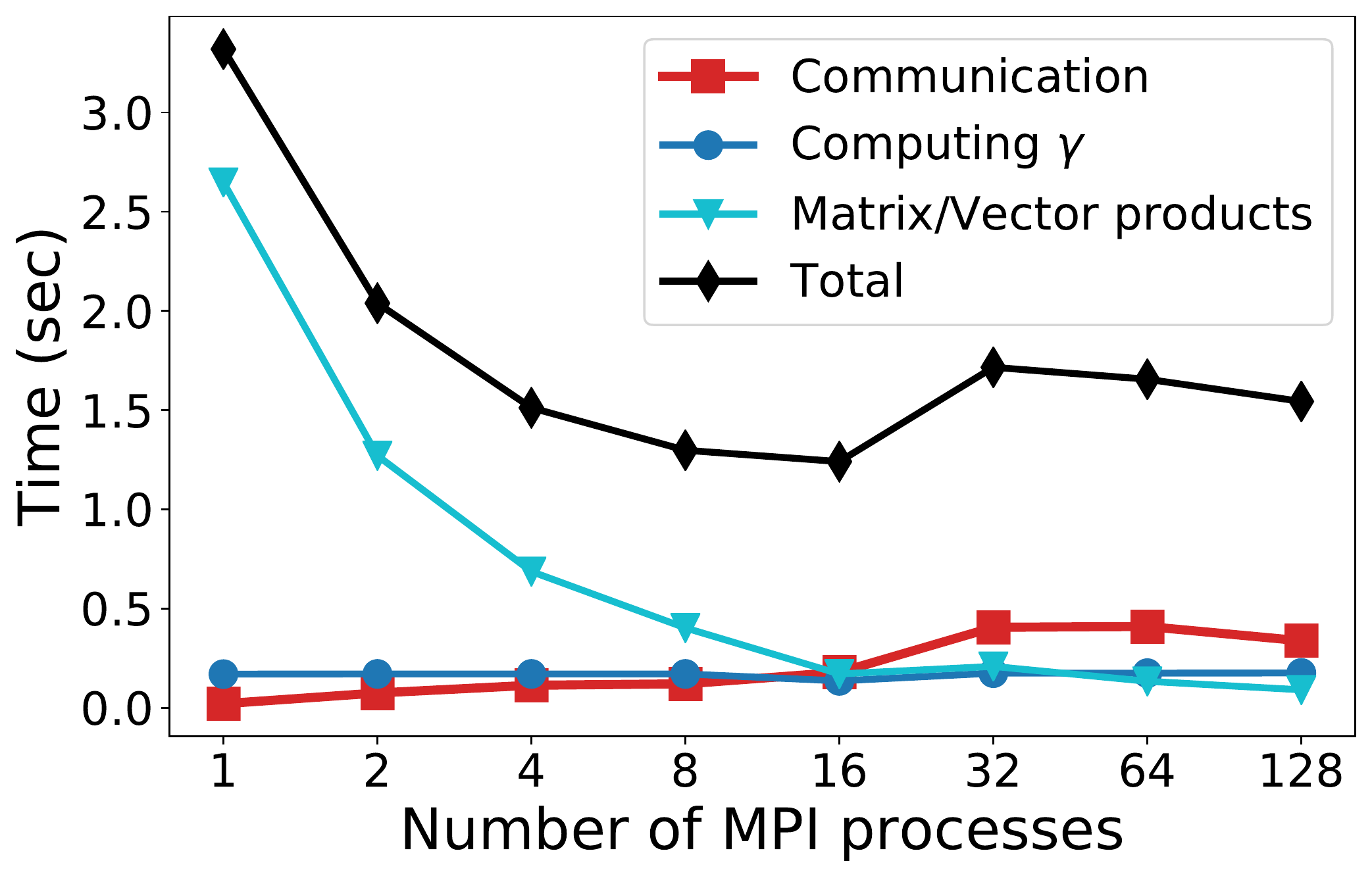}
		\caption{E2006\_tfidf, bLARS}
	\end{subfigure}
	\begin{subfigure}{.245\textwidth}
	\captionsetup{justification=centering}
  		\centering
  		\includegraphics[width=\textwidth]{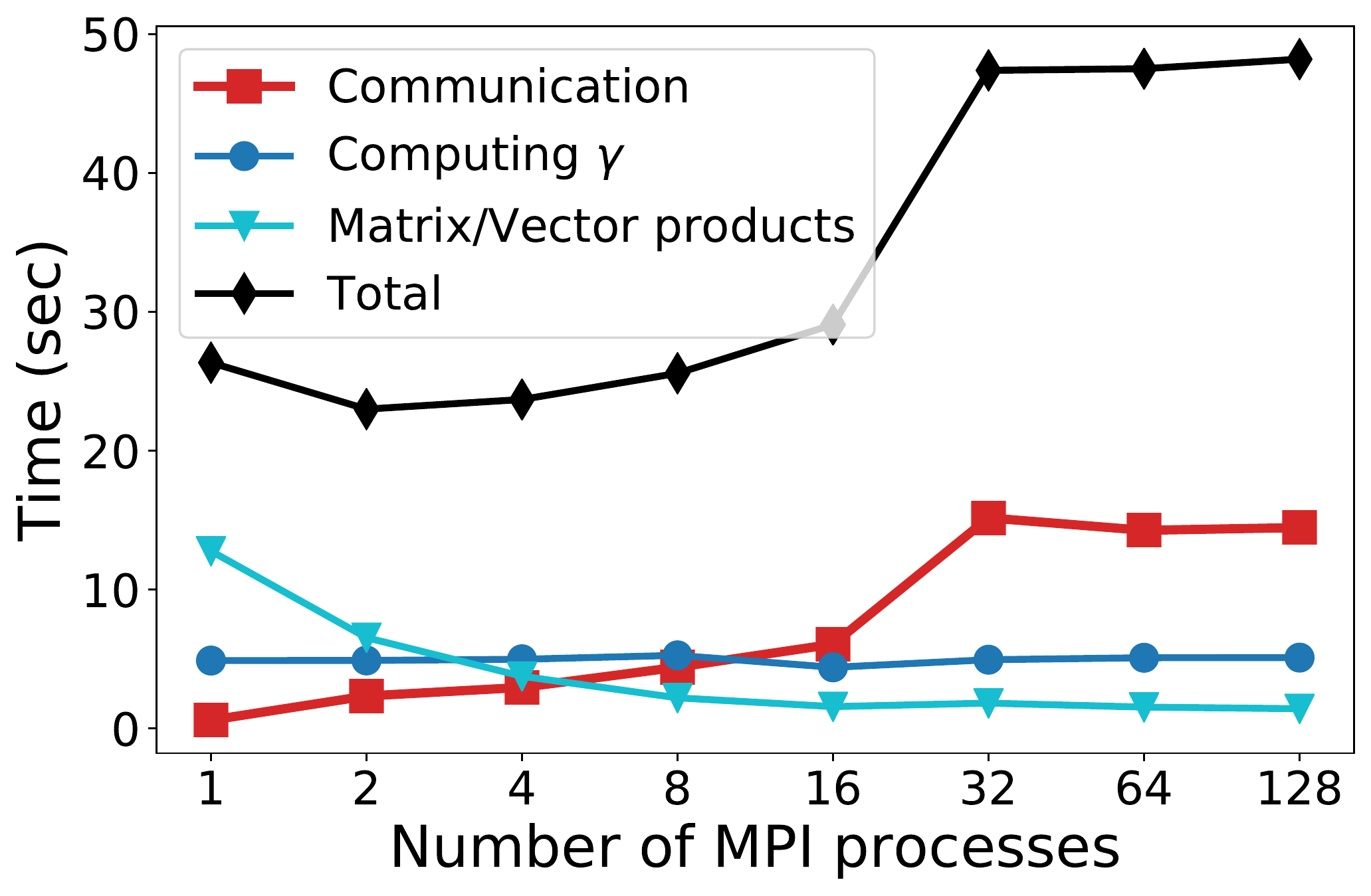}
		\caption{E2006\_log1p, bLARS}
	\end{subfigure}
	\begin{subfigure}{.245\textwidth}
	\captionsetup{justification=centering}
  		\centering
  		\includegraphics[width=\textwidth]{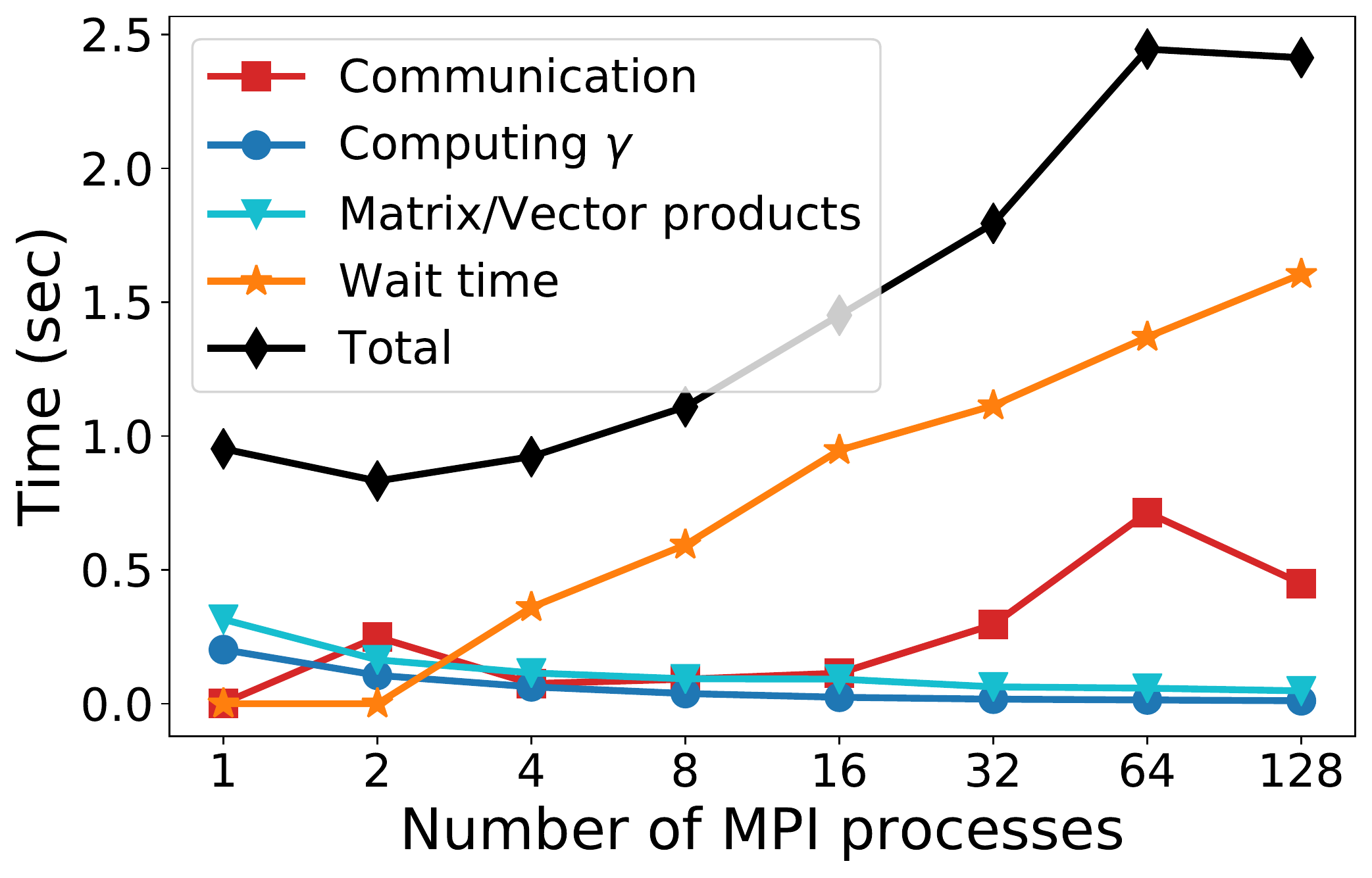}
		\caption{sector, T-bLARS}
	\end{subfigure}%
	\begin{subfigure}{.245\textwidth}
	\captionsetup{justification=centering}
  		\centering
  		\includegraphics[width=\textwidth]{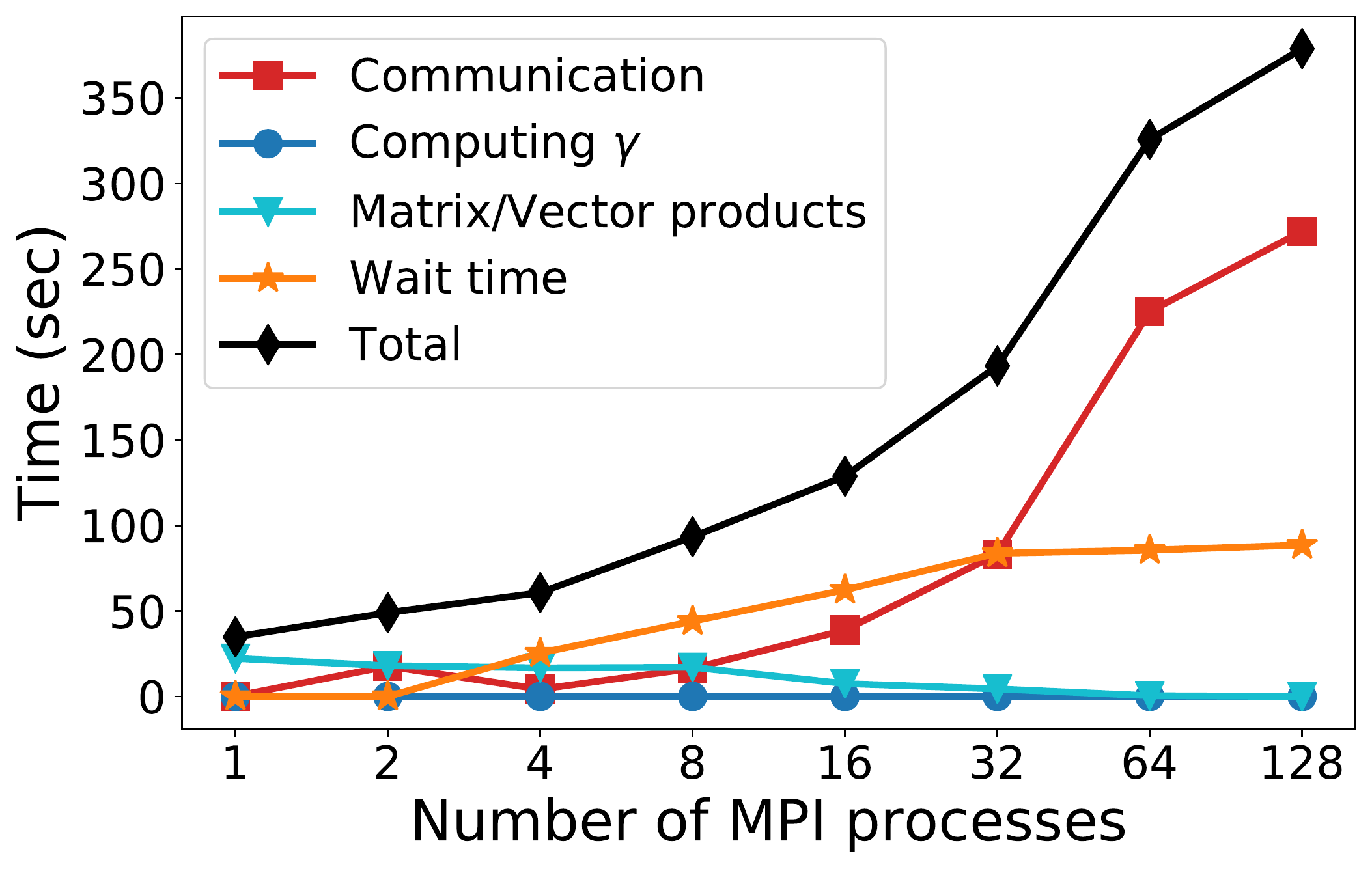}
		\caption{Year., T-bLARS}
	\end{subfigure}
	\begin{subfigure}{.245\textwidth}
	\captionsetup{justification=centering}
  		\centering
  		\includegraphics[width=\textwidth]{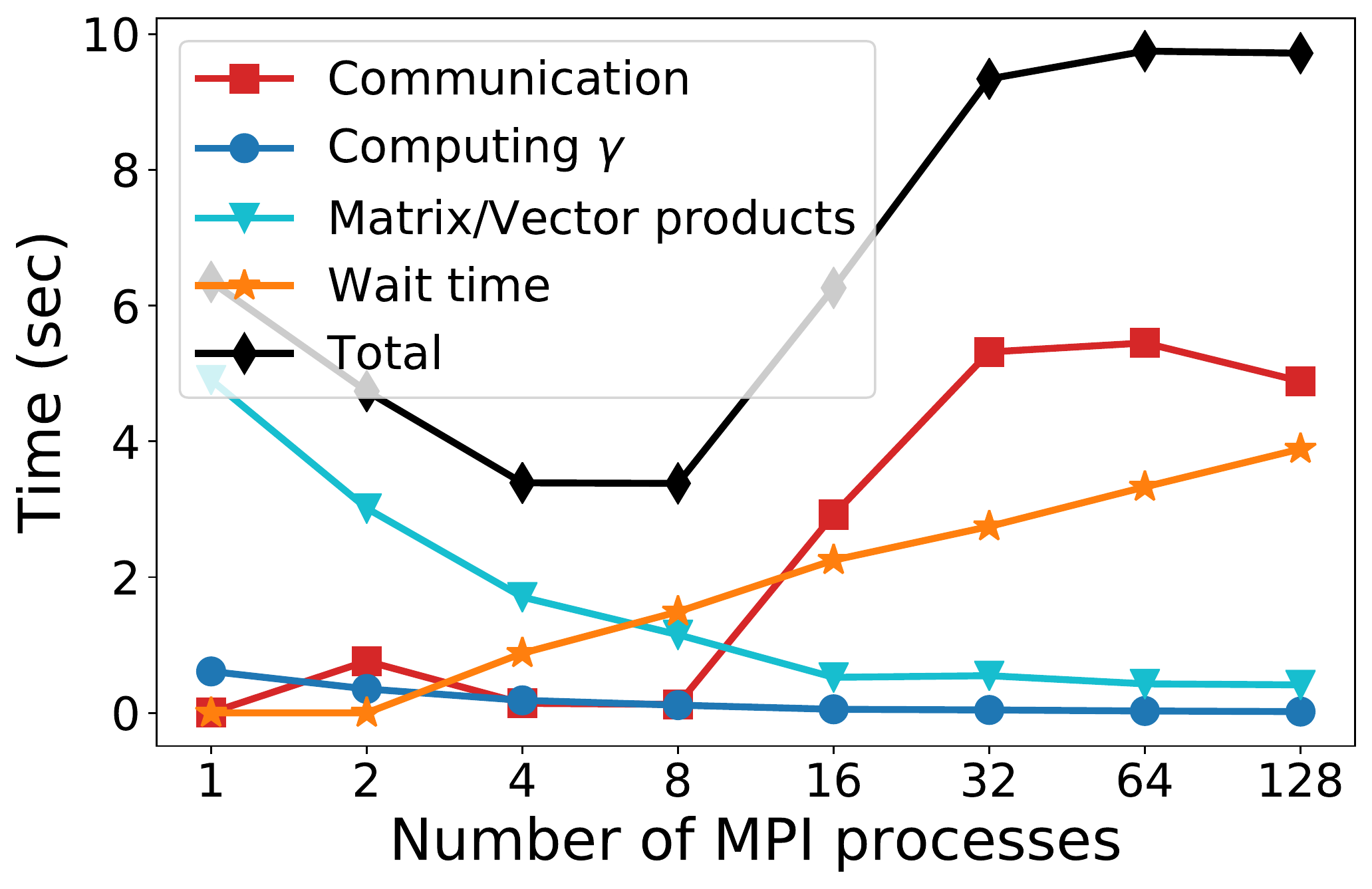}
		\caption{E2006\_tfidf, T-bLARS}
	\end{subfigure}
	\begin{subfigure}{.245\textwidth}
	\captionsetup{justification=centering}
  		\centering
  		\includegraphics[width=\textwidth]{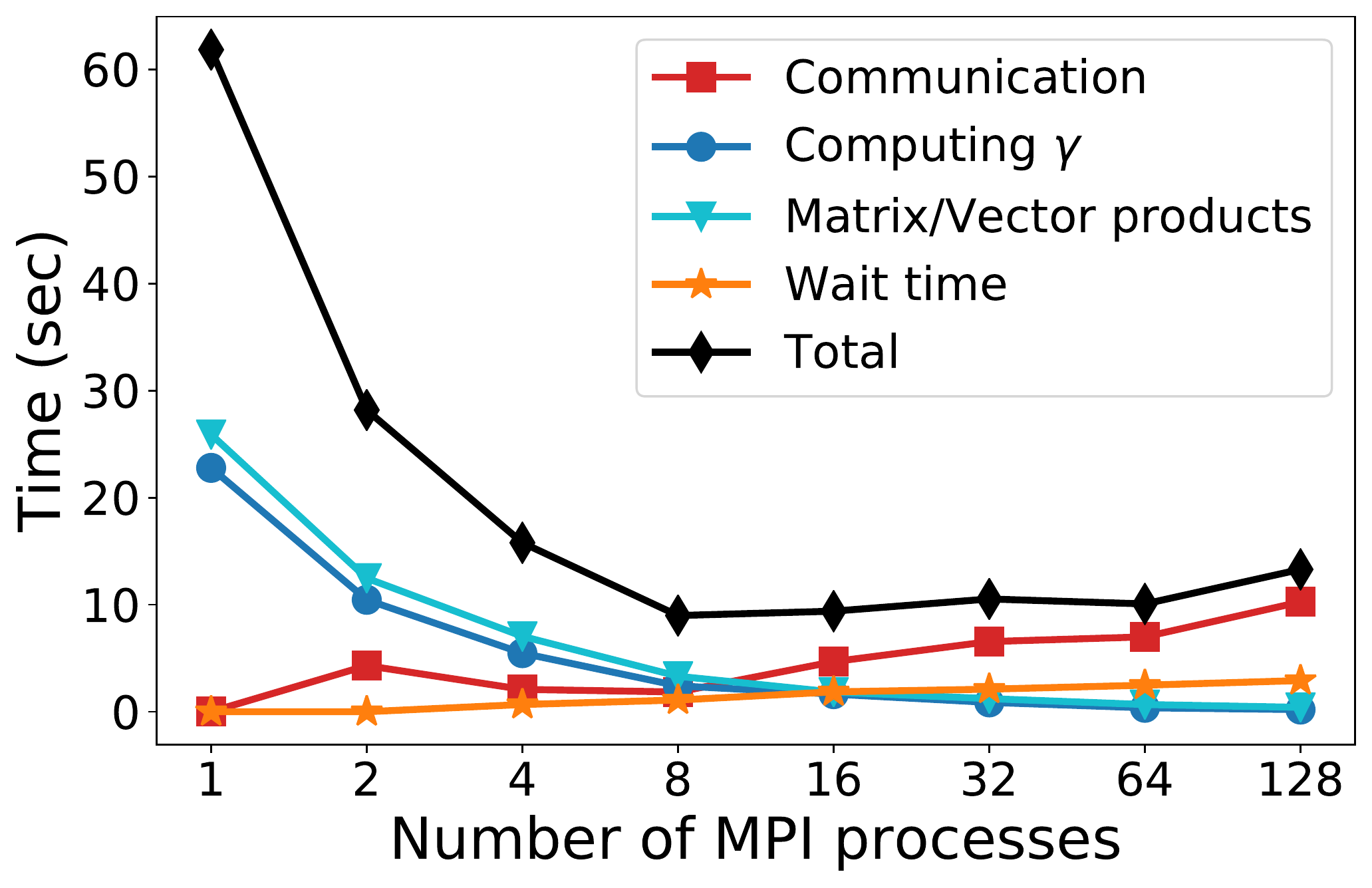}
		\caption{E2006\_log1p, T-bLARS}
	\end{subfigure}
	\caption{Running time breakdown. We fix $b=1$ and vary $P$. The pattern is similar for other $b$.}
	\label{fig:breakdown_real_fixb}
\end{figure}

\begin{figure}[h]
	\centering
	\begin{subfigure}{.245\textwidth}
	\captionsetup{justification=centering}
  		\centering
  		\includegraphics[width=\textwidth]{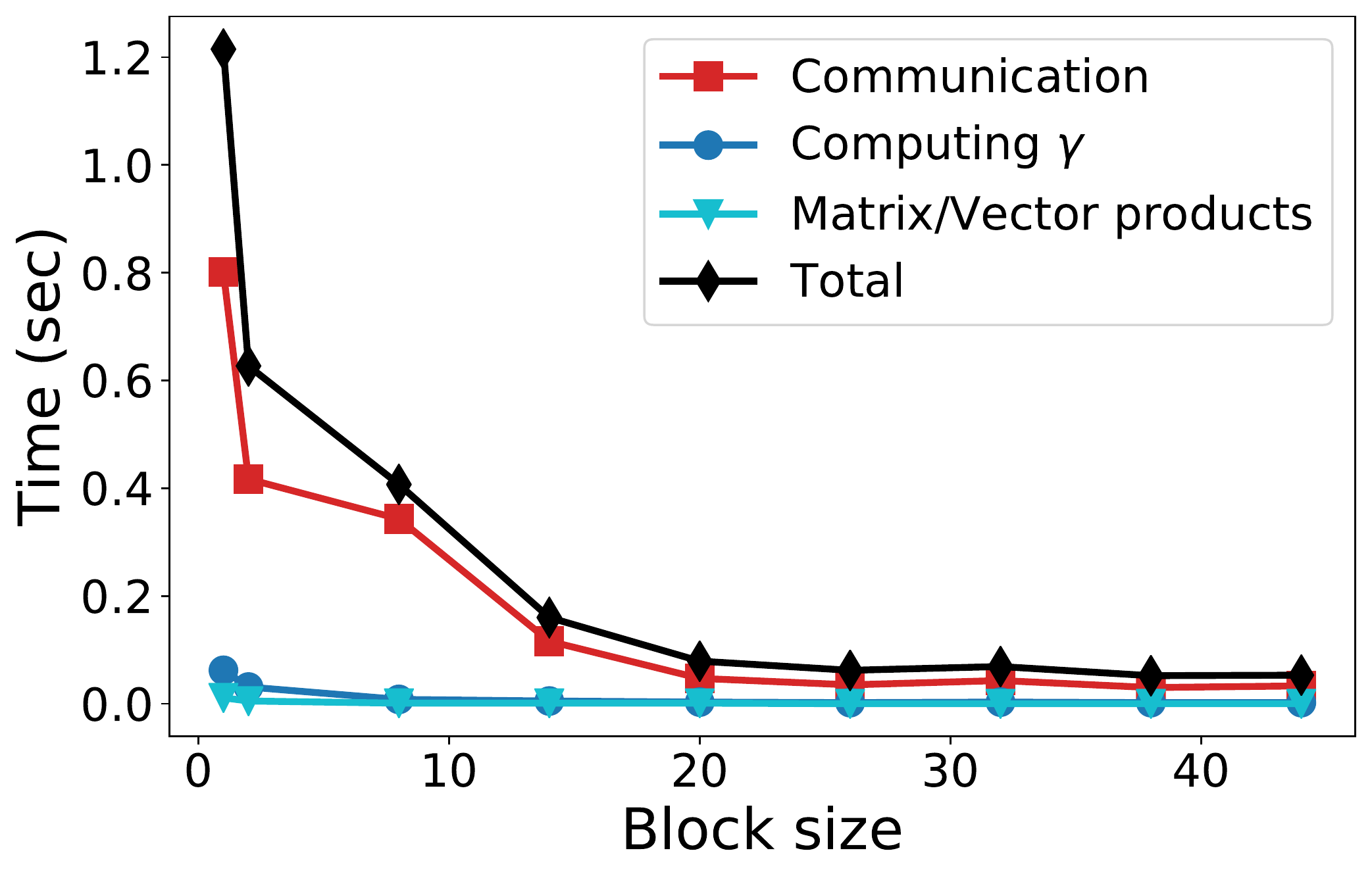}
		\caption{sector, bLARS}
	\end{subfigure}%
	\begin{subfigure}{.245\textwidth}
	\captionsetup{justification=centering}
  		\centering
  		\includegraphics[width=\textwidth]{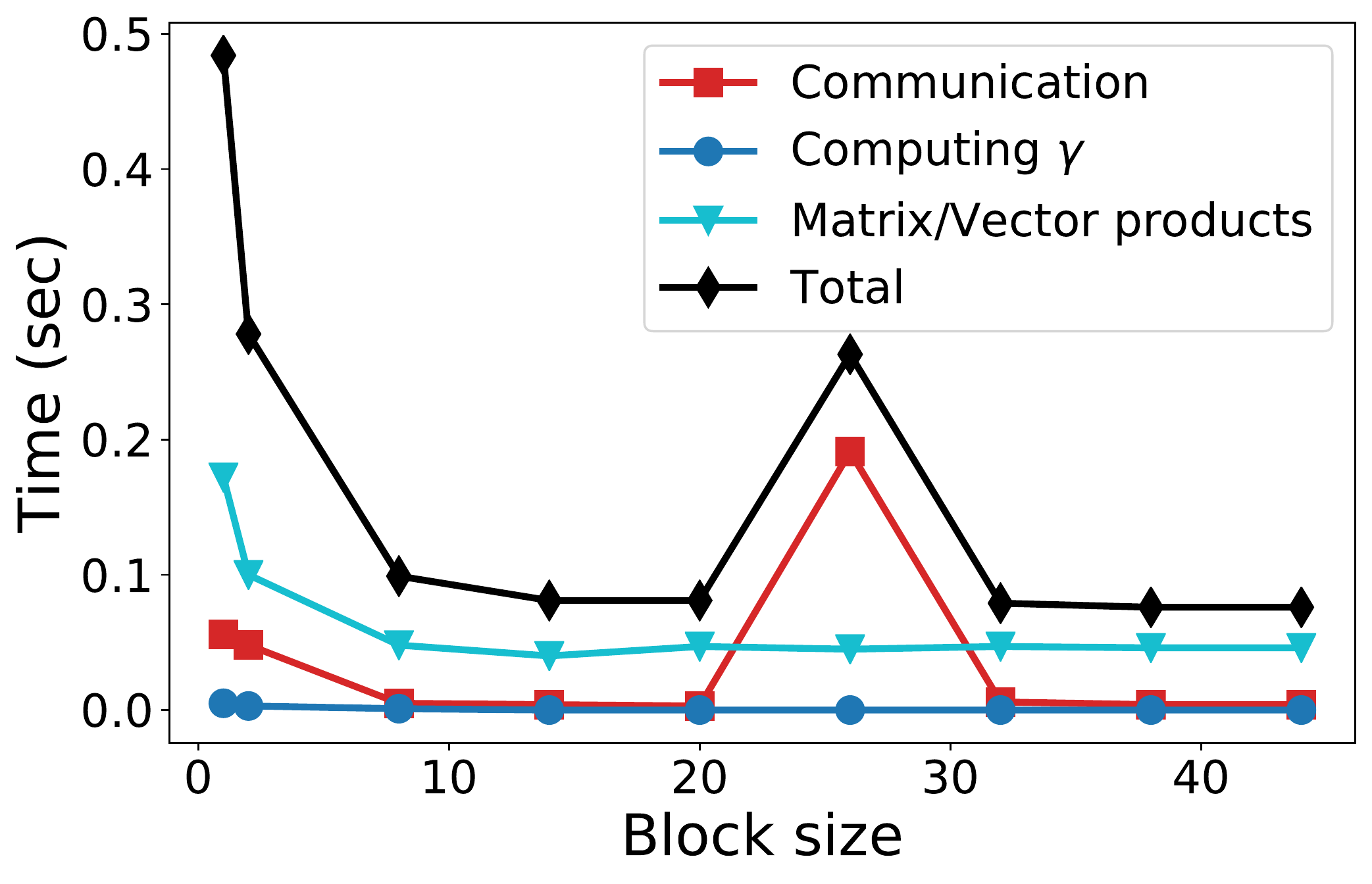}
		\caption{Year., bLARS}
	\end{subfigure}
	\begin{subfigure}{.245\textwidth}
	\captionsetup{justification=centering}
  		\centering
  		\includegraphics[width=\textwidth]{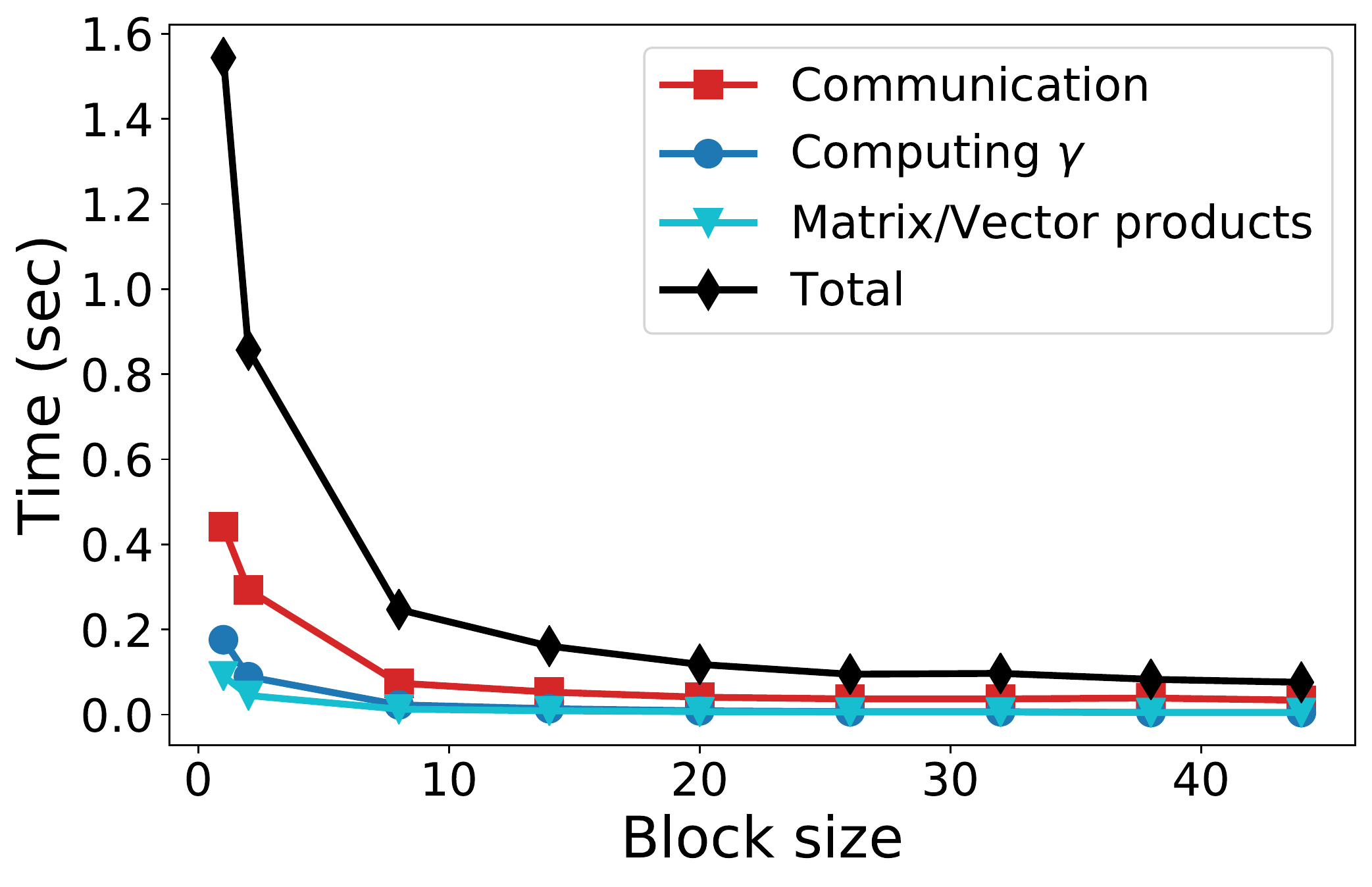}
		\caption{E2006\_tfidf, bLARS}
	\end{subfigure}
	\begin{subfigure}{.245\textwidth}
	\captionsetup{justification=centering}
  		\centering
  		\includegraphics[width=\textwidth]{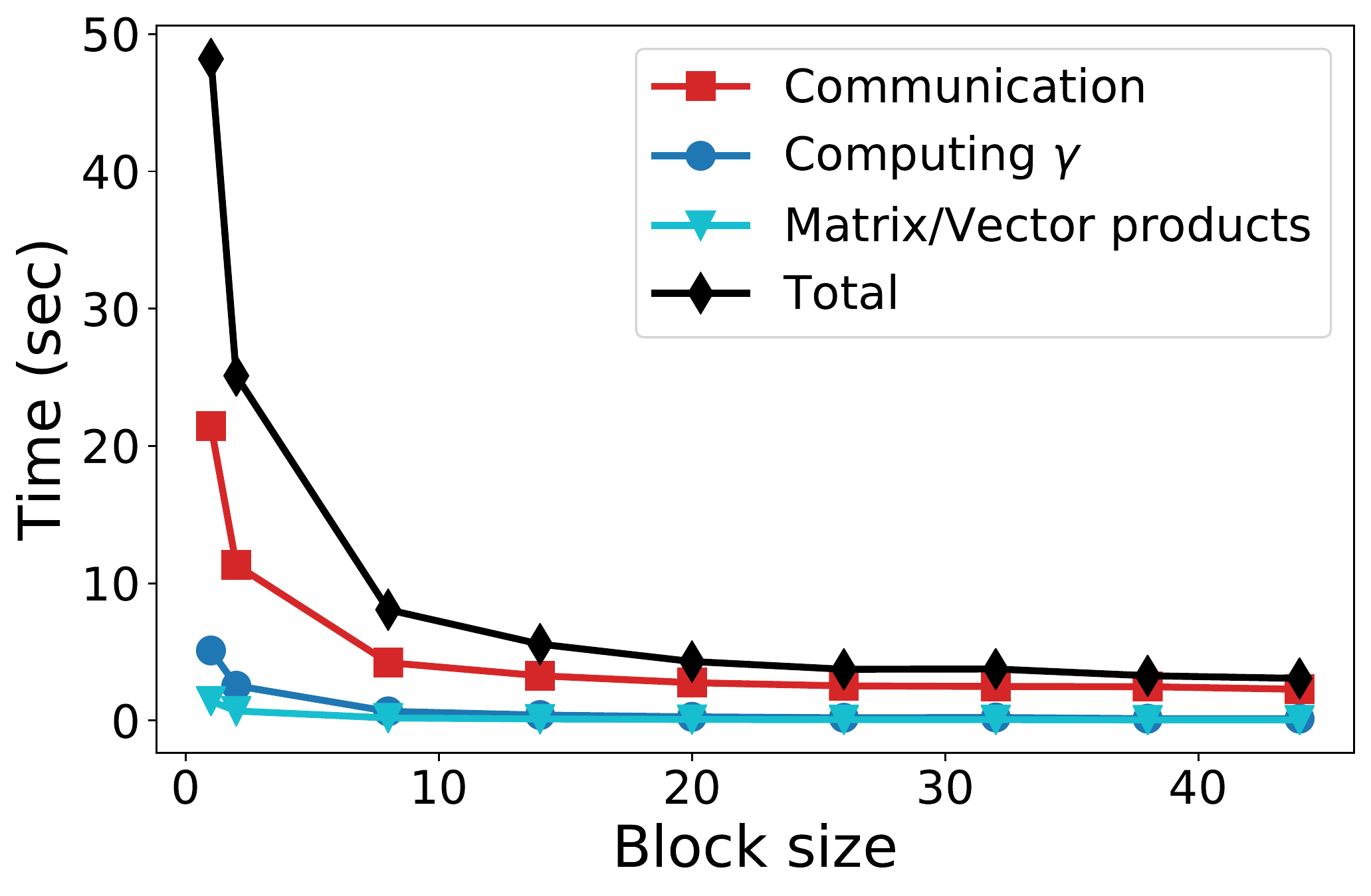}
		\caption{E2006\_log1p, bLARS}
	\end{subfigure}
	\begin{subfigure}{.245\textwidth}
	\captionsetup{justification=centering}
  		\centering
  		\includegraphics[width=\textwidth]{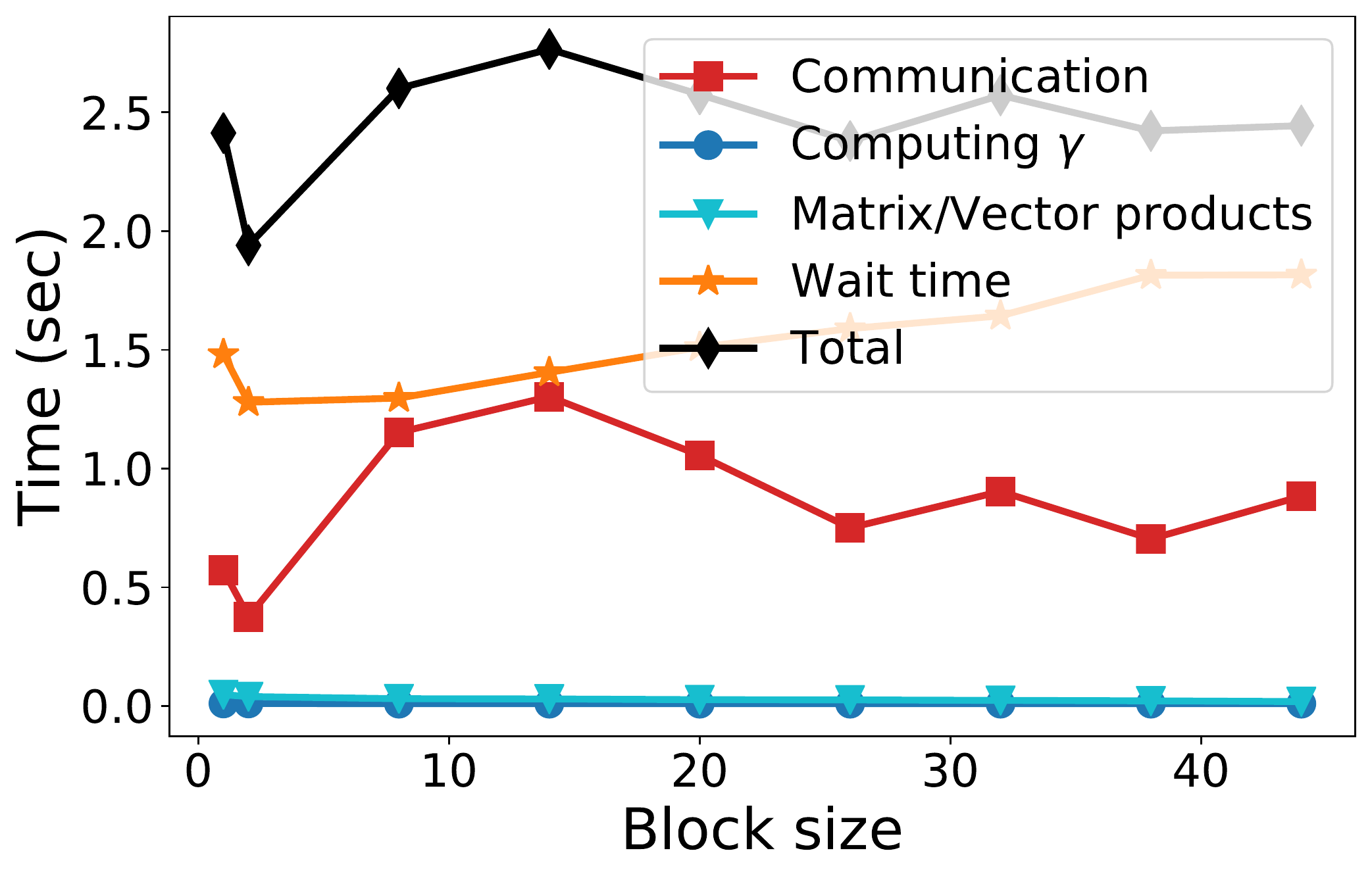}
		\caption{sector, T-bLARS}
	\end{subfigure}%
	\begin{subfigure}{.245\textwidth}
	\captionsetup{justification=centering}
  		\centering
  		\includegraphics[width=\textwidth]{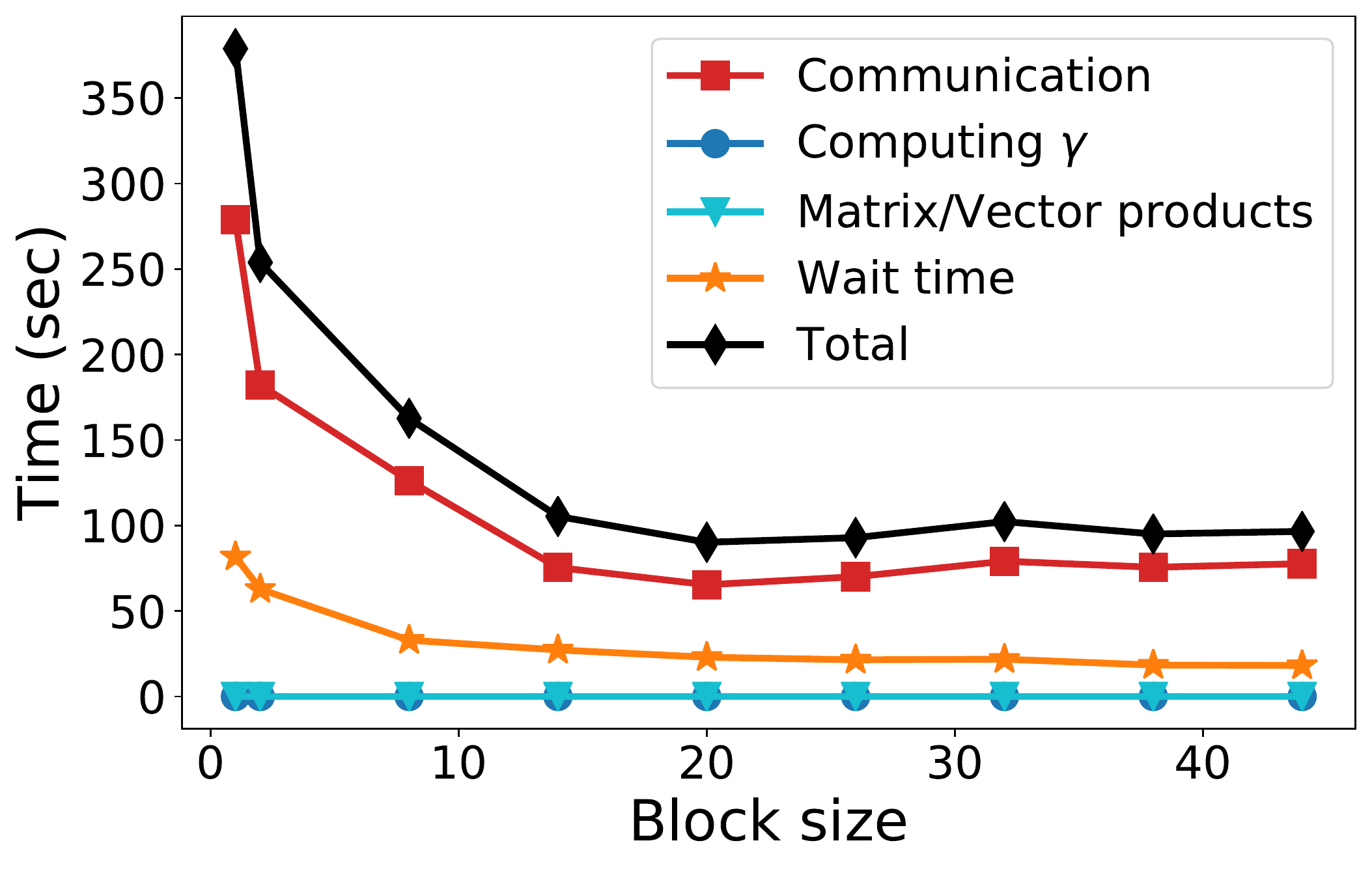}
		\caption{Year., T-bLARS}
	\end{subfigure}
	\begin{subfigure}{.245\textwidth}
	\captionsetup{justification=centering}
  		\centering
  		\includegraphics[width=\textwidth]{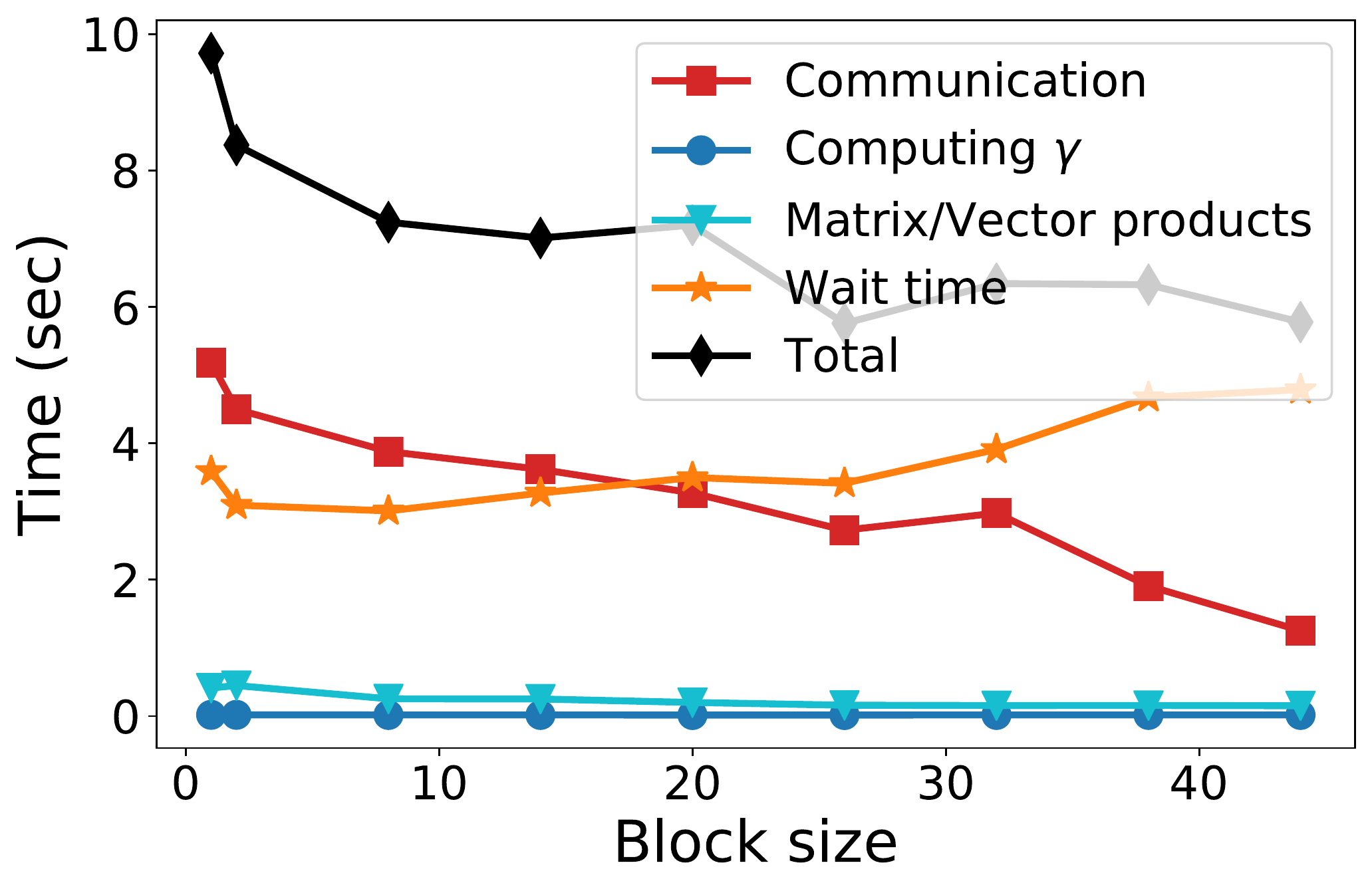}
		\caption{E2006\_tfidf, T-bLARS}
	\end{subfigure}
	\begin{subfigure}{.245\textwidth}
	\captionsetup{justification=centering}
  		\centering
  		\includegraphics[width=\textwidth]{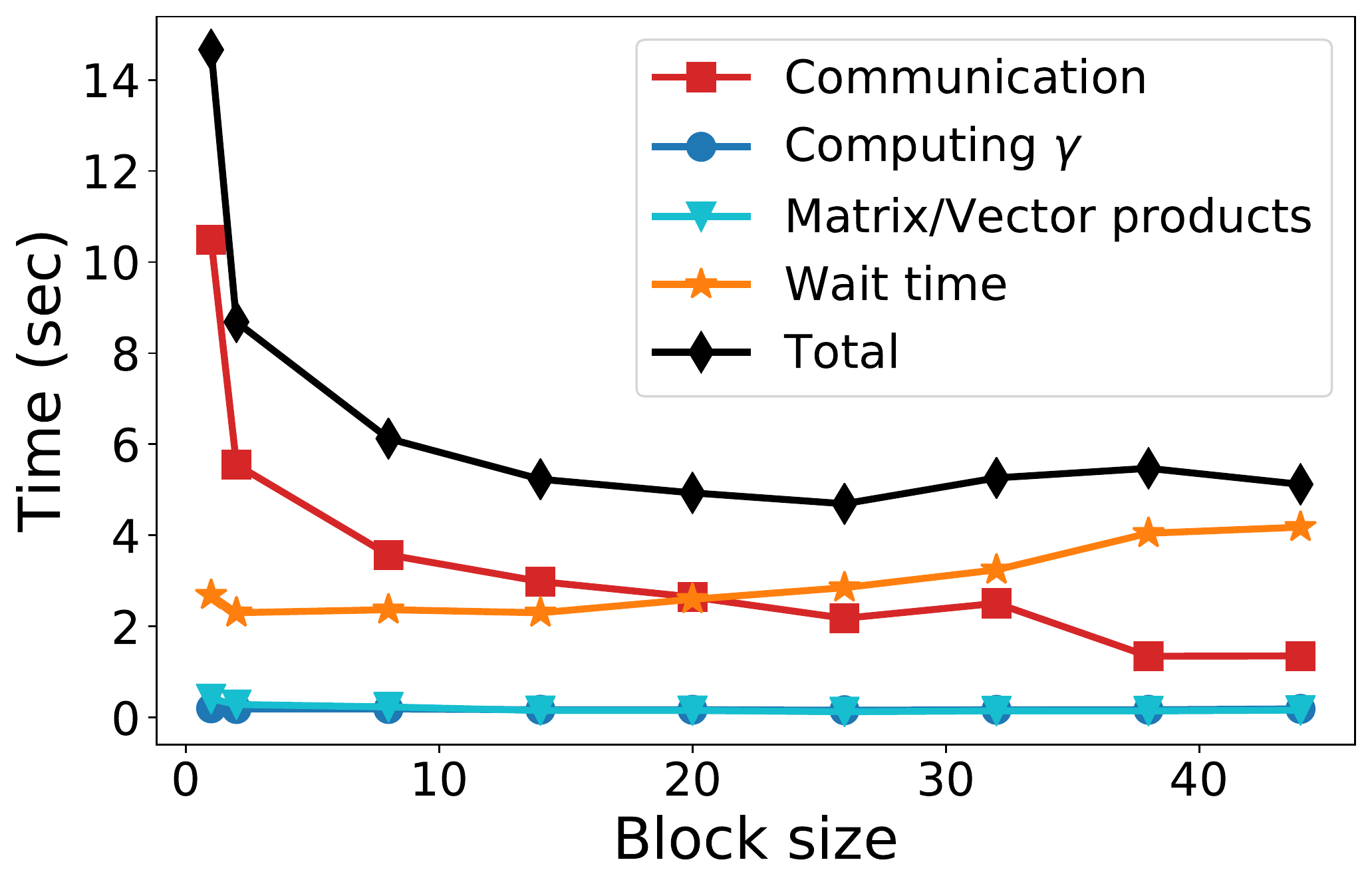}
		\caption{E2006\_log1p, T-bLARS}
	\end{subfigure}
	\caption{Running time breakdown. We fix $P=128$ and vary $b$. The pattern is similar for other $P$.}
	\label{fig:breakdown_real_fixP}
\end{figure}

We show the speedup trends in Figure~\ref{fig:speedup_real}. Note that for $P = b = 1$, the speedup factor for T-bLARS is not identically $1.0$ because T-bLARS performs more matrix-vector products than LARS in this parameter setting. For example, T-bLARS re-computes $\cc_k$ repeatedly due to iterative call to mLARS (Step 4), while in LARS, the vector $\cc_k$ is computed only once and updated iteratively. Overall, bLARS enjoys much higher speedups across all datasets. When the data is not very high-dimensional, i.e., not in the regime $n \gg m$, the total running time of bLARS scales with both $P$ and $b$ as predicted by the asymptotic costs analysis. The largest dataset E2006\_log1p has way more columns than rows, and bLARS slows down when we increase the number of processors beyond 4. On the other hand, apart from E2006\_log1p, T-bLARS does not seem to have a good speedup on other datasets. In order to understand what causes the speedups or the slow-downs, in Figure~\ref{fig:breakdown_real_fixb} (resp. Figure~\ref{fig:breakdown_real_fixP}) we fix $b$ (resp. $P$) and vary $P$ (resp. $b$) and show how the major components of the total running time scales. For arithmetic operations, we plot the time spent on performing matrix-matrix and matrix-vector products and the time spent on computing the step size $\gamma$ separately, as both the cost analysis (cf. Table~\ref{tab:costs_blars}) and subsequent plots show that these are the computation bottlenecks. There is only a very small fraction of the total time spent on other computations, e.g., scalar multiplications, array initializations, and Cholesky factorization and inversion of small-size matrices, so we do not plot all of them explicitly. Note that the binary tree reduction in T-bLARS has $\log P$ serial levels: for a column to become a winner at the root, it has to go through $\log P$ number of competitions sequentially. Therefore, once the candidate columns are selected at leaf nodes and competitions start at non-leaf nodes, there will always be some nodes waiting for the root to broadcast the final winners before starting the next iteration. For T-bLARS we include this wait time in the running time breakdown plots. We estimated the wait time using the average computation time per competition at non-leaf nodes times the number of levels in the tree.

We make some comments about Figure~\ref{fig:breakdown_real_fixb} and Figure~\ref{fig:breakdown_real_fixP}. First, both bLARS and T-bLARS reduce the time spent on matrix-vector products as we increase either $P$ or $b$. The speedup of bLARS mainly comes from the speedup of matrix-vector products. Second, bLARS spent smaller fraction of total time on communication when the data matrix is tall $m \gg n$, e.g., YearPredictionMSD; T-bLARS spent smaller fraction of total time on communication when the data matrix in fat $n \gg m$. This is expected because the number of words communicated for bLARS increases with $n$ and is independent of $m$, while the number of words communicated for T-bLARS increases with $m$ and is independent of $n$. Third, we didn't see a good speedup of T-bLARS for sector, YearPredictionMSD and E2006\_tfidf, because T-bLARS spent a large fraction of time on serial reduction in the binary tree, which overweighs the reduction in time for matrix-vector products. On the other hand, the wait time for serial tournaments for E2006\_log1p took relatively much less time, so T-bLARS obtains good speedups. In general, one can expect T-bLARS to have a good speedup when the ``wait time'' is much less than parallel computation times (e.g., matrix-vector products) at leaf nodes. Our implementation of T-bLARS uses sparse data structures for computations at leaf nodes (to reduce memory requirement) and dense data structures for computations at non-leaf nodes (to reduce overheads). This has put T-bLARS in a slight disadvantage when dealing with sparse data as many arithmetic operations at non-leaf nodes will be unnecessary. We thus expect T-bLARS to achieve better speedups (than the 6x on E2006\_log1p) on dense and high-dimensional data where $n \gg m$. Finally, Figure~\ref{fig:breakdown_real_fixP} shows that the communication cost of both bLARS and T-bLARS tends to decrease as $b$ increases, which is also expected according to Table~\ref{tab:asymptcosts}.

Our experiments indicate that there is a tradeoff between bLARS and T-bLARS. On one side, bLARS is well suited for row-partitioned data and can achieve speedups up to two orders of magnitude. However, the amazing speedup of bLARS comes at the expense of solution quality. One the other side, while T-bLARS is generally slower than bLARS, it has lower residual norms and on average selects columns more accurately than bLARS. For example, for E2006\_log1p, T-bLARS achieves 4x speedup ($P = 64$, $b=2$) while correctly selecting 100\% columns, bLARS only obtains 2x speedup for $b=2$ and has a precision below 80\%. Even though bLARS has up to 27x speedup ($P=4$, $b=38$) for E2006\_log1p, in this setting bLARS only correctly recovers around 30\% columns that would have been selected by LARS.

\section{Conclusions}\label{sec:conclusions}

The two parallel and communication-avoiding methods we have introduced, bLARS and T-bLARS, present valuable methods of least-angle regression that provide higher performance of speed than LARS can normally give.\ 
The choice between the two comes down what priorities and expectations the user has from the solutions generated from these algorithms, e.g., be it higher speed or more resilient accuracy.\

\clearpage

\bibliographystyle{plain}
\bibliography{references}

\begin{thebibliography}{10}

\bibitem{AISS97}
A.~Alexandrov, M.~F. Ionescu, K.~E. Schauser, and C.~Scheiman.
\newblock Log{GP}: {I}ncorporating long messages into the log{P} model for
  parallel computation.
\newblock {\em Journal of parallel and distributed computing}, 44(1):71--79,
  1997.

\bibitem{Ballard13}
G.~Ballard.
\newblock {\em Avoiding Communication in Dense Linear Algebra}.
\newblock PhD thesis, EECS Department, University of California, Berkeley, Aug
  2013.

\bibitem{Ballard14}
G.~Ballard, E.~Carson, J.~Demmel, M~Hoemmen, N.~Knight, and O.~Schwartz.
\newblock Communication lower bounds and optimal algorithms for numerical
  linear algebra.
\newblock {\em Acta Numerica}, 23:1--155, 2014.

\bibitem{CRT06}
E.~J. Cand\'{e}s, J.~Romberg, and T.~Tao.
\newblock Robust uncertainty principles: Exact signal reconstruction from
  highly incomplete frequency information.
\newblock {\em IEEE Trans. Inf. Theory}, 52(2):489--509, 2006.

\bibitem{C69}
L.~Cannon.
\newblock {\em A cellular computer to implement the {K}alman filter algorithm}.
\newblock PhD thesis, Montana State University, Bozeman, MN, 1969.

\bibitem{Carson15}
E.~Carson.
\newblock {\em Communication-Avoiding {K}rylov Subspace Methods in Theory and
  Practice}.
\newblock PhD thesis, EECS Department, University of California, Berkeley, Aug
  2015.

\bibitem{Chang11}
Chih-Chung Chang and Chih-Jen Lin.
\newblock {LIBSVM}: A library for support vector machines.
\newblock {\em ACM Transactions on Intelligent Systems and Technology},
  2:27:1--27:27, 2011.
\newblock Software available at \url{http://www.csie.ntu.edu.tw/~cjlin/libsvm}.

\bibitem{Chrono86}
A.~T. Chronopoulos.
\newblock {\em A class of parallel iterative methods implemented on
  multiprocessors}.
\newblock PhD thesis, Department of Computer Science, University of Illinois,
  Urbana, Illinois, 1986.

\bibitem{chronopoulos96}
A.~T. Chronopoulos and C.~D. Swanson.
\newblock Parallel iterative s-step methods for unsymmetric linear systems.
\newblock {\em Parallel Computing}, 22(5):623--641, 1996.

\bibitem{chronopoulos89a}
A.T. Chronopoulos and C.W. Gear.
\newblock On the efficient implementation of preconditioned s-step conjugate
  gradient methods on multiprocessors with memory hierarchy.
\newblock {\em Parallel Computing}, 11(1):37 -- 53, 1989.

\bibitem{chronopoulos89b}
A.T. Chronopoulos and C.W. Gear.
\newblock s-step iterative methods for symmetric linear systems.
\newblock {\em Journal of Computational and Applied Mathematics}, 25(2):153 --
  168, 1989.

\bibitem{CKPSSSSE93}
D.~Culler, R.~Karp, D.~Patterson, A.~Sahay, K.~E. Schauser, E.~Santos,
  T.~Subramonian, and R.~von Eicken.
\newblock Log{P}: Towards a realistic model of parallel computation.
\newblock {\em Proceedings of the fourth ACM SIGPLAN symposium on Principles
  and practice of parallel programming}, 28(7):1--12, 1993.

\bibitem{mpi4py}
L.~D. Dalcin, R.~R. Paz, P.~A. Kler, and A.~Cosimo.
\newblock Parallel distributed computing using python.
\newblock {\em Advances in Water Resources}, 34(9):1124 -- 1139, 2011.
\newblock New Computational Methods and Software Tools.

\bibitem{DGHL12}
J.~Demmel, L.~Grigori, M.~Hoemmen, and J.~Langou.
\newblock Communication-optimal parallel and sequential {QR} and {LU}
  factorizations.
\newblock {\em SIAM J. Sci. Comput.}, 34(1):A206--A239, 2012.

\bibitem{demmel07}
J.~Demmel, M.~Hoemmen, M.~Mohiyuddin, and K.~Yelick.
\newblock Avoiding communication in computing {K}rylov subspaces.
\newblock Technical Report UCB/EECS-2007-123, EECS Department, University of
  California, Berkeley, Oct 2007.

\bibitem{DFDM18}
A.~Devarakonda, K.~Fountoulakis, J.~Demmel, and M.~Mahoney.
\newblock Avoiding synchronization in first-order methods for sparse convex
  optimization.
\newblock Technical report, 2018.
\newblock Accepted for publication to the 32nd IEEE International Parallel and
  Distributed Processing Symposium.

\bibitem{lars}
B.~Efron, T.~Hastie, I.~Johnstone, and R.~Tibshirani.
\newblock Least angle regression.
\newblock {\em The Annals of Statistics}, 32(2):407--499, 2004.

\bibitem{FR15}
O.~Fercoq and P.~Richt\'{a}rik.
\newblock Accelerated, parallel, and proximal coordinate descent.
\newblock {\em SIAM J. Optim.}, 25(4):1997--2023, 2015.

\bibitem{HTTW07}
T.~Hastie, J.~Taylor, R.~Tibshirani, and G.~Walther.
\newblock Forward stagewise regression and the monotone lasso.
\newblock {\em Electron. J. Statist.}, 1:1--29, 2007.

\bibitem{HTF01}
T.~Hastie, R.~Tibshirani, and J.~Friedman.
\newblock {\em The Elements of Statistical Learning; Data mining, Inference and
  Prediction}.
\newblock Springer Verlag, New York, 2001.

\bibitem{Hoemmen10}
M.~Hoemmen.
\newblock {\em Communication-avoiding {K}rylov subspace methods}.
\newblock PhD thesis, University of California, Berkeley, 2010.

\bibitem{cocoa}
Martin Jaggi, Virginia Smith, Martin Tak\'{a}\v{c}, Jonathan Terhorst, Sanjay
  Krishnan, Thomas Hofmann, and Michael~I. Jordan.
\newblock Communication-efficient distributed dual coordinate ascent.
\newblock In {\em Proceedings of the 27th International Conference on Neural
  Information Processing Systems}, NIPS'14, pages 3068--3076, Cambridge, MA,
  USA, 2014. MIT Press.

\bibitem{kim92}
S.K. Kim and A.T. Chronopoulos.
\newblock An efficient nonsymmetric {L}anczos method on parallel vector
  computers.
\newblock {\em Journal of Computational and Applied Mathematics}, 42(3):357 --
  374, 1992.

\bibitem{mohiyuddin12}
M.~Mohiyuddin.
\newblock {\em Tuning Hardware and Software for Multiprocessors}.
\newblock PhD thesis, EECS Department, University of California, Berkeley, May
  2012.

\bibitem{mohiyuddin09}
M.~Mohiyuddin, M.~Hoemmen, J.~Demmel, and K.~Yelick.
\newblock Minimizing communication in sparse matrix solvers.
\newblock In {\em Proceedings of the Conference on High Performance Computing
  Networking, Storage and Analysis}, SC '09, pages 36:1--36:12, New York, NY,
  USA, 2009. ACM.

\bibitem{Musser1997}
D.~R. Musser.
\newblock Introspective sorting and selection algorithms.
\newblock {\em Softw. Pract. Exper.}, 27(8):983–993, August 1997.

\bibitem{DW17}
D.~Needell and T.~Woolf.
\newblock An asynchronous parallel approach to sparse recovery.
\newblock {\em Information Theory and Applications Workshop (ITA)}, 2017.

\bibitem{nesterov12}
Yu. Nesterov.
\newblock Efficiency of coordinate descent methods on huge-scale optimization
  problems.
\newblock {\em SIAM Journal on Optimization}, 22(2):341--362, 2012.

\bibitem{NG04}
A.~Y. Ng.
\newblock Feature selection, {L}1 vs. {L}2 regularization, and rotational
  invariance.
\newblock pages 78--, 2004.

\bibitem{PHP03}
N.~Park, B.~Hong, and V.~K. Prasanna.
\newblock Tiling, block data layout, and memory hierarchy performance.
\newblock {\em IEEE Transactions on Parallel and Distributed Systems},
  14(7):640--654, 2003.

\bibitem{PH13}
D.~A. Patterson and J.~L. Hennessy.
\newblock {\em Computer organization and design: the hardware/software
  interface}.
\newblock Morgan Kaufman, 2013.

\bibitem{Q04}
M.~J. Quinn.
\newblock {\em Parallel Programming in {C} with {MPI} and {O}pen{MP}}.
\newblock McGraw-Hill, New York, NY, 2004.

\bibitem{recht11}
B.~Recht, C.~R\'{e}, S.~Wright, and F.~Niu.
\newblock Hogwild: A lock-free approach to parallelizing stochastic gradient
  descent.
\newblock In {\em Advances in Neural Information Processing Systems}, pages
  693--701, 2011.

\bibitem{richtarik14}
P.~Richt{\'a}rik and M.~Tak{\'a}{\v{c}}.
\newblock Iteration complexity of randomized block-coordinate descent methods
  for minimizing a composite function.
\newblock {\em Mathematical Programming}, 144(1):1--38, 2014.

\bibitem{SDM10}
J.~Shalf, S.~Dosanjh, and J.~Morrison.
\newblock Exascale computing technology challenges.
\newblock {\em International Conference on High Performance Computing for
  Computational Science - VECPAR 2010}, 6449:1--25, 2010.

\bibitem{Solomonik14}
E.~Solomonik.
\newblock {\em Provably efficient algorithms for numerical tensor algebra}.
\newblock PhD thesis, EECS Department, University of California, Berkeley, Aug
  2014.

\bibitem{TRG2005}
R.~Thakur, R.~Rabenseifner, and W.~Gropp.
\newblock Optimization of collective communication operations in {MPICH}.
\newblock {\em The International Journal of High Performance Computing
  Applications}, 19(1), 2005.

\bibitem{Tib96}
R.~Tibshirani.
\newblock Regression shrinkage and selection via lasso.
\newblock {\em J. Roy. Statist. Soc. Ser. B}, 58:267--288, 1996.

\bibitem{rosendale83}
J.~Van~Rosendale.
\newblock {\em Minimizing inner product data dependencies in conjugate gradient
  iteration}.
\newblock IEEE Computer Society Press, Silver Spring, MD, Jan 1983.

\bibitem{Weis80}
S.~Weisberg.
\newblock {\em Applied linear regression}.
\newblock Wiley, New York, 1980.

\bibitem{williams14}
S.~Williams, M.~Lijewski, A.~Almgren, B.~Van~Straalen, E.~Carson, N.~Knight,
  and J.~Demmel.
\newblock s-step {K}rylov subspace methods as bottom solvers for geometric
  multigrid.
\newblock In {\em Parallel and Distributed Processing Symposium, 2014 IEEE 28th
  International}, pages 1149--1158. IEEE, 2014.

\bibitem{Wright15}
S.~J. Wright.
\newblock Coordinate descent algorithms.
\newblock {\em Math. Program.}, 151(1):3--34, June 2015.

\end{thebibliography}

\end{document}